\documentclass{article}

\usepackage[final]{corl_2021} %
\usepackage{xspace}
\usepackage{url}            %
\usepackage{booktabs}       %
\usepackage{amsfonts}       %
\usepackage{nicefrac}       %
\usepackage{microtype}      %
\usepackage{algorithm}
\usepackage{algpseudocode}
\usepackage{multirow}
\usepackage{enumitem}
\usepackage{amsmath,amssymb}
\usepackage{dsfont}
\usepackage{mathtools}
\usepackage{wrapfig}
\usepackage{graphicx}

\usepackage[export]{adjustbox}
\usepackage{subcaption}
\usepackage[T1]{fontenc}
\usepackage{xcolor} 
\usepackage{siunitx}
\usepackage{soul}
\hypersetup{
breaklinks=true,colorlinks,citecolor=blue
}
\usepackage[title]{appendix}

\title{A System for General In-Hand Object Re-Orientation}

\author{ Tao Chen, Jie Xu, Pulkit Agrawal\\
 Massachusetts Institute of Technology\\
 \tt{\{taochen, jiex, pulkitag\}@mit.edu}}

\newcommand{\figref}[1]{\figurename~\ref{#1}}
\newcommand{\tblref}[1]{Table~\ref{#1}}
\newcommand{\secref}[1]{Section~\ref{#1}}
\newcommand{\equaref}[1]{Equation~(\ref{#1})}
\newcommand{\algoref}[1]{Algorithm~\ref{#1}}
\newcommand{\appref}[1]{Appendix~\ref{#1}}

\newcommand{\ie}{\textit{i.e.}\xspace}
\newcommand{\videourl}{\url{https://taochenshh.github.io/projects/in-hand-reorientation}}

\newcommand{\shadow}{Shadow Hand\xspace}
\newcommand{\isaac}{Isaac Gym\xspace}
\newcommand{\gaussian}{\mathcal{N}}
\newcommand{\uniform}{\mathcal{U}}
\newcommand{\expunilog}{\mathcal{E}}

\newcommand{\expsymbol}{\mathcal{E}}
\newcommand{\stusymbol}{\mathcal{S}}
\newcommand{\expertrnn}{\pi_R^\expsymbol}
\newcommand{\expertmlp}{\pi_M^\expsymbol}
\newcommand{\expert}{\pi^\expsymbol}
\newcommand{\student}{\pi^\stusymbol}
\newcommand{\studentrnn}{\pi_R^\stusymbol}
\newcommand{\studentmlp}{\pi_M^\stusymbol}
\newcommand{\studentspace}{\mathbb{S}^\stusymbol}
\newcommand{\expertspace}{\mathbb{S}^\expsymbol}
\begin{document}
\maketitle

\vspace{-1.2cm}

\begin{figure}[!h]
    \centering
    \includegraphics[width=\linewidth]{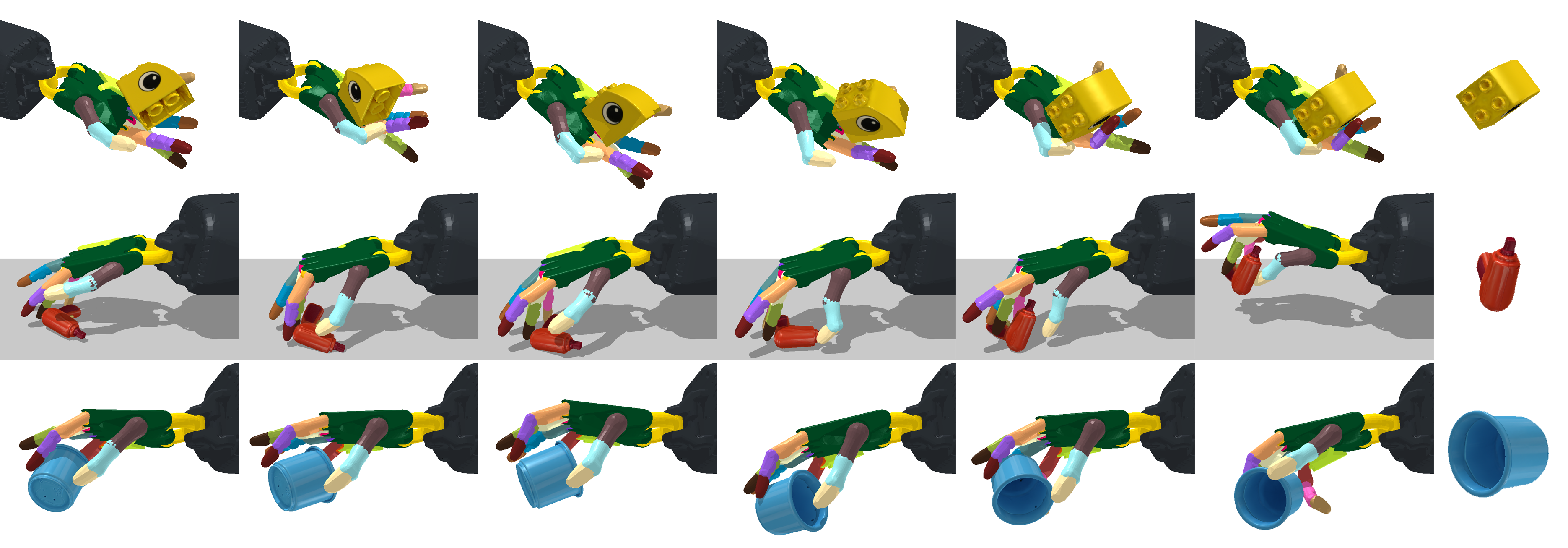}
    \caption{We present a simple framework for learning policies for reorienting a large number of objects in scenarios where the (1) hand faces upward, (2) hand faces downward with a table below the hand and (3) without the support of the table. The object orientation in the rightmost image in each row shows the target orientation.}
    \label{fig:teaser}
\end{figure}

\vspace{-0.5cm}

\begin{abstract}
    In-hand object reorientation has been a challenging problem in robotics due to high dimensional actuation space and the frequent change in contact state between the fingers and the objects. We present a simple model-free framework that can learn to reorient objects with both the hand facing upwards and downwards. We demonstrate the capability of reorienting over $2000$ geometrically different objects in both cases. The learned policies show strong zero-shot transfer performance on new objects. We provide evidence that these policies are amenable to real-world operation by distilling them to use observations easily available in the real world. The videos of the learned policies are available at: \videourl.
\end{abstract}

\keywords{Dexterous manipulation, in-hand manipulation, object reorientation} 

\vspace{-0.25cm}
\section{Introduction}
\vspace{-0.15cm}

A common maneuver in many tasks of daily living is to pick an object, reorient it in hand and then either place it or use it as a tool. Consider three simplified variants of this maneuver shown in Figure~\ref{fig:teaser}. 
The task in the top row requires an upward-facing multi-finger hand to manipulate an \textit{arbitrary} object in a random orientation to a goal configuration shown in the rightmost column. The next two rows show tasks where the hand is facing downward and is required to reorient the object either using the table as a support or without the aid of any support surface respectively. The last task is the hardest because the object is in an intrinsically unstable configuration owing to the downward gravitational force and lack of support from the palm. Additional challenges in performing such manipulation with a multi-finger robotic hand stem from the control space being high-dimensional and reasoning about the multiple transitions in the contact state between the finger and the object. Due to its practical utility and several unsolved issues, in-hand object reorientation remains an active area of research. 

Past work has tackled the in-hand reorientation problem via several approaches: (i) The use of analytical models with powerful trajectory optimization methods~\cite{mordatch2012contact,bai2014dexterous,kumar2014real}. While these methods demonstrated remarkable performance, the results were largely in simulation with simple object geometries and required detailed knowledge of the object model and physical parameters. As such, it remains unclear how to scale these methods to real-world and generalize to new objects. Another line of work has employed (ii) model-based reinforcement learning~\cite{kumar2016optimal,nagabandi2020deep}; or (iii) model-free reinforcement learning with~\cite{gupta2016learning,kumar2016learning,rajeswaran2017learning,radosavovic2020state} and without expert demonstrations~\cite{mudigonda2018investigating,andrychowicz2020learning,akkaya2019solving,zhu2019dexterous}. While some of these works demonstrated learned skills on real robots, it required use of additional sensory apparatus not readily available in the real-world (e.g., motion capture system) to infer the object state, and the learned policies did not generalize to diverse objects. Furthermore, most prior methods operate in the simplified setting of the hand facing upwards. The only exception is pick-and-place, but it does not involve any in-hand re-orientation. A detailed discussion of prior research is provided in Section~\ref{sec:related}. 

In this paper, our goal is to study the object reorientation problem with a multi-finger hand in its general form. We desire (a) manipulation with hand facing upward or downward; (b) the ability of using external surfaces to aid manipulation; (c) the ability to reorient objects of novel shapes to arbitrary orientations; (d) operation from sensory data that can be easily obtained in the real world such as RGBD images and joint positions of the hand. While some of these aspects have been individually demonstrated in prior work, we are unaware of any published method that realizes all four. Our main contribution is building a system that achieves the desiderata. The core of our framework is a model-free reinforcement learning with three key components: teacher-student learning, gravity curriculum, and stable initialization of objects. Our system requires no knowledge of object or manipulator models, contact dynamics or any special pre-processing of sensory observations. We experimentally test our framework using a simulated Shadow hand. Due to the scope of the problem and the ongoing pandemic, we limit our experiments to be in simulation. However, we provide evidence indicating that the learned policies can be transferred to the real world in the future. 

\noindent{\textbf{A Surprising Finding}:} While seemingly counterintuitive, we found that policies that have no access to shape information can manipulate a large number of previously unseen objects in all the three settings mentioned above. At the start of the project, we hypothesized that developing visual processing architecture for inferring shape while the robot manipulates the object would be the primary research challenge. On the contrary, our results show that it is possible to learn control strategies for general in-hand object re-orientation that are shape-agnostic. Our results, therefore, suggest that visual perception may be less important for in-hand manipulation than previously thought. Of course, we still believe that the performance of our system can be improved by incorporating shape information. However, our findings suggest a different framework of thinking: a lot can be achieved without vision, and that vision might be the icing on the cake instead of the cake itself.

\section{Method}
\label{sec:method}

We found that \textit{simple} extensions to existing techniques in robot learning can be used to construct a system for general object reorientation. First, to avoid explicit modeling of non-linear and frequent changes in the contact state between the object and the hand, we use model-free reinforcement learning (RL). An added advantage is that model-free RL is amenable to direct operation from raw point cloud observations, which is preferred for real-world deployment. We found that better policies can be trained faster using \textit{privileged} state information such as the velocities of the object/fingertips that is easily available in the simulator but not in the real world. To demonstrate the possibility of transferring learned policies to the real world in the future, we overcome the need for privileged information using the idea of \textit{teacher-student training}~\citep{lee2020learning, chen2020learning}. In this framework, first, an expert or \textit{teacher} policy ($\expert$) is trained using privileged information. Next, the \textit{teacher} policy guides the learning of a \textit{student} policy ($\student$) that only uses sensory inputs available in the real world. Let the state space corresponding to  $\expert$ and $\student$ be $\expertspace$ and $\studentspace$ respectively. In general, $\expertspace\neq \studentspace$. 

We first trained the teacher policy to reorient more than two thousand objects of diverse shapes (see \secref{sec:learn_teacher}). Next, we detail the method for distilling $\expert$ to a student policy using a reduced state space consisting of only the joint positions of the hand, the object position, and the difference in orientation from the goal configuration (see \secref{subsec:student_nonvision}). However, in the real world, even the object position and relative orientation must be inferred from sensory observation. Not only does this process require substantial engineering effort (e.g., a motion capture or a pose estimation system), but also inferring the pose of a symmetric object is prone to errors. This is because a symmetric object at multiple orientations looks exactly the same in sensory space such as RGBD images. 

To mitigate these issues, we further distill $\expert$ to operate directly from the point cloud and position of all the hand joints (see \secref{subsec:student_vision_policy}). We propose a simple modification that generalizes an existing 2D CNN architecture~\citep{espeholt2018impala} to make this possible. 

The procedure described above works well for manipulation with the hand facing upwards and downwards when a table is available as support. However, when the hand faces downward without an underlying support surface, we found it important to initialize the object in a stable configuration. Finally, because gravity presents the primary challenge in learning policies with a downward-facing hand, we found that training in a curriculum where gravity is slowly introduced (i.e., \textit{gravity curriculum}) substantially improves performance. These are discussed in Section~\ref{subsec: hand_down}.

\vspace{-0.2cm}
\subsection{Learning the teacher policy}
\label{sec:learn_teacher}
\vspace{-0.1cm}

We use model-free RL to learn the teacher policy ($\expert$) for reorienting an object ($\{O_i | i=1, ..., N\}$) from an initial orientation $\alpha^o_0$ to a target orientation $\alpha^g$ ($\alpha^o_0\neq\alpha^g$). At every time step $t$, the agent observes the state $s_t$, executes the action $a_t$ sampled from the policy $\expert$, and receives a reward $r_t$. $\expert$ is optimized to maximize the expected discounted return: $\pi=\arg\max_{\expert}\mathbb{E} \big[ \sum^{T-1}_{t=0}  \gamma^{t} r_t\big]$, where $\gamma$ is the discount factor. The task is successful if the angle difference $\Delta\theta$ between the object's current ($\alpha^o_t$) and the goal orientation ($\alpha^g$) is smaller than the threshold value $\Bar{\theta}$, \ie, $\Delta\theta\leq \Bar{\theta}$. 

To encourage the policy to be smooth, the previous action is appended to the inputs to the policy (\ie, $a_t=\expert(s_t, a_{t-1})$) and large actions are penalized in the reward function. We experiment with two architectures for $\expert$: (1) an MLP policy $\pi_M$, 
(2) an RNN policy $\pi_R$. 
We use PPO~\citep{schulman2017proximal} to optimize $\expert$. 
More details about the training are in \secref{app_subsec:network_arch} and \secref{app_sec:training_details} in the appendix.

\textbf{Observation and action space}: We define $\expertspace$ to include joint, fingertip, object, and goal information as detailed in \tblref{tbl:state_def} in the appendix. Note that $\expertspace$ does not include object shape or information about friction, damping, contact states between the fingers and the object, etc.
We control the joint movements by commanding the relative change in the target joint angle ($q_{t}^{target}$) on each actuated joint (action $a_t\in\mathbb{R}^{20}$): $q_{t+1}^{target} = q_{t}^{target} + a_t\times \Delta t$, where $\Delta t$ is the control time step. We clamp the action command if necessary to make sure $|\Delta q_t^{target}|\leq\SI{0.33}{\radian}$. The control frequency is \SI{60}{\hertz}.

\textbf{Dynamics randomization}: Even though we do not test our policies on a real robot, we train and evaluate policies with domain randomization~\citep{tobin2017domain} to provide evidence that our work has the potential to be transferred to a real robotic system in the future. We randomize the object mass, friction coefficient, joint damping and add noise to the state observation $s_t$ and the action $a_t$. More details about domain randomization are provided in \tblref{tbl:dyn_ran_params} in the appendix.

\subsection{Learning the student policy}
\label{subsec:student_pol}
We distill the teacher $\expert$ into the student policy $\student$ using Dagger~\citep{ross2011reduction}, a learning-from-demonstration method that overcomes the covariate shift problem. We optimize $\student$ by minimizing the KL-divergence between $\student$ and $\expert$: $\student=\arg\min_{\student}D_{KL}\left(\expert(s^\expsymbol_t, a_{t-1}) || \student(s^\stusymbol_t, a_{t-1}))\right)$. Based on observation data available in real-world settings, we investigate two different choices of $\studentspace$.

\subsubsection{Training student policy from low-dimensional state}
\label{subsec:student_nonvision}
In the first case, we consider a non-vision student policy $\student(s_t^\stusymbol, a_{t-1})$.   $s_t^\stusymbol\in\mathbb{R}^{31}$ includes the joint positions $q_t\in\mathbb{R}^{24}$, object position $p^o_t\in\mathbb{R}^3$, quaternion difference between the object's current and target orientation $\beta_t\in\mathbb{R}^4$. In this case, $\studentspace\subset\expertspace$, and we assume the availability of object pose information, but do not require velocity information. We use the same MLP and RNN network architectures used for $\expert$ on $\student$ except the input dimension changes as the state dimension is different.

\subsubsection{Training student policy from vision}
\label{subsec:student_vision_policy}

In the second case, $\studentspace$ only uses direct observations from RGBD cameras and the joint position ($q_t$) of the robotic hand. We convert the RGB and Depth data into a colored point cloud using a pinhole camera model~\cite{andrew2001multiple}. Our vision policy takes as input the voxelized point cloud of the scene $W_t$, $q_t$, and previous action command $a_{t-1}$, and outputs the action $a_t$, \ie, $a_t=\student(W_t, q_t, a_{t-1})$.

\begin{figure}[!tb]
    \centering
    \includegraphics[width=\linewidth]{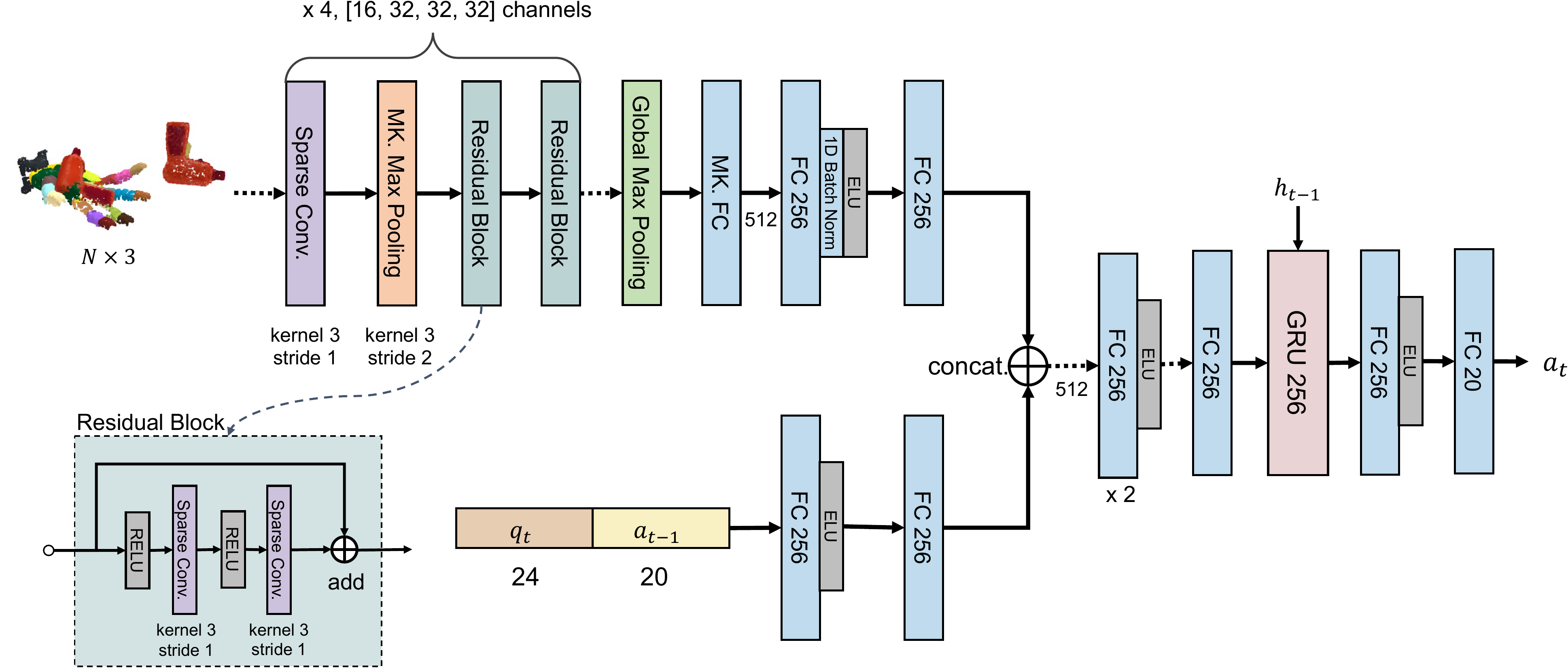}
    \caption{Visual policy architecture. MK stands for Minkowski Engine. $q_t$ is the joint positions and $a_t$ is the action at time step $t$.}
    \label{fig:vision_policy_arch}
\end{figure}

\textbf{Goal specification}: To avoid manually defining per-object coordinate frame for specifying the goal quaternion, we provide the goal to the policy as an object point cloud rotated to the desired orientation $W^g$, \ie, we only show the policy how the object should look like in the end (see the top left of \figref{fig:vision_policy_arch}). The input to $\student$ is the point cloud $W_t = W^s_t\cup W^g$ where $W_t^s$ is the actual point cloud of the current scene obtained from the cameras. Details of obtaining $W_g$ are in \secref{app_sec:training_details}. 

\textbf{Sparse3D-IMPALA-Net}: To convert a voxelized point cloud into a lower-dimensional feature representation, we use a sparse convolutional neural network. We extend the IMPALA policy architecture~\citep{espeholt2018impala} for processing RGB images to process colored point cloud data using 3D convolution. Since many voxels are unoccupied, the use of regular 3D convolution substantially increases computation time. Hence, we use Minkowski Engine~\citep{choy20194d}, a PyTorch library for sparse tensors, to design a 3D version of IMPALA-Net with sparse convolutions (\textit{Sparse3D-IMPALA-Net})\footnote{We also experimented with a 3D sparse convolutional network based on ResNet18, and found that 3D IMPALA-Net works better.}. 
The Sparse3D-IMPALA network takes as input the point cloud $W_t$, and outputs an embedding vector which is concatenated with the embedding vector of $(q_t, a_{t-1})$. Afterward, a recurrent network is used and outputs the action $a_t$.
The detailed architecture is illustrated in \figref{fig:vision_policy_arch}.

\textbf{Mitigating the object symmetry issue}: $\expert$ is trained with the the ground-truth state information $s_t^\expsymbol$ including the object orientation $\alpha_t^o$ and goal orientation $\alpha^g$. %
The vision policy does not take any orientation information as input. If an object is symmetric, the two different orientations of the object may correspond to the same point cloud observation. This makes it problematic to use the difference in orientation angles ($\Delta\theta\leq \Bar{\theta}$) as the stopping and success criterion.
To mitigate this issue, we use Chamfer distance~\citep{fan2017point} to compute the distance between the object point cloud in $\alpha^o_t$ and the goal point cloud (i.e., the object rotated by $\alpha^g$) as the evaluation criterion. The Chamfer distance is computed as $d_{C}=\sum_{a\in W_t^o}\min_{b\in W^g} \left\Vert a-b\right\Vert_2^2 + \sum_{b\in W^g}\min_{a\in W_t^o} \left\Vert a-b\right\Vert_2^2$, where $W_t^o$ is the object point cloud in its current orientation. Both $W_t^o$ and $W^g$ are scaled to fit in a unit sphere for computing $d_C$. We check Chamfer distance in each rollout step. If $d_C\leq \Bar{d}_C$ ($\Bar{d}_C$ is a threshold value for $d_C$), we consider the episode to be successful. Hence, the success criterion is $(\Delta\theta\leq\Bar{\theta}) \lor (d_C\leq \Bar{d}_C)$. In training, if the success criterion is satisfied, the episode is terminated and used for updating $\student$.

\section{Experimental Setup}
\label{sec:env_setup}
We use the simulated \shadow~\citep{shadowhand} in NVIDIA \isaac~\citep{makoviychuk2021isaac}. \shadow is an anthropomorphic robotic hand with 24 degrees of freedom (DoF). We assume the base of the hand to be fixed. Twenty joints are actuated by agonist–antagonist tendons and the remaining four are under-actuated. 

\textbf{Object datasets}:
We use the EGAD dataset~\citep{morrison2020egad} and YCB dataset~\citep{calli2015ycb} that contain objects with diverse shapes (see \figref{fig:egad_ycb_dataset}) for in-hand manipulation experiments. EGAD contains $2282$ geometrically diverse textureless object meshes, while the YCB dataset includes textured meshes for objects of daily life with different shapes and textures. We use the $78$ YCB object models collected with the Google scanner. Since most YCB objects are too big for in-hand manipulation, we proportionally scale down the YCB meshes. To further increase the diversity of the datasets, we create $5$ variants for each object mesh by randomly scaling the mesh. More details of the object datasets are in \secref{app_subsec:dataset}.

\begin{figure}[!tb]
\vspace{-0.5cm}
\centering
\begin{subfigure}{0.24\textwidth}
  \centering
  \includegraphics[width=\linewidth]{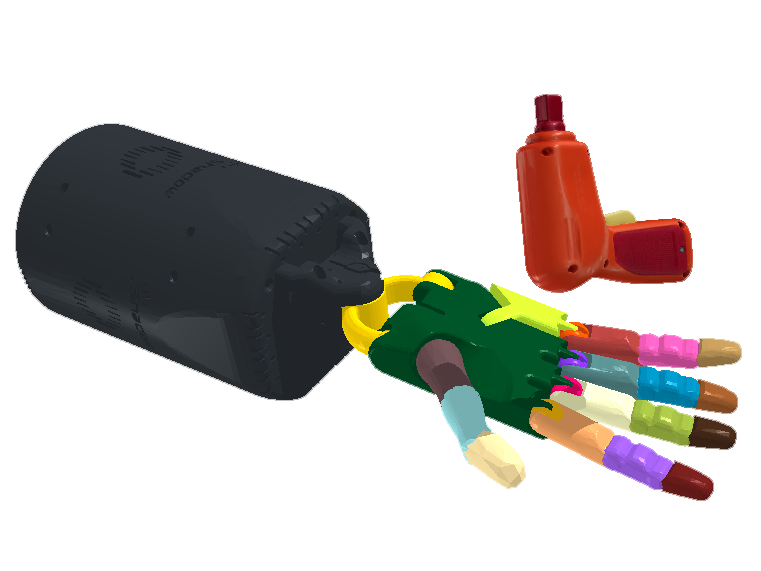}
  \caption{}
  \label{fig:up}
\end{subfigure}%
\begin{subfigure}{0.24\textwidth}
  \centering
  \includegraphics[width=\linewidth]{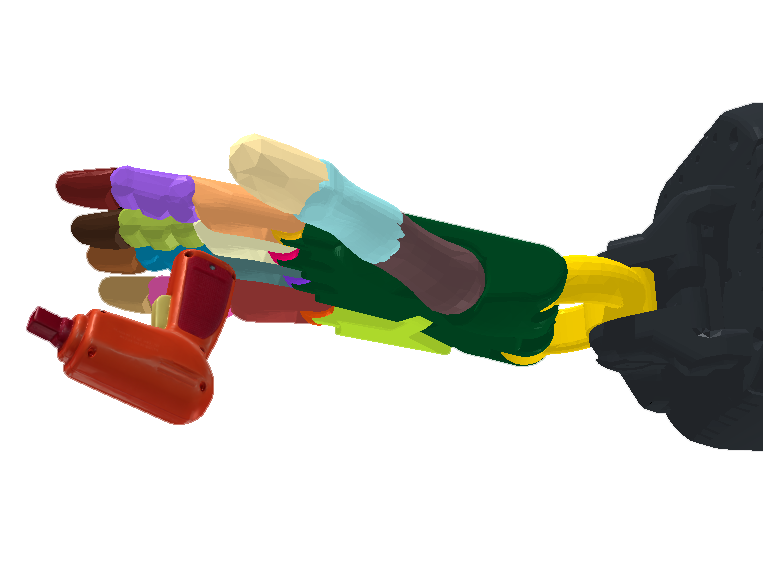}
  \caption{}
  \label{fig:down_random}
\end{subfigure}%
\begin{subfigure}{0.24\textwidth}
  \centering
  \includegraphics[width=\linewidth]{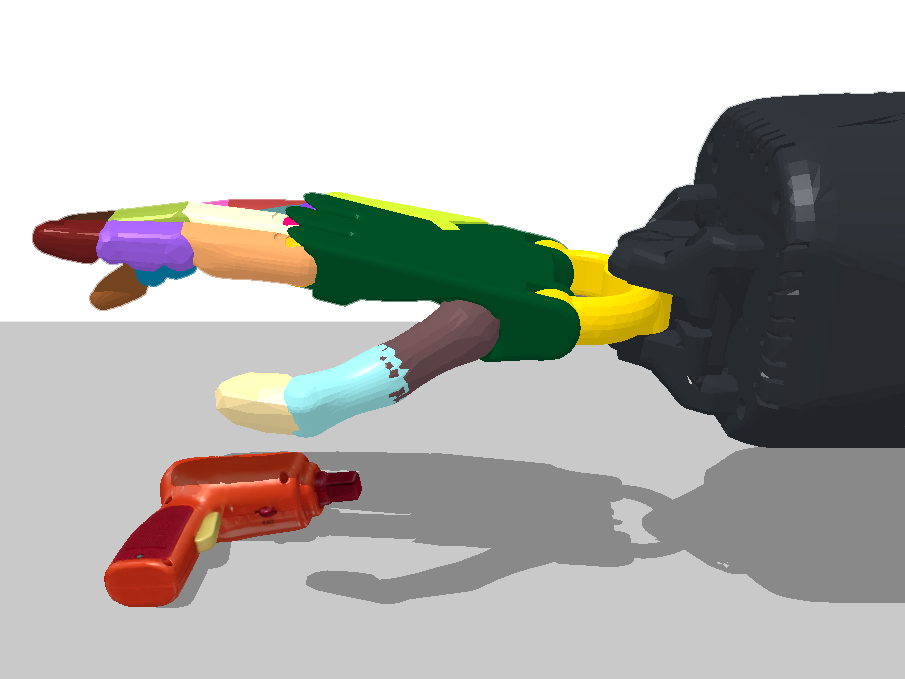}
  \caption{}
  \label{fig:down_table}
\end{subfigure}%
\begin{subfigure}{0.24\textwidth}
  \centering
  \includegraphics[width=\linewidth]{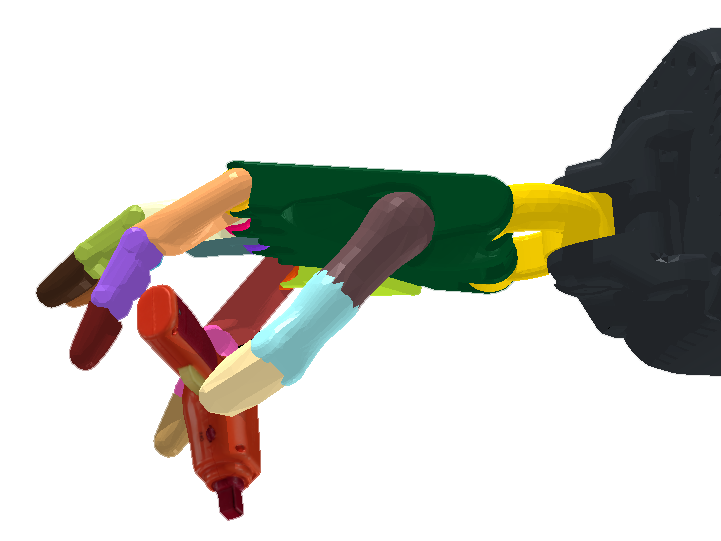}
  \caption{}
  \label{fig:down_pregrasp}
\end{subfigure}\\
\caption{Examples of initial poses of the hand and object. \textbf{(a)}: hand faces upward. \textbf{(b)}, \textbf{(c)}, \textbf{(d)}: hand faces downward. \textbf{(b)}: both the hand and the object are initialized with random poses . \textbf{(c)}: there is a table below the hand. \textbf{(d)}: the hand and the object are initialized  from the lifted poses. }
\label{fig:exp_setup}

\end{figure}

\textbf{Setup for visual observations}:
For the vision experiments, we trained policies for the scenario of hand facing upwards. We place two RGBD cameras above the hand (\figref{fig:camera_pos}). The data from these cameras is combined to create a point cloud observation of the scene~\citep{andrew2001multiple}. For downstream computation, the point cloud is voxelized with a resolution of \SI{0.003}{\meter}. 

\textbf{Setup with the upward facing hand}: We first consider the case where the \shadow faces upward and is required to reorient objects placed in the hand (see \figref{fig:up}). We use the coordinate system where the $z$-axis points vertically upward and the $xy$-plane denotes the horizontal plane. The object pose is initialized with the following procedure: $xy$ position of the object's center of mass (COM) $p_{0,xy}^o$ is randomly sampled from a square region of size $\SI{0.09}{\meter}\times\SI{0.09}{\meter}$. The center of this square is approximately located on the intersection of the middle finger and the palm so that the sampling region covers both the fingers and the palm. The $z$ position of the object is fixed to \SI{0.13}{\meter} above the base of the hand to ensure that the object does not collide with the hand at initialization. The initial and goal orientations are randomly sampled from the full $SO(3)$ space.

\textbf{Setup with the downward facing hand}: Next, we consider the cases where the hand faces downward. We experiment with two scenarios: with and without a table below the hand. In the first case, we place a table with the tabletop being \SI{0.12}{\meter} below the hand base. We place objects in a random pose between the hand and the table so that the objects will fall onto the table. We will describe the setup for the second case (without a table) in \secref{subsec:reorient_air}. 

\textbf{Evaluation criterion}: For non-vision experiments, a policy rollout is considered a success if $\theta\leq\Bar{\theta}$. $\Bar{\theta}=\SI{0.1}{\radian}$. For vision experiments, we also check $d_C\leq \Bar{d}_C$ as another criterion and $\Bar{\theta}=\SI{0.2}{\radian}, \Bar{d}_C=0.01$.
The initial and goal orientation are randomly sampled from $SO(3)$ space in all the experiments.   We report performance as the percentage of successful episodes when the agent is tasked to reorient each training object $100$ times from arbitrary start to goal orientation. We report the mean and standard deviation of success rate from $3$ seeds.

\section{Results}
\label{sec:results}

We evaluate the performance of reorientation policies with the hand facing upward and downward. Further we analyze the generalization of the learned policies to unseen object shapes. 

\subsection{Reorient objects with the hand facing upward}
\label{subsec: result_hand_up}

\paragraph{Train a teacher policy with full-state information}
We train our teacher MLP and RNN policies using the full state information using all objects in the EGAD and YCB datasets separately. The progression of success rate during training is shown in \figref{fig:egad_learn_upward} in \appref{app_subsec:hand_up} . \figref{fig:egad_learn_upward} also shows that using privileged information substantially speeds up policy learning. 
Results reported in \tblref{tbl:up_success_rate_short} indicate that the RNN policies achieve a success rate greater than 90\% on the EGAD dataset (entry B1) and greater than 80\% on the YCB dataset (entry G1) without any explicit knowledge of the object shape\footnote{More quantitative results on the MLP policies are available in \tblref{tbl:up_success_rate} in the appendix.}. This result is surprising because apriori one might believe that shape information is important for in-hand reorientation of diverse objects.

The visualization of policy rollout reveals that the agent employs a clever strategy that is invariant to object geometry for re-orienting objects. The agent throws the object in the air with a spin and catches it at the precise time when the object's orientation matches the goal orientation. Throwing the object with a spin is a dexterous skill that automatically emerges! One possibility for the emergence of this skill is that we used very light objects. This is not true because we trained with objects in the range of 50-150g which spans many hand-held objects used by humans (e.g., an egg weighs about 50g, a glass cup weighs around 100g, iPhone 12 weighs 162g, etc.). To further probe this concern, we evaluated zero-shot performance on objects weighing up to 500g\footnote{We change the mass of the YCB objects to be in the range of $[0.3,  0.5]$kg, and test $\expertrnn$ from the YCB dataset on these new objects. The success rate is around $75\%$.} and found that the learned policy can successfully reorient them. We provide further analysis in the appendix showing that forces applied by the hand for such manipulation are realistic. While there is still room for the possibility that the learned policy is exploiting the simulator to reorient objects by throwing them in the air, our analysis so far indicates otherwise. 

Next, to understand the failure modes, we collected one hundred unsuccessful trajectories on YCB dataset and manually analyzed them. The primary failure is in manipulating long, small, or thin objects, which accounts for $60\%$ of all errors. In such cases, either the object slips through the fingers and falls, or is hard to be manipulated when the objects land on the palm. Another cause of failures ($19\%$) is that objects are reoriented close to the goal orientation but not close enough to satisfy $\Delta\theta<\Bar{\theta}$.
Finally, the performance on YCB is lower than EGAD because objects in the YCB dataset are more diverse in their aspect ratios. Scaling these objects by constraining $l_{\max} \in[0.05, 0.12]$m (see \secref{sec:env_setup}) makes some of these objects either too small, too big, or too thin and consequently results in failure (see \figref{fig:failure_cases_up}). A detailed object-wise quantitative analysis of performance is reported in appendix \figref{fig:success_rate_ycb_up}. Results confirm that sphere-like objects such as tennis balls and orange are easiest to reorient, while long/thin objects such as knives and forks are the hardest to manipulate. 

\renewcommand{\arraystretch}{1.1}
\begin{table}
\centering
\caption{\small{Success rates (\%) of policies tested on different dynamics distribution. $\Bar{\theta}=0.1$\si{\radian}. DR: domain randomization and observation/action noise. X$\rightarrow$Y: distill policy X into policy Y. The full table is in \tblref{tbl:up_success_rate}.}}
\label{tbl:up_success_rate_short}
\resizebox{\columnwidth}{!}{
\begin{tabular}{cccc|ccc} 
\hline
\multicolumn{4}{c}{}                                                                                                       & 1                                   & 2               & 3  \\ 
\hline
\multirow{2}{*}{Exp. ID} & \multirow{2}{*}{Dataset} & \multirow{2}{*}{State} & \multicolumn{1}{c}{\multirow{2}{*}{Policy}} & \multicolumn{2}{c}{Train without DR}                  & Train with DR           \\ 
\cline{5-7}
                         &                          &                        & \multicolumn{1}{c}{}                        & \multicolumn{1}{l}{Test without DR} & Test with DR    & Test with DR            \\ 
\hline
B                        & \multirow{2}{*}{EGAD}    & Full state             & RNN                                         & 95.95 $\pm$ 0.8                     & 84.27 $\pm$ 1.0 & 88.04 $\pm$ 0.6         \\
E                        &                          & Reduced state          & RNN$\rightarrow$RNN                         & 91.96 $\pm$ 1.5                     & 78.30 $\pm$ 1.2 & 80.29 $\pm$ 0.9         \\ 
\hline
G                        & \multirow{2}{*}{YCB}     & Full state             & RNN                                         & 80.40 $\pm$ 1.6                     & 65.16 $\pm$ 1.0 & 72.34 $\pm$ 0.9         \\
J                        &                          & Reduced state          & RNN$\rightarrow$RNN                         & 81.04 $\pm$ 0.5                     & 64.93 $\pm$ 0.2 & 65.86 $\pm$ 0.7         \\
\hline
\end{tabular}
}
\end{table}

\renewcommand{\arraystretch}{1}

\vspace{-0.2cm}
\paragraph{Train a student policy with a reduced state space}
The student policy state is $s_t^\stusymbol\in\mathbb{R}^{31}$. In \tblref{tbl:up_success_rate_short} (entries E1 and J1), we can see that $\studentrnn$ can get similarly high success rates as $\expertrnn$. The last two columns in \tblref{tbl:up_success_rate_short} also show that the policy is more robust to dynamics variations and observation/action noise after being trained with domain randomization.

\subsection{Reorient objects with the hand facing downward}
\label{subsec: hand_down}

The results above demonstrate that when the hand faces upwards, RL can be used to train policies for reorienting a diverse set of geometrically different objects. A natural question to ask is, does this still hold true when the hand is flipped upside down? Intuitively, this task is much more challenging because the objects will immediately fall down without specific finger movements that stabilize the object. Because with the hand facing upwards, the object primarily rests on the palm, such specific finger movements are not required. Therefore, the hand facing downwards scenario presents a much harder exploration challenge. To verify this hypothesis, we trained a policy with the downward-facing hand, objects placed underneath the hand (see \figref{fig:down_random}), and using the same reward function (\equaref{eqn:rot_reward}) as before. Unsurprisingly, the success rate was $0\%$. The agent's failure can be attributed to policy needing to learn to both stabilize the object under the effect of gravity and simultaneously reorient it. Deploying this policy simply results in an object falling down, confirming the hard-exploration nature of this problem.

\subsubsection{Reorient objects on a table}
\label{subsec:reorient_table}
To tackle the hard problem of reorienting objects with the hand facing downward, we started with a simplified task setup that included a table under the hand (see \figref{fig:down_table}). Table eases exploration by preventing the objects from falling. We train $\expertmlp$ using the same reward function \equaref{eqn:rot_reward} on objects sampled from the EGAD and YCB datasets. The success rate using an MLP policy using full state information for EGAD and YCB is $95.31\%\pm 0.9\%$ and $81.59\%\pm 0.7\%$ respectively. Making use of external support for in-hand manipulation has been a challenging problem in robotics. Prior work approach this problem by building analytical models and constructing motion cones~\citep{chavan2018hand}, which is challenging for objects with complex geometry. Our experiments show that model-free RL provides an effective alternative for learning manipulation strategies capable of using external support surfaces.

\subsubsection{Reorient objects in air with hand facing downward}
\label{subsec:reorient_air}
To enable the agent to operate in more general scenarios, we tackled the re-orientation problem with the hand facing downwards and without any external support. In this setup, one might hypothesize that object shape information (e.g., from vision) is critical because finding the strategy in \secref{subsec: result_hand_up} is not easy when the hand needs to overcome gravity and stabilize the object while reorienting it. We experimentally verify that even in this case, the policies achieve a reasonably high success rate without any knowledge of object shape.

\textbf{A good pose initialization is what you need}:
The difficulty of directly training the RL policies when the hand faces downward is mainly because of the hard-exploration issue in learning to catch the objects that are moving downward. However, catching is not necessary for the reorientation. Even for human, we only reorient the object after we grasp it. More specifically, we first train an object-lifting policy to lift objects from the table, collect the ending state (joint positions $q_T$, object position $p^o_T$ and orientation $\alpha^o_T$) in each successful lifting rollout episode, and reset the hand and objects to these states (velocities are all $0$) for the pose initialization in training the reorientation policy.  The objects have randomly initialized poses and are dropped onto the table. We trained a separate RNN policy for each dataset using the reward function in \secref{app_sec:training_details}. The success rate on the EGAD dataset is $97.80\%$, while the success rate on the YCB dataset is $90.11\%$. Note that objects need to be grasped first to be lifted. Our high success rates on object lifting also indicate that using an anthropomorphic hand makes object grasping an easy task, while many prior works~\citep{mousavian2019graspnet, mahler2017dex} require much more involved training techniques to learn grasping skills with parallel-jaw grippers. After we train the lifting policy, we collect about $250$ ending states for each object respectively from the successful lifting episodes. In every reset during the reorientation policy training, ending states are randomly sampled and used as the initial pose of the fingers and objects. With a good pose initialization, policies are able to learn to reorient objects with high success rates. $\expertrnn$ trained on EGAD dataset gets a success rate more than $80\%$ while $\expertrnn$ trained on YCB dataset gets a success rate greater than $50\%$. More results on the different policies with and without domain randomization are available in \tblref{tbl:down_air_success_rate} in the appendix. This setup is challenging  because if at any time step in an episode the fingers take a bad action, the object will fall. 

\textbf{Improving performance using gravity curriculum}:
Since the difficulty of training the re-orientation policy with the hand facing downward is due to the gravity, we propose to build a gravity curriculum to learn the policy $\expert$. Since $\expert$ already performs very well on EGAD objects, we apply gravity curriculum to train $\expert$ on YCB objects. Our gravity curriculum is constructed as follows: we start the training with $g=\SI{1}{\meter\per\second^2}$, then we gradually decrease $g$ in a step-wise fashion if the evaluation success rate ($w$) is above a threshold value ($\Bar{w}$) until $g=-\SI{9.81}{\meter\per\second^2}$. More details about gravity curriculum are in \secref{subsec:grav_curr}. In \tblref{tbl:down_air_success_rate} (Exp Q and T) in the appendix, we can see that adding gravity curriculum ($g$-curr) significantly boost the success rates on the YCB dataset.

\renewcommand{\arraystretch}{1.1}
\begin{table}
\centering

\caption{Performance of the student policy when the hand faces upward and downward}
\label{tbl:summary_up_down}
\small{
\begin{tabular}{cccc} 
\hline
Dataset & Upward          & Downward (air)  & Downward (air, $g$-curr)  \\ 
\hline
EGAD    & 91.96 $\pm$ 1.5 & 74.10 $\pm$ 2.3 &   $\diagup$ \\
YCB     & 81.04 $\pm$ 0.5 & 45.22 $\pm$ 2.1  & 67.33 $\pm$ 1.9 \\ 
\hline                 
\end{tabular}}
\end{table}

\renewcommand{\arraystretch}{1}

\subsection{Zero-shot policy transfer across datasets}
\label{subsec:zero_shot}
We have shown the testing performance on the same training dataset so far. How would the policies work on a different dataset? To answer this, we test our policies across datasets: policies trained with EGAD objects are now tested with YCB objects and vice versa. We used the RNN policies trained with full-state information and reduced-state information respectively (without domain randomization) and tested them on the other dataset with the hand facing upward and downward. In the case of the hand facing downward, we tested the RNN policy trained with gravity curriculum.  \tblref{tbl:zero_shot} shows that policies still perform well on the untrained dataset.

\begin{table}[!tb]
    \begin{minipage}[t]{.47\linewidth}
    \vspace{-0.5cm}
      \caption{\small{Zero-shot RNN policy transfer success rates (\%) across datasets. \textbf{U.} (\textbf{D.}) means hand faces upward (downward). \textbf{FS} (\textbf{RS}) means using full-state (reduced-state) information. }}
      \label{tbl:zero_shot}
      \centering
      
        \renewcommand{\arraystretch}{1.1}
        \resizebox{\columnwidth}{!}{
        \begin{tabular}{c|ccc} 
\cline{1-3}
\multicolumn{1}{c}{} & EGAD~$\rightarrow$~YCB & YCB~$\rightarrow$~EGAD &                       \\ 
\cline{1-3}
\textbf{U.FS}                  & $68.82\pm1.7$          & $96.41\pm 1.2$         &                       \\
\textbf{U.RS}               & $59.64\pm 1.8$         & $96.38\pm 1.3$         &                       \\
\textbf{D.FS}                & $62.73\pm 2.2$         & $85.45\pm 2.9$         &                       \\
\textbf{D.RS}                 & $55.30\pm 1.3$         & $77.91\pm 2.1$         &                       \\ 
\cline{1-3}
\multicolumn{1}{c}{} &                        &                        &                       \\
\multicolumn{1}{l}{} & \multicolumn{1}{l}{}   & \multicolumn{1}{l}{}   & \multicolumn{1}{l}{} 
\end{tabular}
}
\renewcommand{\arraystretch}{1}

    \end{minipage}%
    \hfill
    \begin{minipage}[t]{.48\linewidth}
    \vspace{-0.5cm}
      \centering
        \caption{\small{Vision policy success rate ($\Bar{\theta}=\SI{0.2}{\radian}, \Bar{d}_C=0.01$)}}
\label{tbl:vision_pol_noise}
        \resizebox{0.95\columnwidth}{!}{
        \renewcommand{\arraystretch}{1.1}
\begin{tabular}{clc} 
\hline
\multicolumn{2}{c}{Object}                                        & Success rate (\%)  \\ 
\hline
\begin{minipage}{.06\textwidth}
\includegraphics[width=\linewidth]{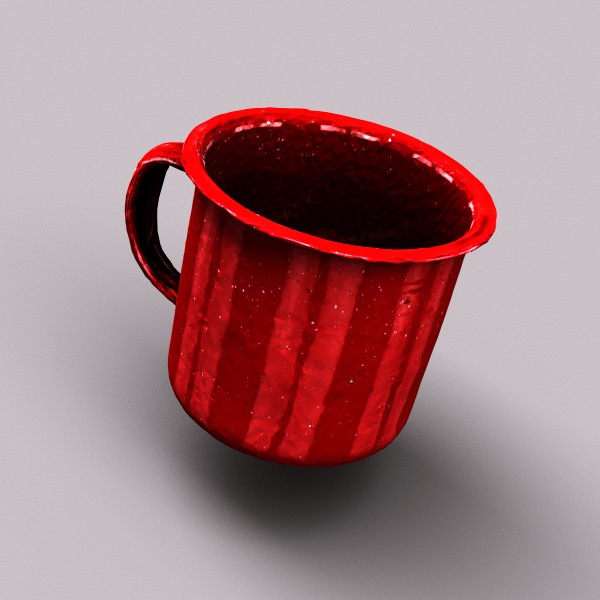}
\end{minipage} & 025\_mug &    $89.67 \pm 1.2$               \\
\begin{minipage}{.06\textwidth}
\includegraphics[width=\linewidth]{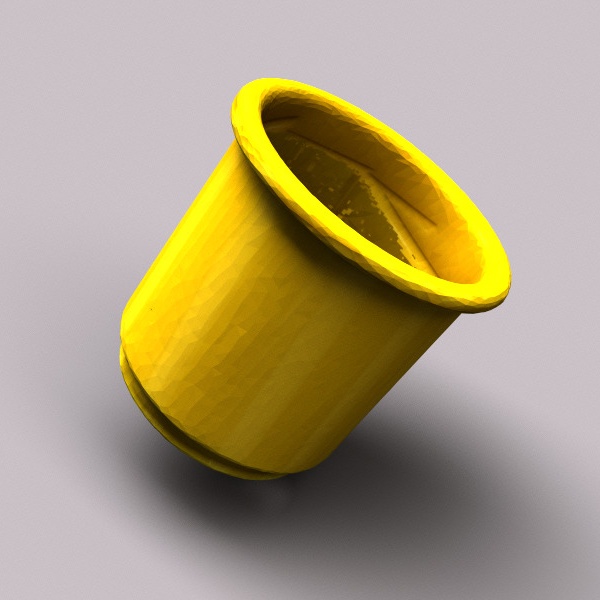}      
\end{minipage}& 065-d\_cups                         &  $68.32\pm 1.9$                       \\
\begin{minipage}{.06\textwidth}
\includegraphics[width=\linewidth]{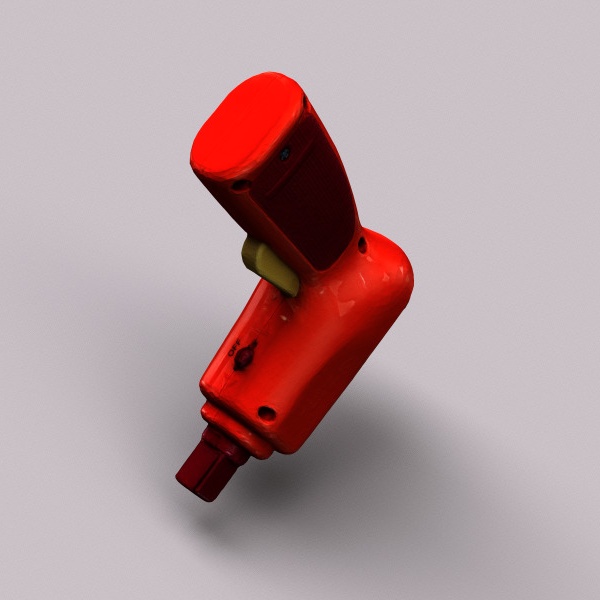}
\end{minipage}&072-b\_toy\_airplane            & $84.52 \pm 1.4$    \\
\begin{minipage}{.06\textwidth}
\includegraphics[width=\linewidth]{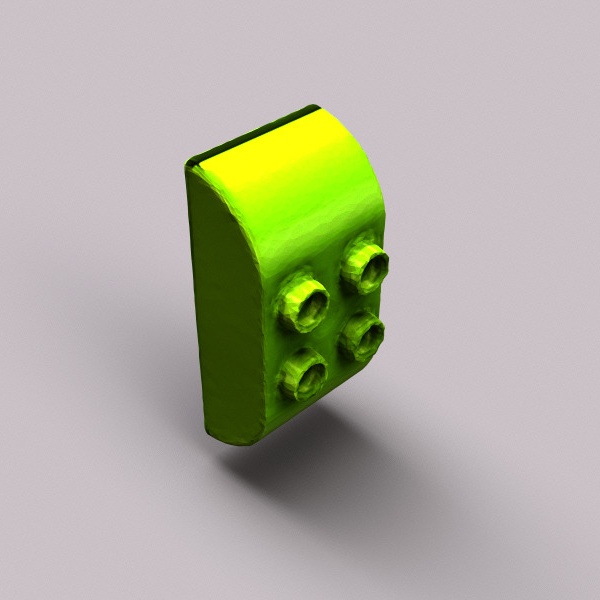}
\end{minipage}& 073-a\_lego\_duplo &      $58.16 \pm 3.1$   \\
 \begin{minipage}{.06\textwidth}
\includegraphics[width=\linewidth]{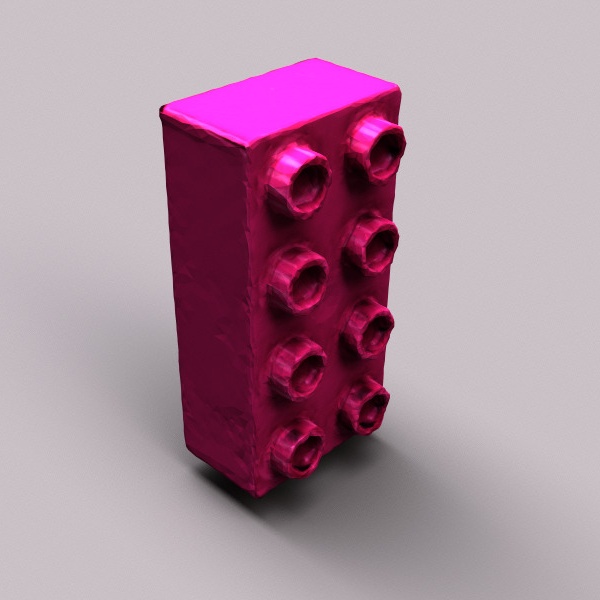}
\end{minipage} &073-c\_lego\_duplo &     $50.21 \pm 3.7$   \\
\begin{minipage}{.06\textwidth}
\includegraphics[width=\linewidth]{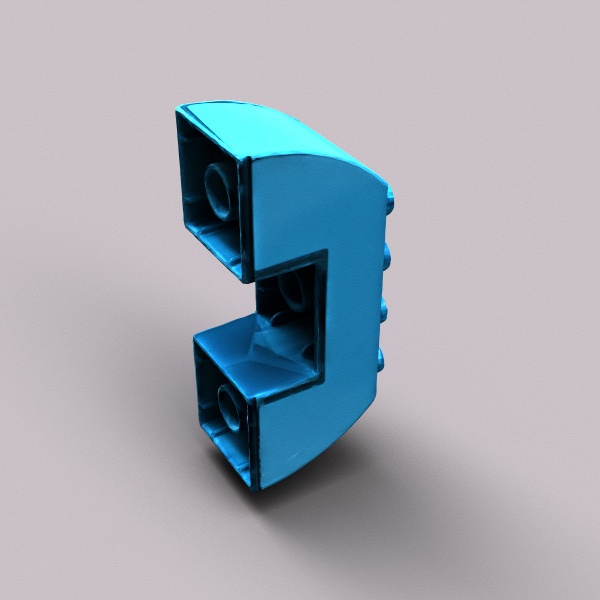}
\end{minipage}&073-e\_lego\_duplo & $66.57\pm 3.1$ \\
\hline
\end{tabular}
\renewcommand{\arraystretch}{1.}
}
    \end{minipage} 
\vspace{-1.cm}
\end{table}

\subsection{Object Reorientation with RGBD sensors}
\label{subsec:vision_exp}
 In this section, we investigate whether we can train a vision policy to reorient objects with the hand facing upward. As vision-based experiments require much more compute resources, we train one vision policy for each object individually on six objects shown in \tblref{tbl:vision_pol_noise}. We leave training a single vision policy for all objects to future work. We use the expert MLP policy trained in \secref{subsec: result_hand_up} to supervise the vision policy. We also performed data augmentation on the point cloud input to the policy network at each time step in both training and testing. The data augmentation includes the random translation of the point cloud, random noise on the point positions, random dropout on the points, and random variations on the point color. More details about the data augmentation are in \secref{app_subsec:vision_noise}. We can see from \tblref{tbl:vision_pol_noise} that reorienting the non-symmetric objects including the toy and the mug has high success rates (greater than $80\%$). While training the policy for symmetric objects is much harder, \tblref{tbl:vision_pol_noise} shows that using $d_C$ as an episode termination criterion enables the policies to achieve a success rate greater than $50\%$.

\section{Related Work}
\label{sec:related}

Dexterous manipulation has been studied for decades, dating back to \citep{salisbury1982articulated, mason1989robot}. In contrast to parallel-jaw grasping, pushing, pivoting~\citep{karayiannidis2016adaptive}, or pick-and-place, dexterous manipulation typically involves continuously controlling force to the object through the fingertips of a robotic hand~\citep{dafle2014extrinsic}. Some prior works used analytical kinematics and dynamics models of the hand and object, and used trajectory optimization to output control policies~\citep{mordatch2012contact, bai2014dexterous, sundaralingam2019relaxed} or employed kinodynamic planning to find a feasible motion plan~\citep{rus1999hand}. However, due to the large number of active contacts on the hand and the objects, model simplifications such as simple finger and object geometries are usually necessary to make the optimization or planning tractable. \citet{sundaralingam2019relaxed} moved objects in hand but assumes that there is no contact breaking or making between the fingers and the object. \citet{furukawa2006dynamic} achieved a high-speed dynamic regrasping motion on a cylinder using a high-speed robotic hand and a high-speed vision system. Prior works have also explored the use of a vision system for manipulating an object to track a planned path~\citep{calli2018path}, detecting manipulation modes~\citep{calli2018learning}, precision manipulation~\citep{calli2017vision} with a limited number of objects with simple shapes using a two-fingered gripper. Recent works have explored the application of reinforcement learning to dexterous manipulation. Model-based RL works learned a linear ~\citep{kumar2016optimal, kumar2016learning} or deep neural network~\citep{nagabandi2020deep} dynamics model from the rollout data, and used online optimal control to rotate a pen or Baoding balls on a Shadow hand. However, when the system is unstable, collecting informative trajectories for training a good dynamics model that generalizes to different objects remains challenging. Another line of works uses model-free RL algorithms to learn a dexterous manipulation policy. For example, \citet{andrychowicz2020learning} and \citet{akkaya2019solving} learned a controller to reorient a block or a Rubik's cube. \citet{van2015learning} learned the tactile informed policy via RL for a three-finger manipulator to move an object on the table. To reduce the sample complexity of model-free learning, \citep{radosavovic2020state, zhu2019dexterous, rajeswaran2017learning, jeong2020learning, gupta2016learning} combined imitation learning with RL to learn to reorient a pen, open a door, assemble LEGO blocks, etc. However, collecting expert demonstration data from humans is expensive, time-consuming, and even incredibly difficult for contact-rick tasks~\citep{rajeswaran2017learning}. Our method belongs to the category of model-free learning. We use the teacher-student learning paradigm to speed up the deployment policy learning. Our learned policies also generalize to new shapes and show strong zero-shot transfer performance. To the best of our knowledge, our system is the first work that demonstrates the capabilities of reorienting a diverse set of objects that have complex geometries with both the hand facing upward and downward. A recent work~\citep{huang2021geometry} (after our CoRL submission) learns a shape-conditioned policy to reorient objects around $z$-axis with an upward-facing hand. Our work tackles more general tasks (more diverse objects, any goal orientation in $SO(3)$, hand faces upward and downward) and shows that even without knowing any object shape information, the policies can get surprisingly high success rates in these tasks.

\section{Discussion and Conclusion}

Our results show that model-free RL with simple deep learning architectures can be used to train policies to reorient a large set of geometrically diverse objects. Further, for learning with the hand facing downwards, we found that a good pose initialization obtained from a lifting policy was necessary, and the gravity curriculum substantially improved performance. The agent also learns to use an external surface (i.e., the table). The most surprising observation is that information about shape is not required despite the fact that we train a single policy to manipulate multiple objects. Perhaps in hindsight, it is not as surprising -- after all, humans can close their eyes and easily manipulate novel objects into a specific orientation. %
Our work can serve a strong baseline for future in-hand object reorientation works that incorporate object shape in the observation space.%

While we only present results in simulation, we also provide evidence that our policies can be transferred to the real world. The experiments with domain randomization indicate that learned policies can work with noisy inputs. Analysis of peak torques during manipulation  (see \figref{fig:torque} in the appendix) also reveals that generated torque commands are feasible to actuate on an actual robotic hand.

Finally,  \figref{fig:success_rate_ycb_up} and \figref{fig:success_rate_ycb_down} in the appendix show that the success rate varies substantially with object shape. This suggests that in the future, a training curriculum based on object shapes can improve performance. Another future work is to directly train one vision policy for a diverse set of objects. A major bottleneck in vision-based experiments is the demand for much larger GPU memory. Learning visual representations of point cloud data that can ease the computational bottleneck is, therefore, an important avenue for future research.

\acknowledgments{
We thank the anonymous reviewers for their helpful comments in revising the paper. We thank the members of Improbable AI lab for providing valuable feedback on research idea formulation and manuscript. This research is funded by Toyota Research Institute, Amazon Research Award, and DARPA Machine Common Sense Program. We also acknowledge the MIT SuperCloud and Lincoln Laboratory Supercomputing Center for providing HPC resources that have contributed to the research results reported within this paper.}

\bibliography{ref}   %
\newpage
\begin{appendices}
\setcounter{figure}{0}
\setcounter{table}{0}
\setcounter{footnote}{0}
\renewcommand{\thefigure}{\Alph{section}.\arabic{figure}}
\renewcommand{\thetable}{\Alph{section}.\arabic{table}}

\section{Evidence indicating transfer to real-world}

Due to the scope of the problem and the ongoing pandemic, we limit our experiments to be in simulation. However, we provided evidence indicating that the learned policies can be transferred to the real world in the future in the paper.  We summarize this evidence as follows.

\paragraph{Convex decomposition}
The objects after the convex decomposition still have geometrically different and complex geometries as shown in \figref{fig:convex_decomp}. The objects in the EGAD dataset are 3D printable. The YCB objects are available in the real world.

\paragraph{Action space}
We control the finger joints via relative position control as explained in \secref{sec:learn_teacher}. This suffers less sim-to-real gap compared to using torque control on the joints directly.

\paragraph{Student policies}
We designed two student policies and both of them use the observation data that can be readily acquired from the real world. The first student policy only requires the joint positions and the object pose. Object pose can be obtained using a pose estimation system or a motion capture system in the real world. Our second student policy only require the point cloud of the scene and the joint positions. We can get the point cloud in the real world by using RGBD cameras such as Realsense D415, Azure Kinect, etc.

\paragraph{Domain randomization}
We also trained and tested our policies with domain randomization. We randomized object mass, friction, joint damping, tendon damping, tendon stiffness, etc. \tblref{tbl:dyn_ran_params} lists all the parameters we randomized in our experiments. We also add noise to the state observation and action commands as shown in \tblref{tbl:dyn_ran_params}. For the vision experiments, we also added noise (various ways of data augmentation including point position jittering, color jiterring, dropout, etc.) to the point cloud observation in training and testing as explained in \secref{app_subsec:vision_noise}. 

The results in \tblref{tbl:up_success_rate}, \tblref{tbl:down_air_success_rate}, and \tblref{tbl:vision_pol_noise} show that even after adding randomization/noise, we can still get good success rates with the trained policies. Even though we cannot replicate the true real-world setups in the simulation, our results with domain randomization indicates a high possibility that our policies can be transferred to the real Shadow hand. Prior works~\cite{andrychowicz2020learning} have also shown the domain randomization can effectively reduce the sim-to-real gap.

\paragraph{Torque analysis}
We also conducted torque analysis as shown in \secref{app_subsec:torque_analysis}. We can see that the peak torque values remain in an reasonable and affordable range for the Shadow hand. This indicates that our learned policies are less likely to cause motor overload on the real Shadow hand.

\newpage

\section{Environment Setup}

\begin{figure}[!h]
\centering
    \begin{subfigure}[t]{0.18\textwidth}
        \centering
        \includegraphics[width=\linewidth]{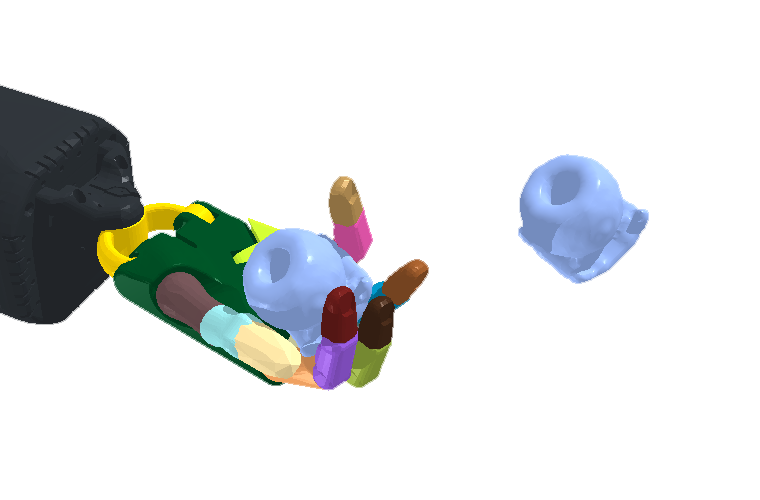}
    \end{subfigure}
    \begin{subfigure}[t]{0.18\textwidth}
        \centering
        \includegraphics[width=\linewidth]{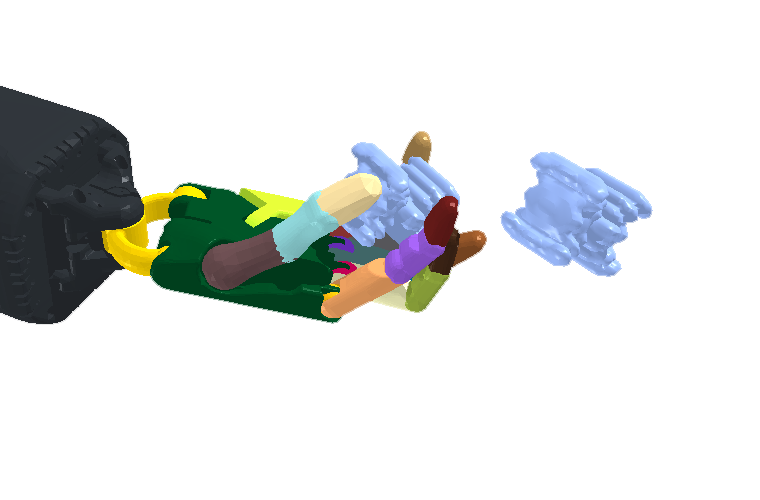} 
    \end{subfigure}
    \begin{subfigure}[t]{0.18\textwidth}
        \centering
        \includegraphics[width=\linewidth]{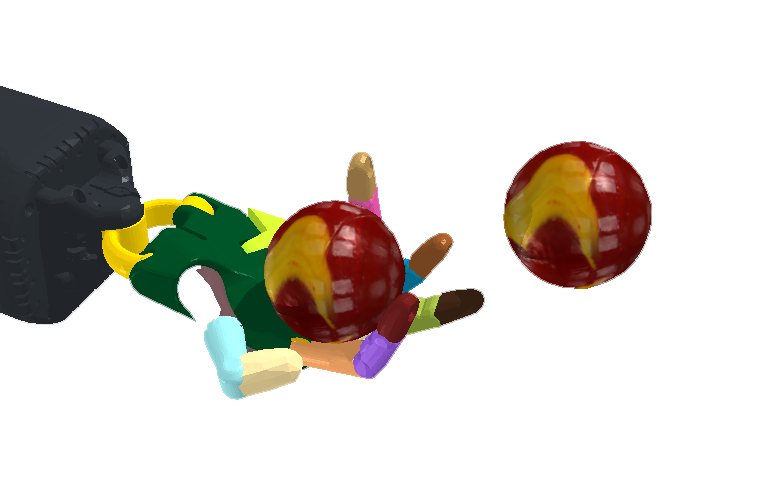} 
    \end{subfigure}
    \begin{subfigure}[t]{0.18\textwidth}
        \centering
        \includegraphics[width=\linewidth]{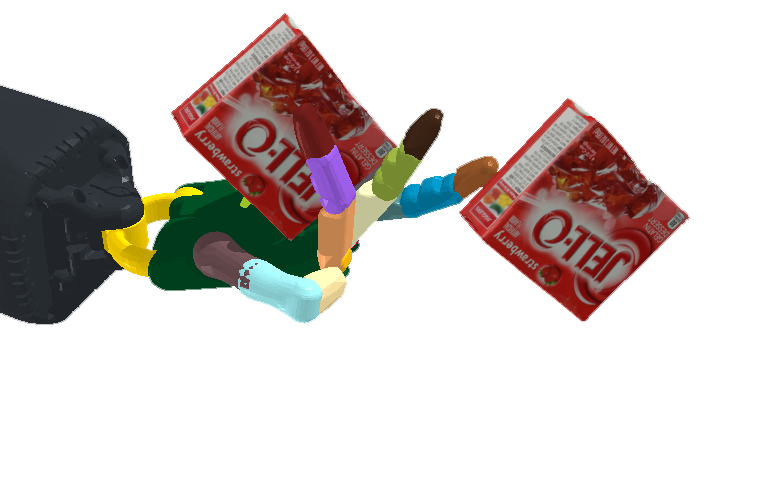} 
    \end{subfigure}
        \begin{subfigure}[t]{0.18\textwidth}
        \centering
        \includegraphics[width=\linewidth]{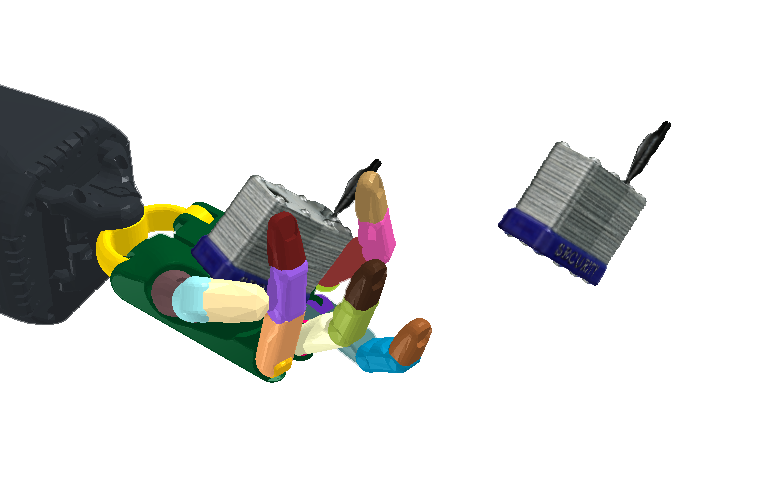} 
    \end{subfigure}%
    \\
    \begin{subfigure}[t]{0.18\textwidth}
        \centering
        \includegraphics[width=\linewidth]{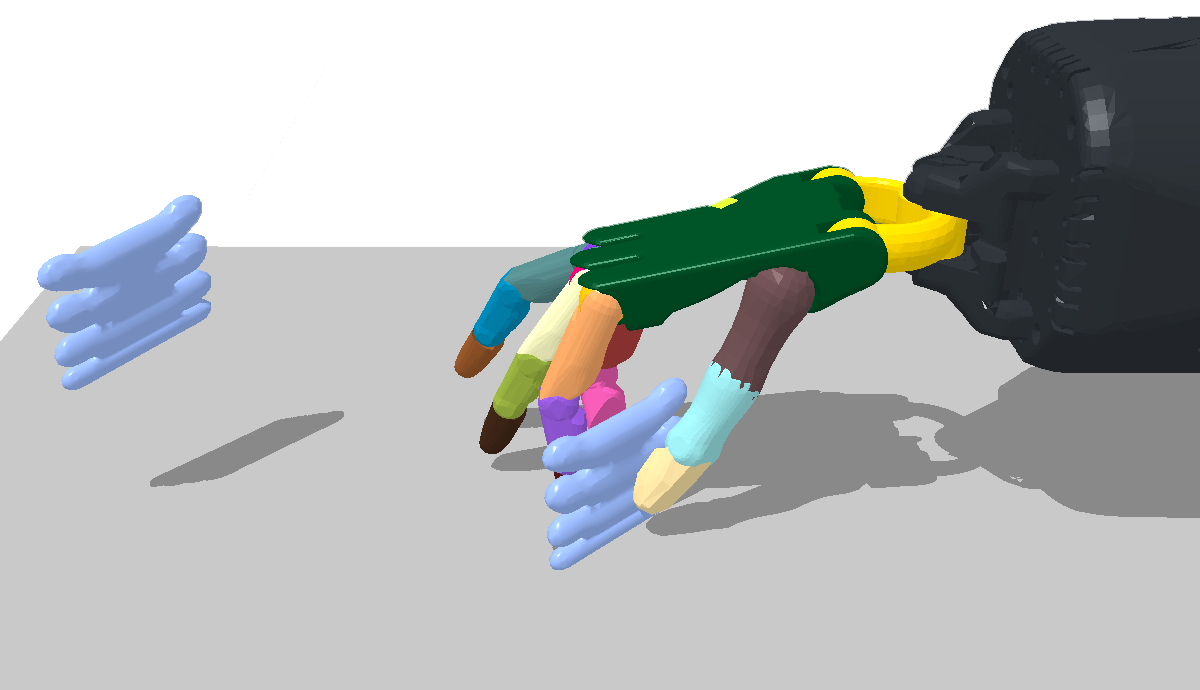} 
    \end{subfigure}
    \begin{subfigure}[t]{0.18\textwidth}
        \centering
        \includegraphics[width=\linewidth]{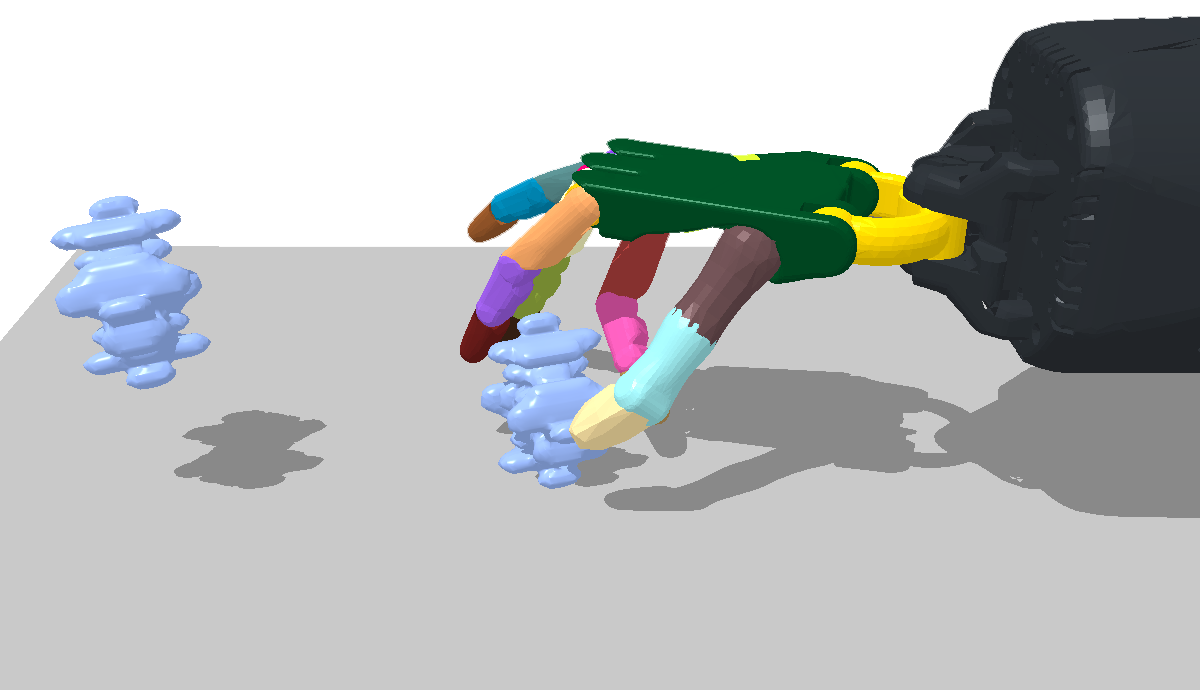} 
    \end{subfigure}
    \begin{subfigure}[t]{0.18\textwidth}
        \centering
        \includegraphics[width=\linewidth]{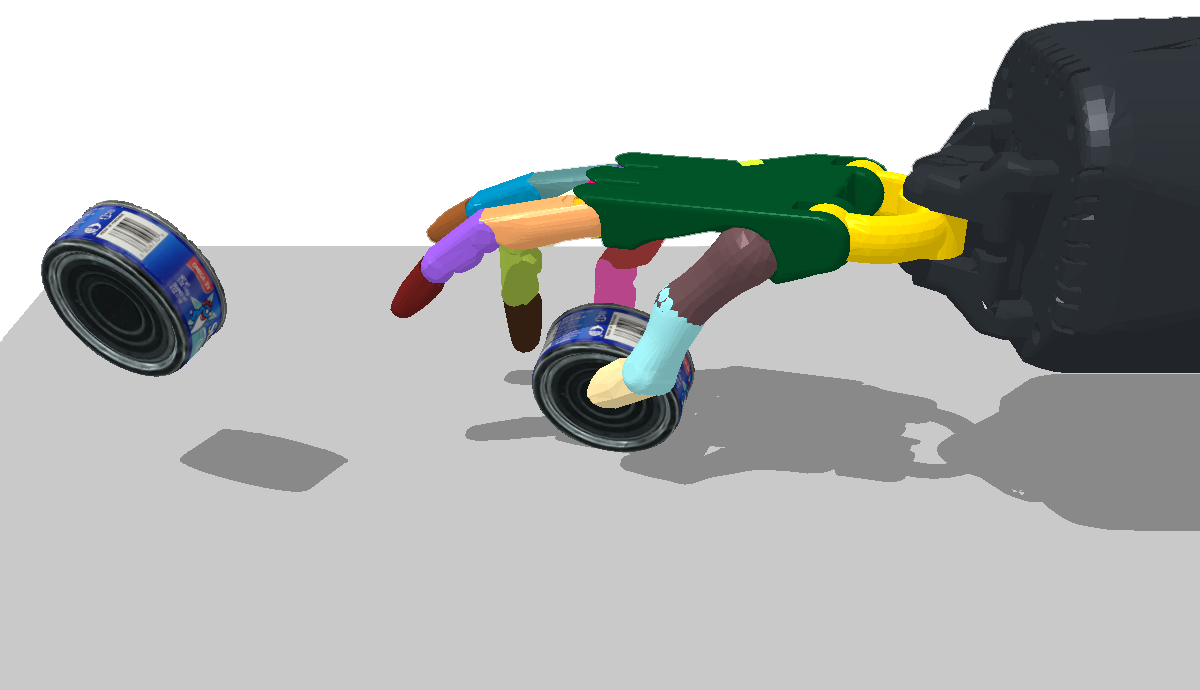} 
    \end{subfigure}
    \begin{subfigure}[t]{0.18\textwidth}
        \centering
        \includegraphics[width=\linewidth]{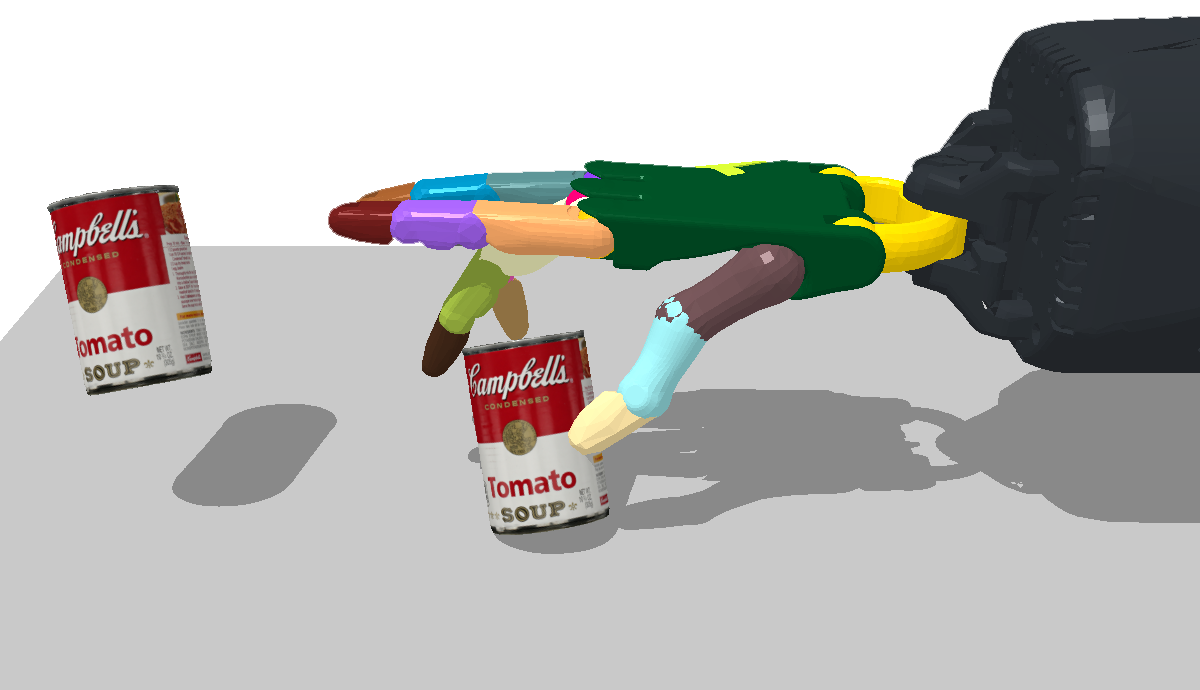} 
    \end{subfigure}
        \begin{subfigure}[t]{0.18\textwidth}
        \centering
        \includegraphics[width=\linewidth]{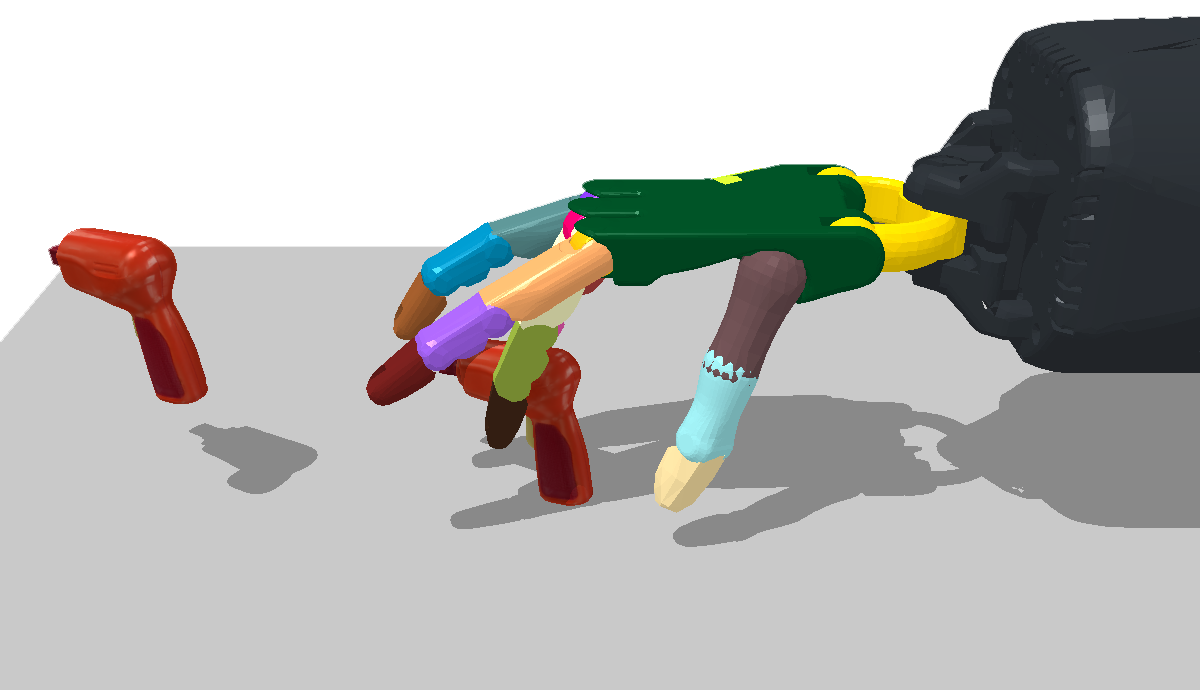} 
    \end{subfigure}%
    \\
     \begin{subfigure}[t]{0.18\textwidth}
        \centering
        \includegraphics[width=\linewidth]{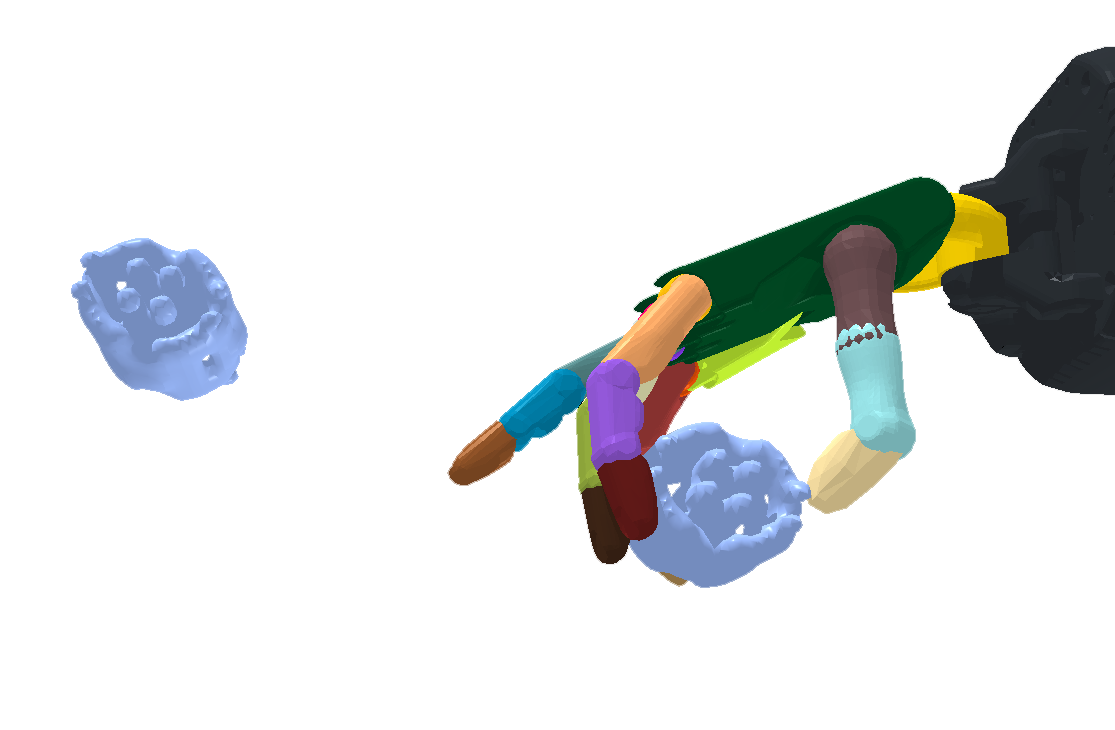} 
    \end{subfigure}
    \begin{subfigure}[t]{0.18\textwidth}
        \centering
        \includegraphics[width=\linewidth]{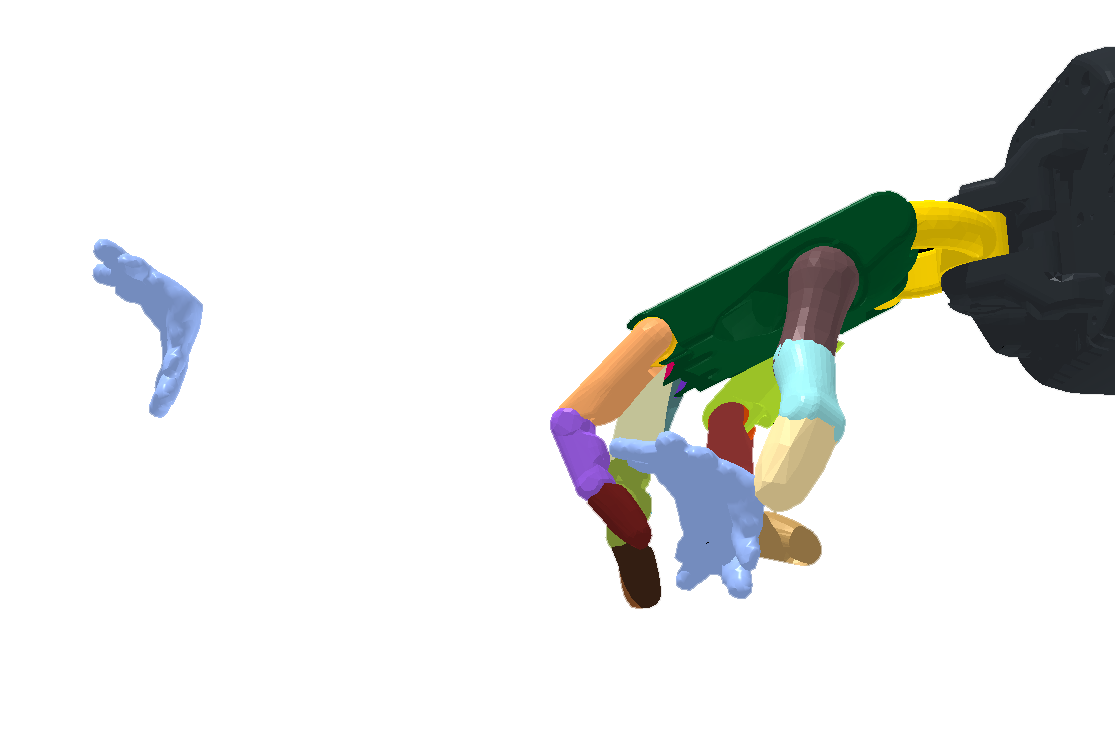} 
    \end{subfigure}
    \begin{subfigure}[t]{0.18\textwidth}
        \centering
        \includegraphics[width=\linewidth]{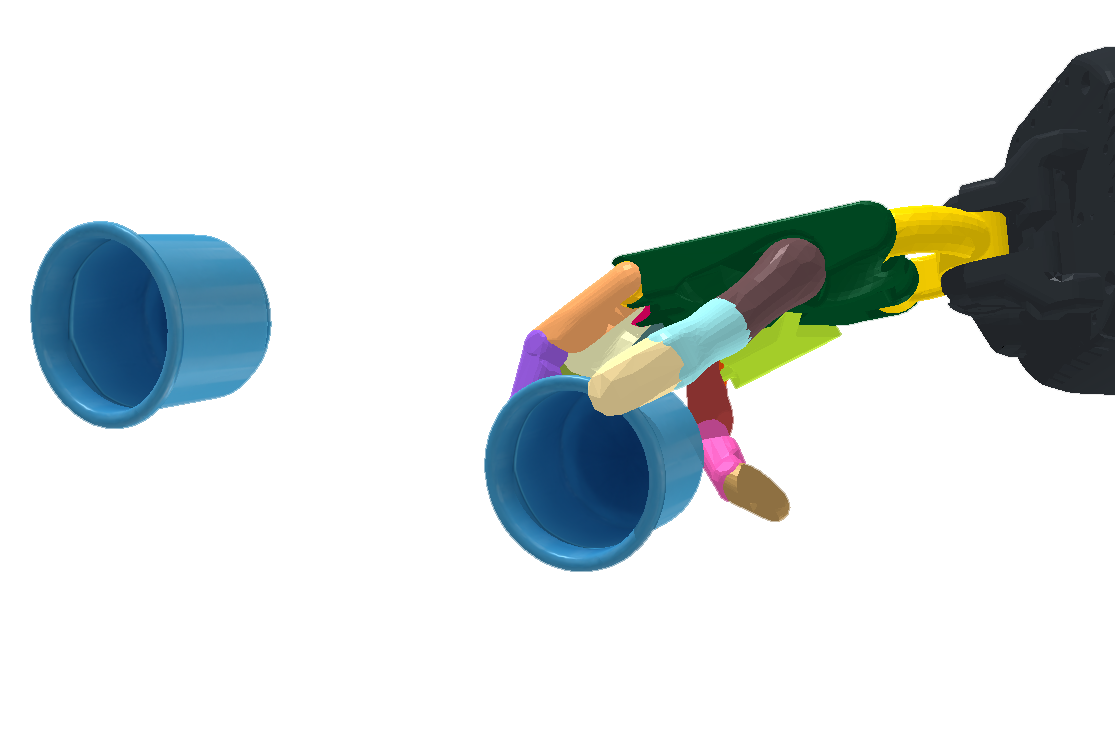} 
    \end{subfigure}
    \begin{subfigure}[t]{0.18\textwidth}
        \centering
        \includegraphics[width=\linewidth]{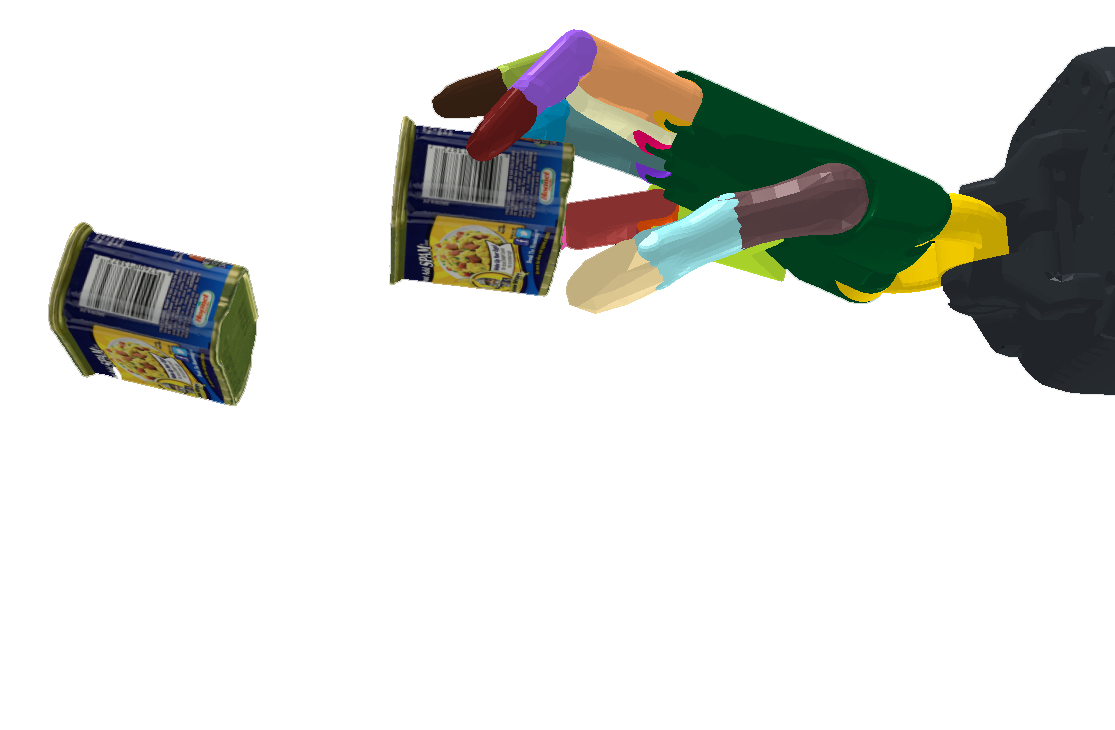} 
    \end{subfigure}
        \begin{subfigure}[t]{0.18\textwidth}
        \centering
        \includegraphics[width=\linewidth]{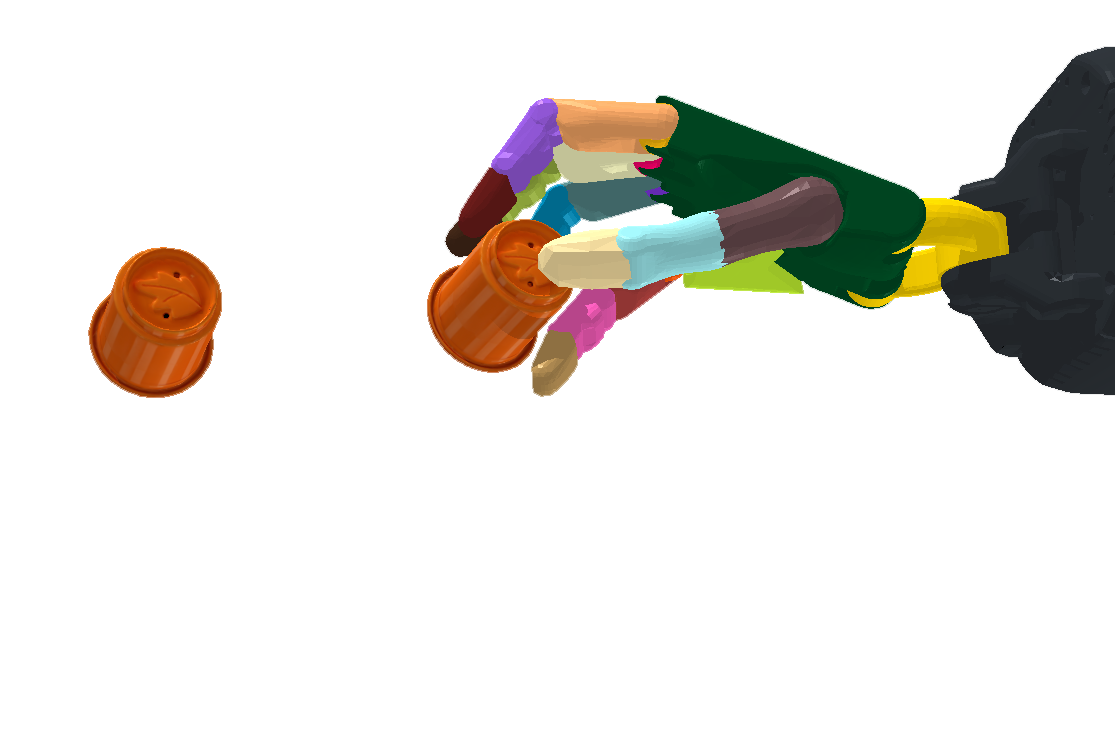} 
    \end{subfigure}
    \caption{We learn policies that can reorient many objects in three scenarios respectively: (1) hand faces upward, (2) hand faces downward with a table below the hand, (3) hand faces downward without any table. The extra object in each figure shows the desired orientation.}
\label{fig:reorient_examples}
\end{figure}

\subsection{State definition}
\label{app_subsec:state_def}

The full state space $\expertspace$ includes joint, fingertip, object and goal information detailed in \tblref{tbl:state_def}. To compute the angle difference between two quaternion orientations $\alpha_1$ and $\alpha_2$, we first compute the difference rotation quaternion: $\beta=\alpha_1\alpha_2^{-1}$. Then the angle difference (distance between two rotations) $\Delta\theta$ is computed as the angle of $\beta$ from the axis-angle representation of $\beta$.

\renewcommand{\arraystretch}{1.3}
\begin{table}[!htb]
\small{
\caption{Full state $s_t^\expsymbol\in\mathbb{R}^{134}$ information. Orientations are in the form of quaternions.}
\label{tbl:state_def}
\centering
\setlength\tabcolsep{3pt}
\begin{tabular}{ll|ll|ll} 
\hline
Parameter                      & Description           & Parameter                                                          & Description                  & Parameter                                                          & Description                                                        \\ 
\hline
$q_t\in\mathbb{R}^{24}$        & joint positions       & $v^f_t\in\mathbb{R}^{15}$ & fingertip linear velocities  & $\alpha^g\in\mathbb{R}^{4}$  & object goal orientation                                             \\
$\dot{q}_t\in\mathbb{R}^{24}$  & joint velocities      & $w^f_t\in\mathbb{R}^{15}$& fingertip angular velocities & $v^o_t\in\mathbb{R}^{3}$ & object linear velocity                                             \\
$p^f_t\in\mathbb{R}^{15}$      & fingertip positions   & $p^o_t\in\mathbb{R}^{3}$ & object position              & $w^o_t\in\mathbb{R}^{3}$ & object angular velocity  \\
$\alpha^f_t\in\mathbb{R}^{20}$ & fingertip orientation & $\alpha^o_t\in\mathbb{R}^{4}$                                      & object orientation           & $\beta_t\in\mathbb{R}^{4}$                                         &
$\alpha^o_t(\alpha^g)^{-1}$
   \\
\hline
\end{tabular}
}
\end{table}
\renewcommand{\arraystretch}{1}

\subsection{Dataset}
\label{app_subsec:dataset}
 We use two object datasets (EGAD and YCB) in our paper. To further increase the diversity of the datasets, we create $5$ variants for each object mesh by randomly scaling the mesh. The scaling ratios are randomly sampled such that the longest side of the objects' bounding boxes $l_{\max}$ lies in $[0.05, 0.08]$m for EGAD objects, and $l_{\max} \in [0.05, 0.12]$m for YCB objects. The mass of each object is randomly sampled from $[0.05, 0.15]$kg. When we randomly scale YCB objects, some objects become very small and/or thin, making the reorientation task even more challenging. In total, we use $11410$ EGAD object meshes and $390$ YCB object meshes for training.

\begin{figure}[!htb]
    \centering
    \begin{subfigure}[t]{0.11\textwidth}
        \centering
        \includegraphics[width=\linewidth]{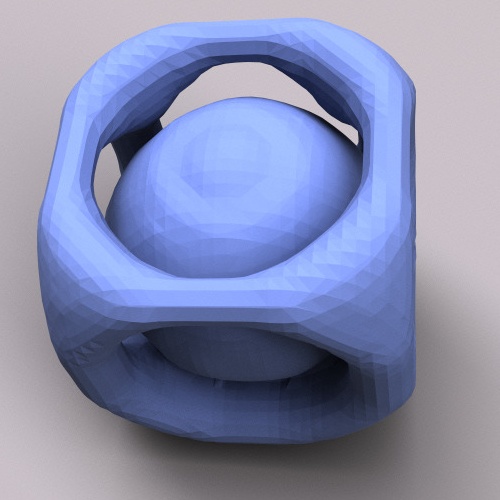} 
    \end{subfigure}
    \begin{subfigure}[t]{0.11\textwidth}
        \centering
        \includegraphics[width=\linewidth]{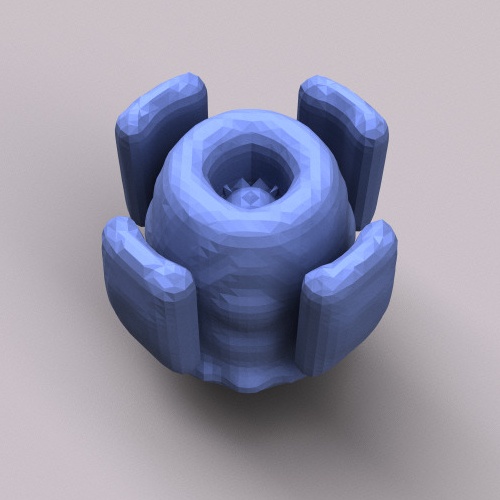} 
    \end{subfigure}
    \begin{subfigure}[t]{0.11\textwidth}
        \centering
        \includegraphics[width=\linewidth]{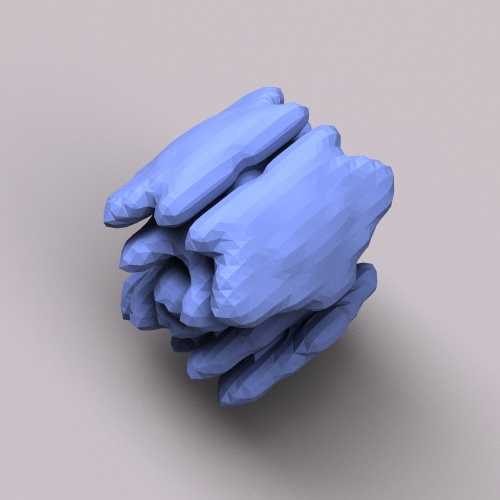} 
    \end{subfigure}
    \begin{subfigure}[t]{0.11\textwidth}
        \centering
        \includegraphics[width=\linewidth]{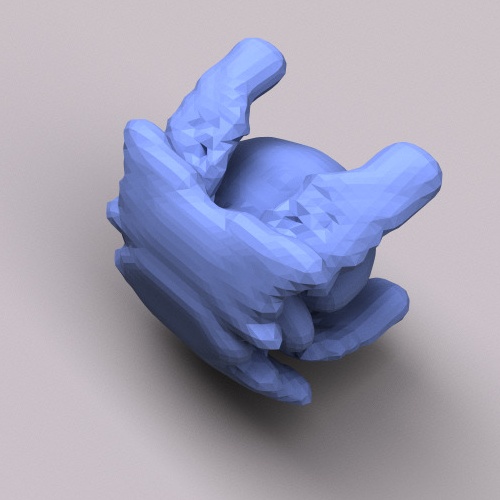} 
    \end{subfigure}
        \begin{subfigure}[t]{0.11\textwidth}
        \centering
        \includegraphics[width=\linewidth]{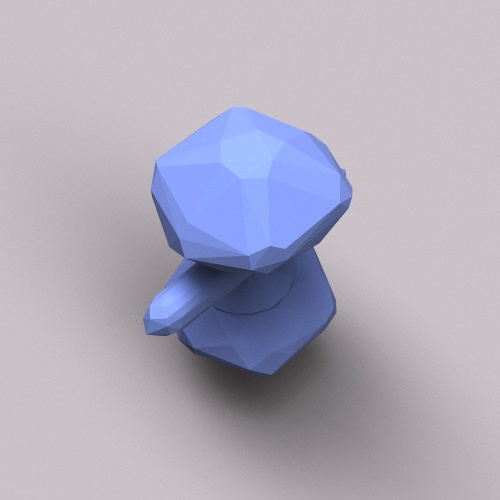} 
    \end{subfigure}
        \begin{subfigure}[t]{0.11\textwidth}
        \centering
        \includegraphics[width=\linewidth]{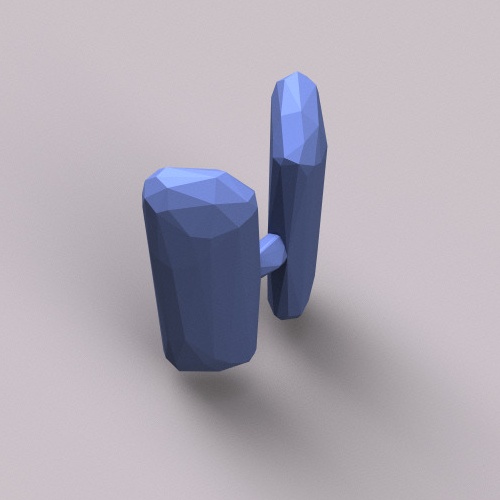} 
    \end{subfigure}
    \begin{subfigure}[t]{0.11\textwidth}
        \centering
        \includegraphics[width=\linewidth]{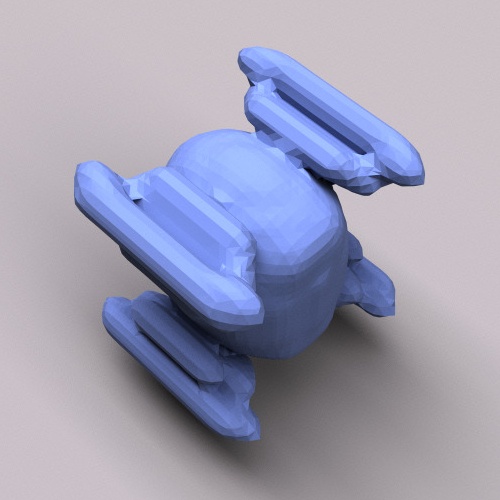} 
    \end{subfigure}
        \begin{subfigure}[t]{0.11\textwidth}
        \centering
        \includegraphics[width=\linewidth]{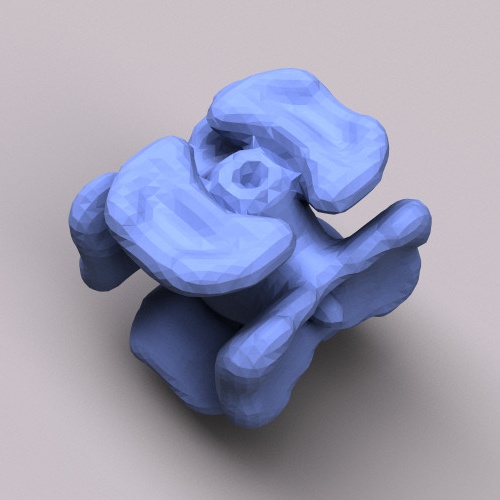} 
    \end{subfigure}
    \\\vspace{0.02cm}
    \begin{subfigure}[t]{0.11\textwidth}
        \centering
        \includegraphics[width=\linewidth]{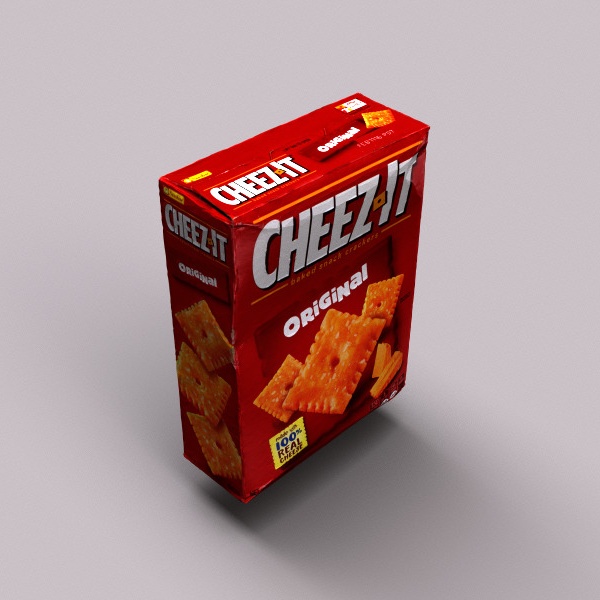} 
    \end{subfigure}
    \begin{subfigure}[t]{0.11\textwidth}
        \centering
        \includegraphics[width=\linewidth]{figures/ycb/025_mug.jpeg} 
    \end{subfigure}
    \begin{subfigure}[t]{0.11\textwidth}
        \centering
        \includegraphics[width=\linewidth]{figures/ycb/065-d_cups.jpeg} 
    \end{subfigure}
    \begin{subfigure}[t]{0.11\textwidth}
        \centering
        \includegraphics[width=\linewidth]{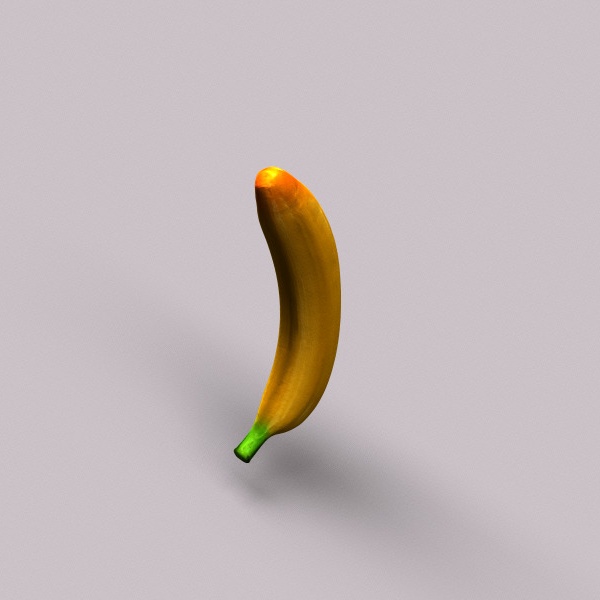} 
    \end{subfigure}
    \begin{subfigure}[t]{0.11\textwidth}
        \centering
        \includegraphics[width=\linewidth]{figures/ycb/072-b_toy_airplane.jpeg} 
    \end{subfigure}
    \begin{subfigure}[t]{0.11\textwidth}
        \centering
        \includegraphics[width=\linewidth]{figures/ycb/073-c_lego_duplo.jpeg} 
    \end{subfigure}
        \begin{subfigure}[t]{0.11\textwidth}
        \centering
        \includegraphics[width=\linewidth]{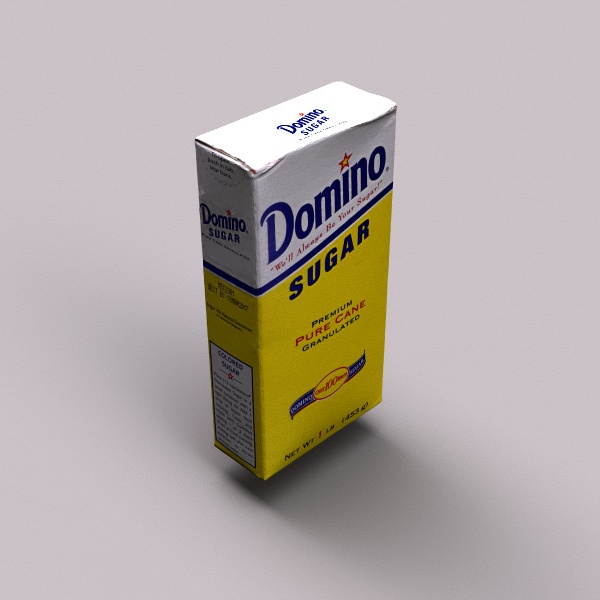} 
    \end{subfigure}
    \begin{subfigure}[t]{0.11\textwidth}
        \centering
        \includegraphics[width=\linewidth]{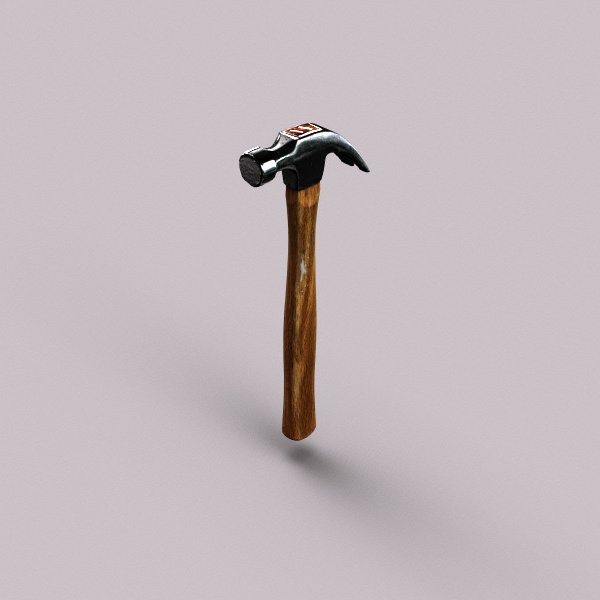} 
    \end{subfigure}%
    \caption{First row: examples of EGAD objects. Second row: examples of YCB objects.}
     \label{fig:egad_ycb_dataset}
\end{figure}

\begin{figure}[!htb]
    \centering
    \begin{subfigure}[t]{0.16\textwidth}
        \centering
        \includegraphics[width=\linewidth]{figures/vhacd/egad/A2_v.jpeg} 
    \end{subfigure}
    \begin{subfigure}[t]{0.16\textwidth}
        \centering
        \includegraphics[width=\linewidth]{figures/vhacd/egad/A4_v.jpeg} 
    \end{subfigure}
    \begin{subfigure}[t]{0.16\textwidth}
        \centering
        \includegraphics[width=\linewidth]{figures/vhacd/egad/B5_v.jpeg} 
    \end{subfigure}
    \begin{subfigure}[t]{0.16\textwidth}
        \centering
        \includegraphics[width=\linewidth]{figures/vhacd/egad/B6_v.jpeg} 
    \end{subfigure}
        \begin{subfigure}[t]{0.16\textwidth}
        \centering
        \includegraphics[width=\linewidth]{figures/vhacd/egad/C4_v.jpeg} 
    \end{subfigure}
    \begin{subfigure}[t]{0.16\textwidth}
        \centering
        \includegraphics[width=\linewidth]{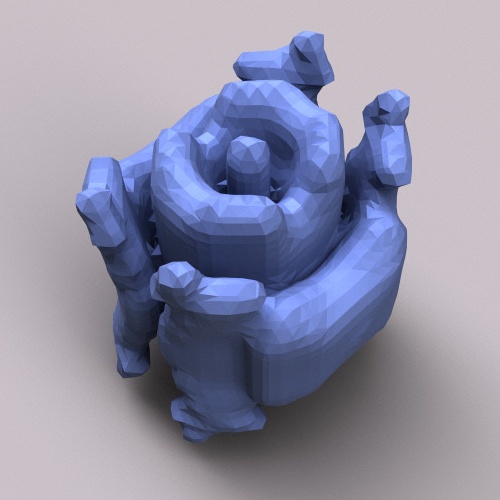} 
    \end{subfigure}%
    \\\vspace{0.02cm}
    \begin{subfigure}[t]{0.16\textwidth}
        \centering
        \includegraphics[width=\linewidth]{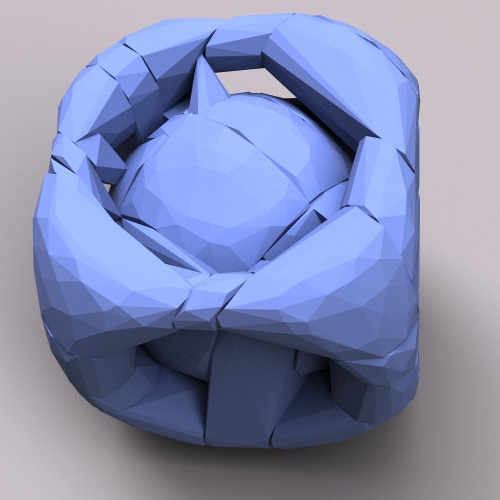} 
    \end{subfigure}
    \begin{subfigure}[t]{0.16\textwidth}
        \centering
        \includegraphics[width=\linewidth]{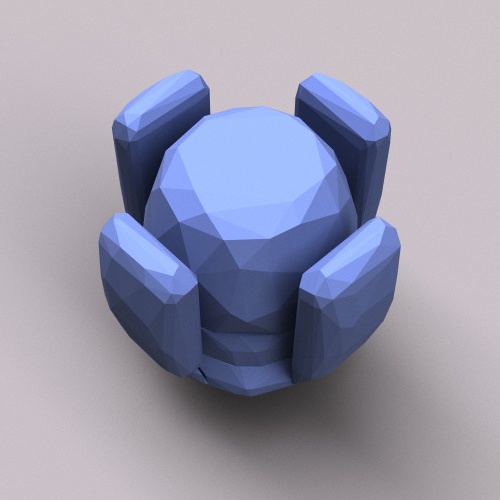} 
    \end{subfigure}
    \begin{subfigure}[t]{0.16\textwidth}
        \centering
        \includegraphics[width=\linewidth]{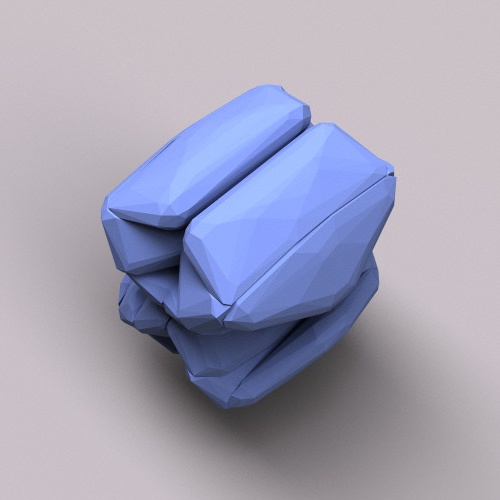} 
    \end{subfigure}
    \begin{subfigure}[t]{0.16\textwidth}
        \centering
        \includegraphics[width=\linewidth]{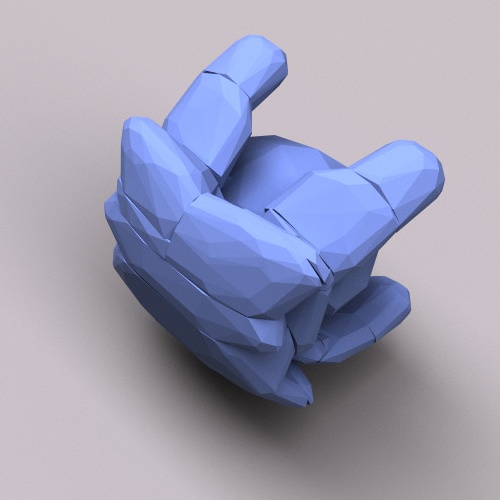} 
    \end{subfigure}
        \begin{subfigure}[t]{0.16\textwidth}
        \centering
        \includegraphics[width=\linewidth]{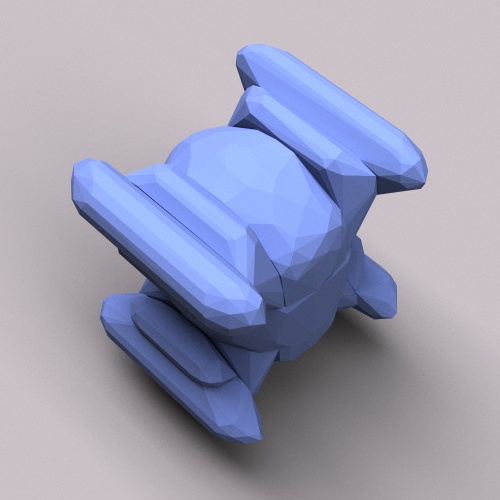} 
    \end{subfigure}
    \begin{subfigure}[t]{0.16\textwidth}
        \centering
        \includegraphics[width=\linewidth]{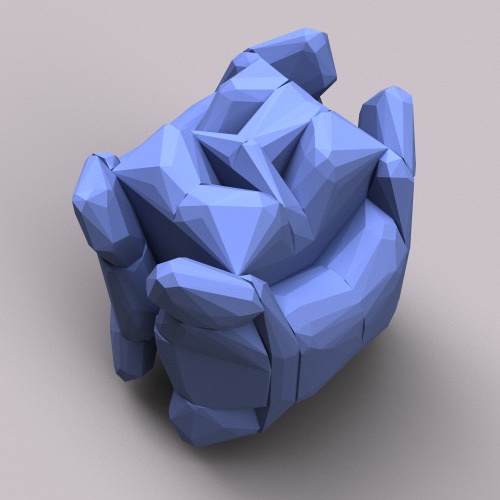} 
    \end{subfigure}%
    \\
    \vspace{0.4cm}
     \begin{subfigure}[t]{0.16\textwidth}
        \centering
        \includegraphics[width=\linewidth]{figures/vhacd/egad/D4_v.jpeg} 
    \end{subfigure}
    \begin{subfigure}[t]{0.16\textwidth}
        \centering
        \includegraphics[width=\linewidth]{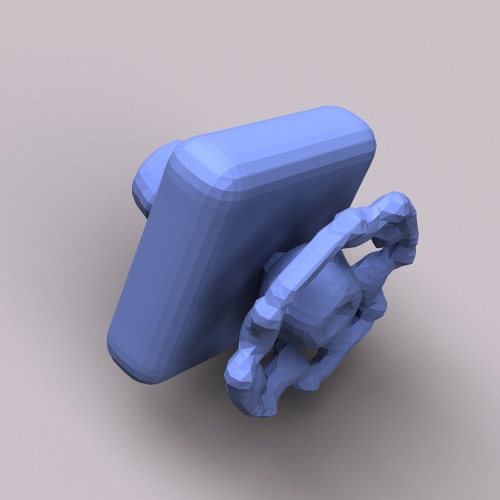} 
    \end{subfigure}
    \begin{subfigure}[t]{0.16\textwidth}
        \centering
        \includegraphics[width=\linewidth]{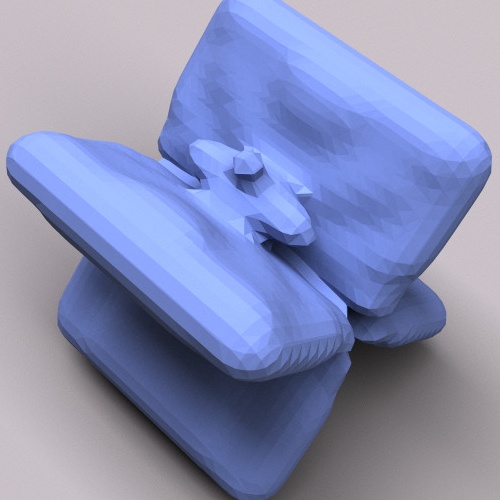} 
    \end{subfigure}
    \begin{subfigure}[t]{0.16\textwidth}
        \centering
        \includegraphics[width=\linewidth]{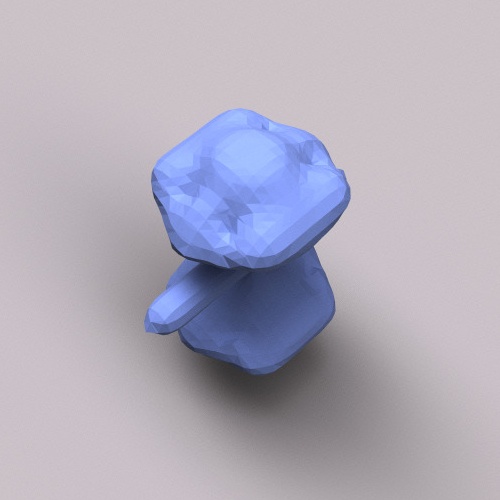} 
    \end{subfigure}
        \begin{subfigure}[t]{0.16\textwidth}
        \centering
        \includegraphics[width=\linewidth]{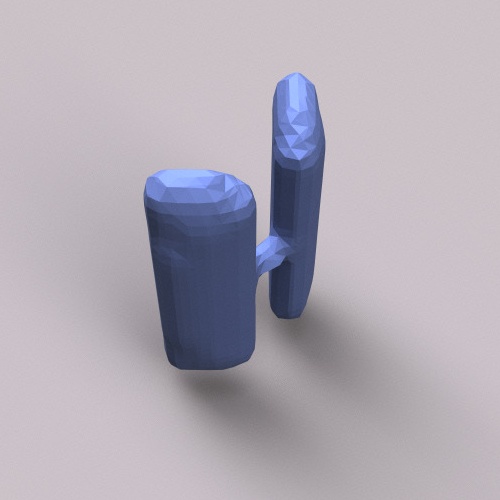} 
    \end{subfigure}
    \begin{subfigure}[t]{0.16\textwidth}
        \centering
        \includegraphics[width=\linewidth]{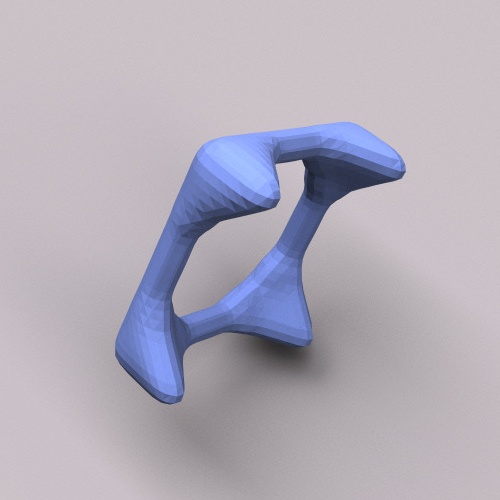} 
    \end{subfigure}%
    \\\vspace{0.02cm}
    \begin{subfigure}[t]{0.16\textwidth}
        \centering
        \includegraphics[width=\linewidth]{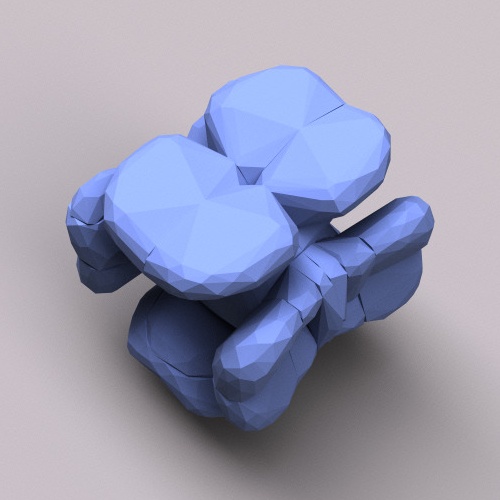} 
    \end{subfigure}
    \begin{subfigure}[t]{0.16\textwidth}
        \centering
        \includegraphics[width=\linewidth]{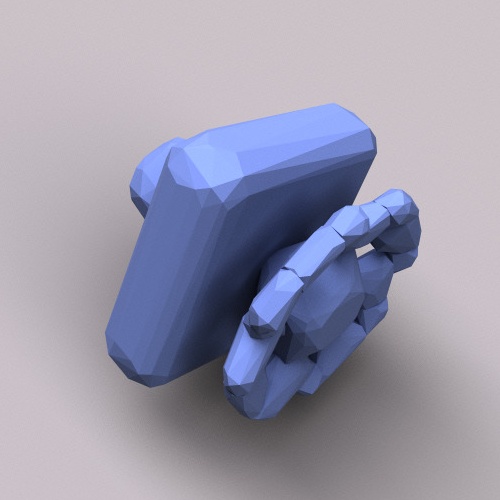} 
    \end{subfigure}
    \begin{subfigure}[t]{0.16\textwidth}
        \centering
        \includegraphics[width=\linewidth]{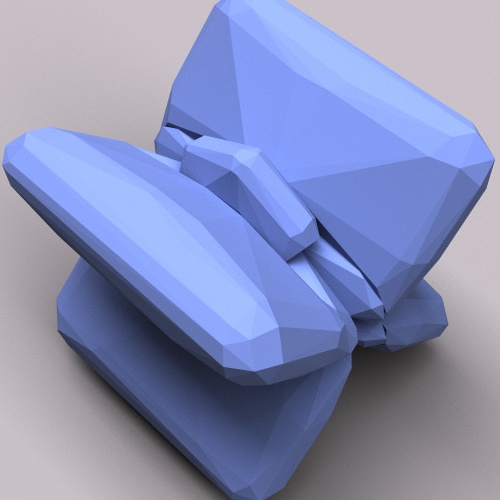} 
    \end{subfigure}
    \begin{subfigure}[t]{0.16\textwidth}
        \centering
        \includegraphics[width=\linewidth]{figures/vhacd/egad/F6_c.jpeg} 
    \end{subfigure}
        \begin{subfigure}[t]{0.16\textwidth}
        \centering
        \includegraphics[width=\linewidth]{figures/vhacd/egad/G2_c.jpeg} 
    \end{subfigure}
    \begin{subfigure}[t]{0.16\textwidth}
        \centering
        \includegraphics[width=\linewidth]{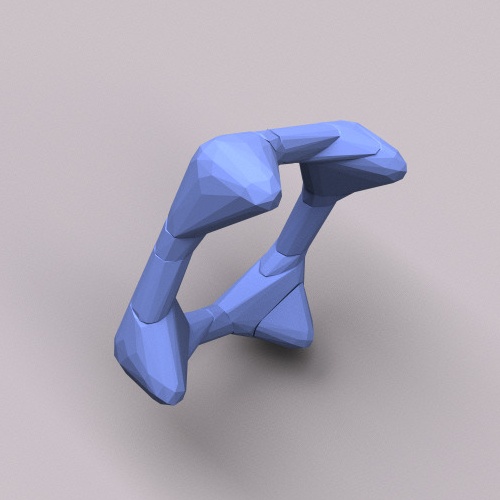} 
    \end{subfigure}%
    \caption{Examples of EGAD objects. The first and third row shows the visual mesh of the objects. The second and fourth row show the corresponding collision mesh (after V-HACD decomposition).}
     \label{fig:convex_decomp}
\end{figure}

\figref{fig:egad_ycb_dataset} shows examples from the EGAD and YCB dataset. We can see that these objects are geometrically different and have complex shapes. We also use V-HACD~\citep{mamou2016volumetric} to perform an approximate convex decomposition on the object meshes for fast collision detection in the simulator. \figref{fig:convex_decomp} shows the object shapes before and after the decomposition. After the decomposition, the objects are still geometrically different. 

\subsection{Camera setup}
\label{subsec:camera_setup}

\begin{figure}[!htb]
    \centering
    \includegraphics[height=0.3\linewidth]{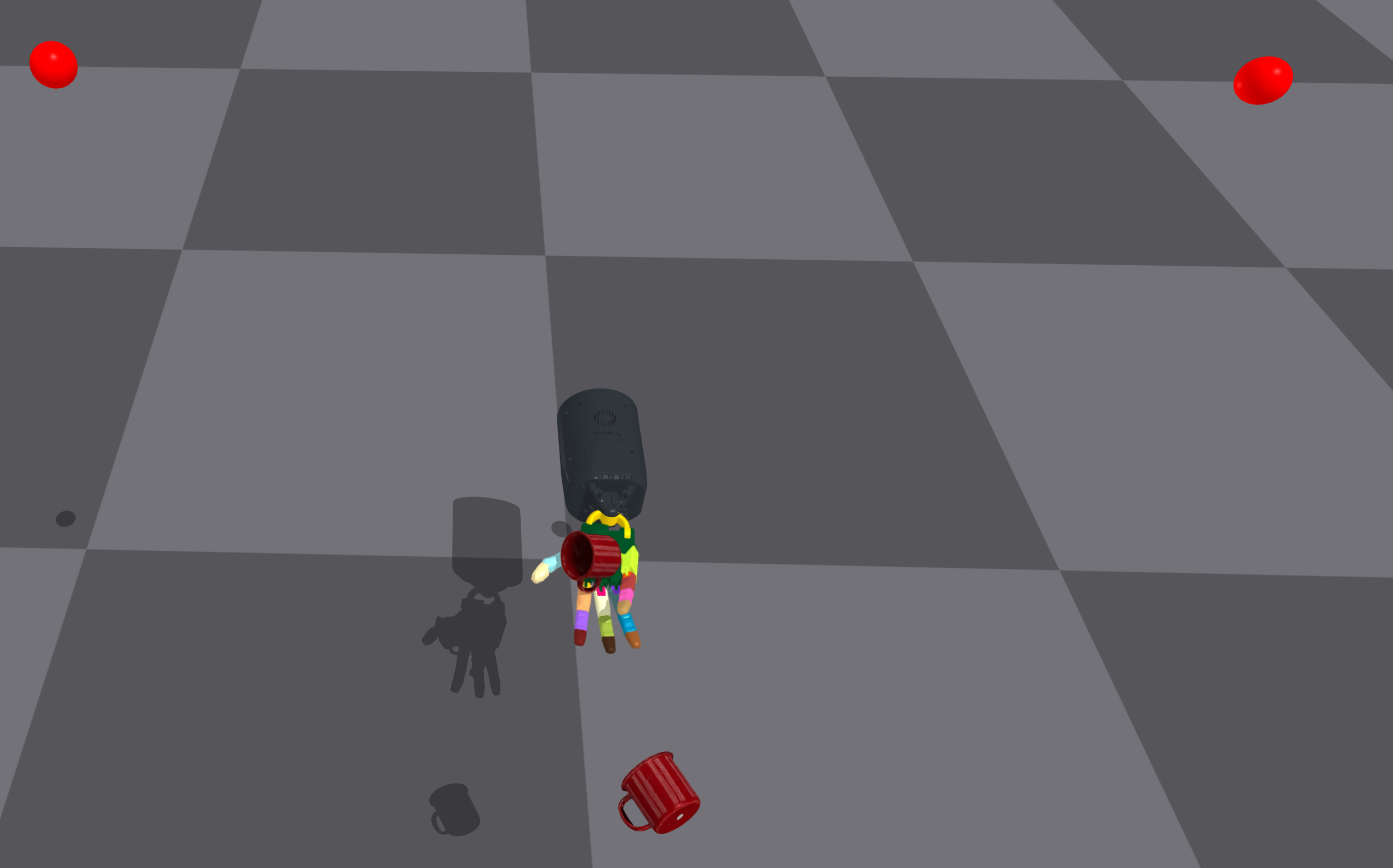}
    \includegraphics[angle=90,height=0.3\linewidth]{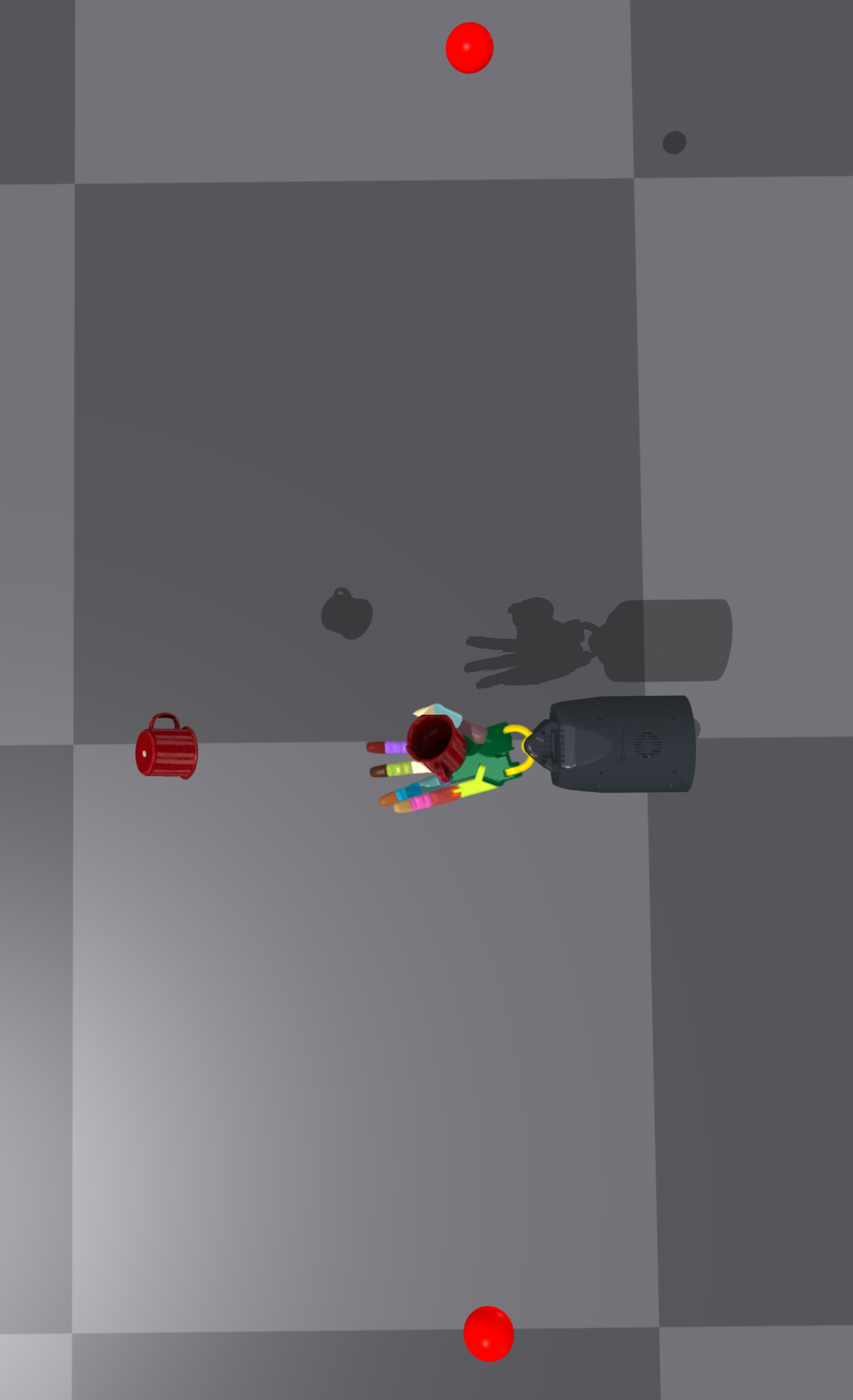}
    \caption{Camera positions} 
    \label{fig:camera_pos}
\end{figure}

We placed two RGBD cameras above the hand, as shown in \figref{fig:camera_pos}. In ISAAC gym, we set the camera pose by setting its position and focus position. The two cameras' positions are shifted from the Shadow hand's base origin by $[-0.6, -0.39, 0.8]$ and $[0.45, -0.39, 0.8]$ respectively. And their focus points are the points shifted from the Shadow hand's base origin by $[-0.08, -0.39, 0.15]$ and $[0.045, -0.39, 0.15]$ respectively.

\section{Experiment Setup}
\label{app_sec:exp_setup}

\subsection{Network architecture}
\label{app_subsec:network_arch}

For the non-vision policies, we experimented with two architectures: The MLP policy $\pi_M$ consists of $3$ hidden layers with $512, 256, 256$ neurons respectively. The RNN policy $\pi_R$ has $3$ hidden layers ($512 - 256 - 256$), followed by a $256$-dim GRU layer and one more $256$-dim hidden layer. We use the exponential linear unit (ELU)~\citep{clevert2015fast} as the activation function.

For our vision policies, we design a sparse convolutional network architecture (\textit{Sparse3D-IMPALA-Net}). As shown in \figref{fig:vision_policy_arch}, the point cloud $W_t$ is processed by a series of sparse CNN residual modules and projected into an embedding vector. $q_t$ and $a_{t-1}$ are concatenated together and projected into an embedding vector via an MLP. Both embedding vectors from $W_t$ and $(q_t, a_{t-1})$ are concatenated and passed through a recurrent network to output the action $a_t$. 

\subsection{Training details}
\label{app_sec:training_details}
All the experiments in the paper were run on at most $2$ GPUs with a $32$GB memory. We use PPO~\citep{schulman2017proximal} to learn $\expert$. \tblref{tbl:hyper_params} lists the hyperparameters for the experiments. We use 40K parallel environments for data collection. We update the policy with the rollout data for $8$ epochs after every $8$ rollout steps for the MLP policies and $50$ rollout steps for the RNN policies. A rollout episode is terminated (reset) if the object is reoriented to the goal orientation successfully, or the object falls, or the maximum episode length is reached. To learn the student policies $\student$, we use Dagger\citep{ross2011reduction}. While Dagger typically keep all the state-action pairs for training the policy, we do Dagger in an online fashion where $\student$ only learns from the latest rollout data. 

For the vision experiments, the number of parallel environments is 360 and we update policy after every $50$ rollout steps from all the parallel environments. The batch size is 30. We sample 15000 points from the reconstructed point cloud of the scene from 2 cameras for the scene point cloud $W_t^s$ and sample 5000 points from the object CAD mesh model for the goal point cloud $W^g$.

We use Horovod~\citep{sergeev2018horovod} for distributed training and Adam~\citep{kingma2014adam} optimizer for neural network optimization.

\textbf{Reward function for reorientation}: For training $\expert$ for the reorientation task, we modified the reward function proposed in \citep{makoviychuk2021isaac} to be:
\begin{align}
\label{eqn:rot_reward}
    r(s_t, a_t) = c_{\theta_1}\frac{1}{|\Delta \theta_t|+\epsilon_\theta}+c_{\theta_2}\mathds{1}(|\Delta \theta_t| < \Bar{\theta}) + c_3\left\Vert a_t\right\Vert_2^2
\end{align}
where $c_{\theta_1}>0$, $c_{\theta_2}>0$ and $c_3<0$ are the coefficients, $\Delta\theta_t$ is the difference between the current object orientation and the target orientation, $\epsilon_\theta$ is a constant, $\mathds{1}$ is an indicator function that identifies whether the object is in the target orientation. The first two reward terms encourage the policy to reorient the object to the desired orientation while the last term suppresses large action commands.

\textbf{Reward function for object lifting}: To train the lifting policy, we use the following reward function: 
\begin{align}
    r(s_t, a_t) &=  c_{h_1}\frac{1}{|\Delta h_t|+\epsilon_h}+c_{h_2}\mathds{1}(|\Delta h_t| < \Bar{h}) + c_3\left\Vert a_t\right\Vert_2^2
\end{align}
where $\Delta h_t=\max(p_t^{b,z} - p_t^{o,z}, 0)$ and $p_t^{b,z}$ is the height ($z$ coordinate) of the Shadow Hand base frame, $p_t^{o,z}$ is the height of the object, $\Bar{h}$ is the threshold of the height difference. The objects have randomly initialized poses and are dropped onto the table.

\textbf{Goal specification for vision policies}: We obtain $W^g$ by sampling $5000$ points from the object's CAD mesh using the Barycentric coordinate, rotating the points by the desired orientation, and translating them so that these points are next to the hand. Note that one can also put the object in the desired orientation right next to the hand in the simulator and render the entire scene altogether to remove the need for CAD models. We use CAD models for $W^g$ just to save the computational cost of rendering another object while we still use RGBD cameras to get $W_t^s$.

\renewcommand{\arraystretch}{1.2}
\begin{table}[!htb]
\centering
\caption{Hyperparameter Setup}
\label{tbl:hyper_params}
\resizebox{\columnwidth}{!}{
\begin{tabular}{cccccc} 
\hline
Hyperparameter       & Value  & Hyperparameter  & Value & Hyperparameter                                                                            & Value  \\ 
\hline
Num. batches         & 5      & Entropy coeff.  & 0.    & Num. pts sampled from $W_t^s$                                                             & 15000  \\
Actor learning rate  & 0.0003 & GAE parameter   & 0.95  & Num. pts sampled from $W^g$                                                               & 5000   \\
Critic learning rate & 0.001  & Discount factor & 0.99  & Num. envs                                                                                 & 40000  \\
Num. epochs          & 8      & Episode length  & 300   & \begin{tabular}[c]{@{}c@{}}Num. rollout steps per \\policy update (MLP/RNN)\end{tabular} & 8/50   \\
Value loss coeff.    & 0.0005 & PPO clip range  & 0.1   &    $c_{\theta_1} $                                                                                      &  1      \\
$c_{\theta_2}$ & 800 & $c_3$ & 0.1 & $\epsilon_\theta$ & 0.1\\
$\Bar{\theta}$ & 0.1\si{\radian} & $c_{h_1}$ & 0.05 & $\epsilon_h$ & 0.02 \\
$\Bar{h}$ & 0.04 & $c_{h_2}$ & 800 & &\\
\hline
\end{tabular}
}
\end{table}
\renewcommand{\arraystretch}{1.}

\begin{table}
\centering
\caption{Mesh Parameters}
\label{tbl:mesh_params}
\begin{tabular}{ll} 
\hline
Parameter                                        & Range             \\ 
\hline
longest side of the bounding box of EGAD objects & {[}0.05, 0.08]m   \\
longest side of the bounding box of YCB objects  & {[}0.05, 0.12]m   \\
mass of each object                              & {[}0.05, 0.15]kg  \\
No. of EGAD meshes                               & 2282              \\
No. of YCB meshes                                & 78                \\
No. of variants per mesh                         & 5                 \\
Voxelization resolution                          & 0.003 m           \\
\hline
\end{tabular}
\end{table}

\subsection{Dynamics randomization}
\tblref{tbl:dyn_ran_params} list all the randomized parameters as well the state observation noise and action command noise.
\renewcommand{\arraystretch}{1.2}
\begin{table}[!htb]
\centering
\caption{Dynamics Randomization and Noise}
\label{tbl:dyn_ran_params}
\resizebox{\columnwidth}{!}{
\begin{tabular}{cccccc} 
\hline
Parameter              & Range                      & Parameter                                                                                 & Range                        & Parameter        & Range                                                                                                                                                                                                                      \\ 
\hline
state observation      & $+\uniform(-0.001, 0.001)$ & action                                                                                    & $+\gaussian(0, 0.01)$        & joint stiffness  & $\times\expunilog(0.75, 1.5)$                                                                                                                                                                                              \\
object mass            & $\times\uniform(0.5, 1.5)$ & joint lower range                & $+\gaussian(0, 0.01)$        & tendon damping   & $\times\expunilog(0.3, 3.0)$                                                                                                                                                                                               \\
object static friction & $\times\uniform(0.7, 1.3)$ & joint upper range & $+\gaussian(0, 0.01)$        & tendon stiffness & $\times\expunilog(0.75, 1.5)$                                                                                                                                                                                              \\
finger static friction & $\times\uniform(0.7, 1.3)$ & joint damping                                                                             & $\times\expunilog(0.3, 3.0)$ &                  &                                                                                                                                                                                                                            \\ 
\hline
\multicolumn{6}{l}{\begin{tabular}[c]{@{}l@{}}$\gaussian(\mu, \sigma)$: Gaussian distribution with mean $\mu$ and standard deviation $\sigma$.\\ $\uniform(a, b)$: uniform distribution between $a$ and $b$. $\expunilog(a, b)=\exp^{\uniform(\log(a), \log(b))}$.\\ $+$: the sampled value is added to the original value of the variable. $\times$: the original value is scaled by the sampled value.\end{tabular}}    \\
\hline
\end{tabular}
}
\end{table}
\renewcommand{\arraystretch}{1}

Comparing the Column 1 and Column 2 in \tblref{tbl:up_success_rate}, we can see that if we directly deploy the policy trained without domain randomization into an environment with different dynamics, the performance drops significantly. If we train policies with domain randomization (Column 3), the policies are more robust and the performance only drops slightly compared to Column 1 in most cases. The exceptions are on C3 and H3. In these two cases, the $\studentmlp$ policies collapsed in training during the policy distillation along with the randomized dynamics.

\subsection{Gravity curriculum}
\label{subsec:grav_curr}

We found building a curriculum on gravity helps improve the policy learning for YCB objects when the hand faces downward. \algoref{alg:grav} illustrates the process of building the gravity curriculum. In our experiments, we only test on the training objects once (one random initial and goal orientation) to get the surrogate average success rate $w$ on all the objects during training. $\Bar{w}=0.8, g_0=\SI{1}{\meter\per\second^2}, \Delta g = \SI{-0.5}{\meter\per\second^2}, K=3, L=20, \Delta T_{\min}=40$.

\begin{algorithm}
\caption{Gravity Curriculum}\label{alg:grav}
\begin{algorithmic}[1]
\State Initialize an empty FIFO queue $Q$ of size $K$, $\Delta T = 0, g=g_{0}$
\For{$i \gets 1$ to $M$}                    
    \State $\tau$ = \texttt{rollout\_policy($\pi_\theta$)} \Comment{get rollout trajectory}
    \State $\pi_\theta$ = \texttt{optimize\_policy($\pi_\theta$, $\tau$)} \Comment{update policy}
    \State $\Delta T = \Delta T + 1$
    \If{$ i \mod L = 0$}
    \State $w$ = \texttt{evaluate\_policy($\pi_\theta$)} \Comment{evaluate the policy, get the success rate $w$}
    \State append $w$ to the queue $Q$
    \If{\texttt{avg}$(Q) > \Bar{w}$ and $\Delta T> \Delta T_{\min}$}
    \State $g=\max(g-\Delta g, -9.81) $
    \State $\Delta T = 0$
    \EndIf
    
    \EndIf
\EndFor
\end{algorithmic}
\end{algorithm}

\section{Supplementary Results}

\subsection{Hand faces upward}
\label{app_subsec:hand_up}

\begin{figure}[!htb]
    \centering
    \includegraphics[width=0.32\linewidth]{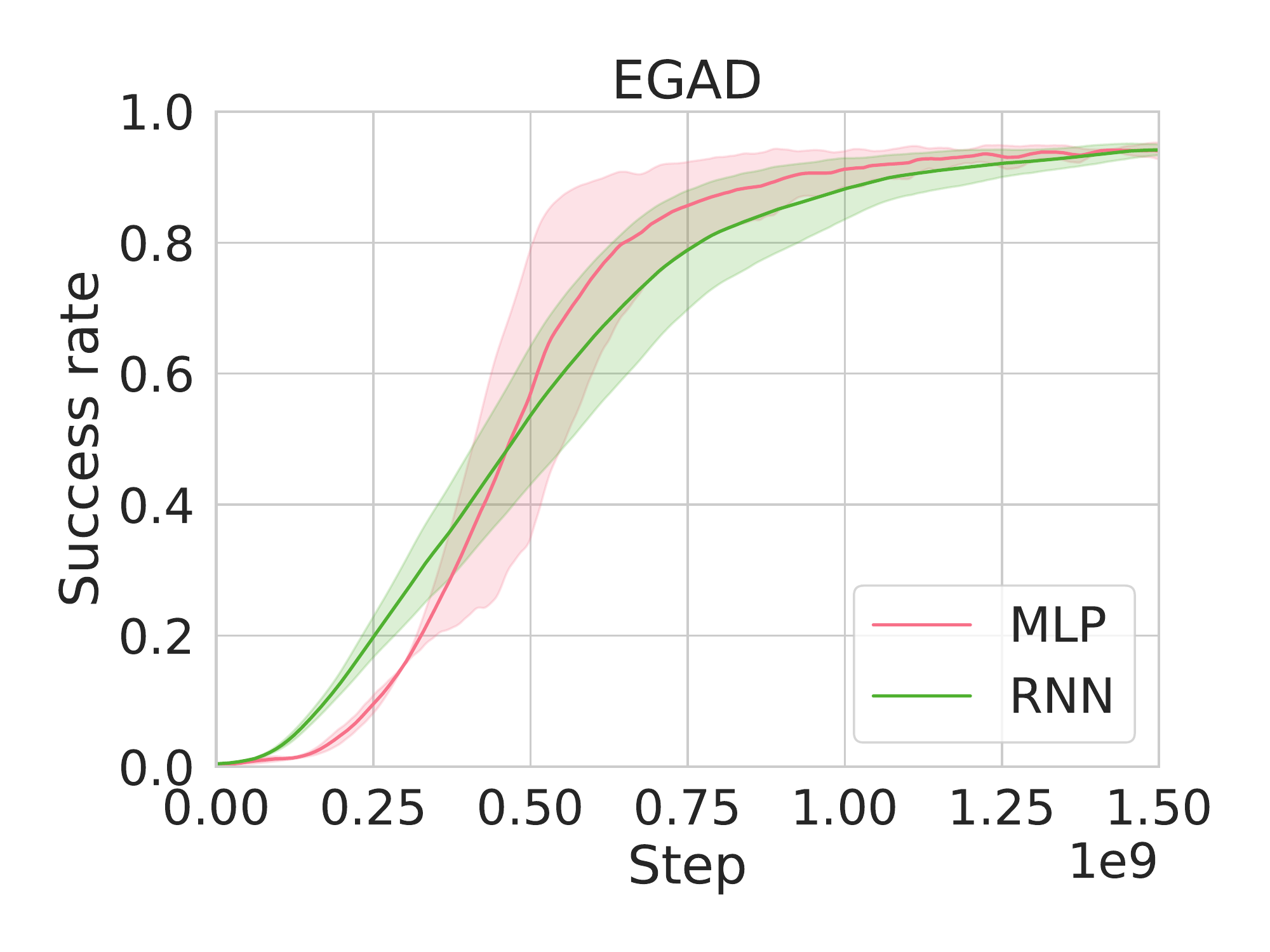}
    \includegraphics[width=0.32\linewidth]{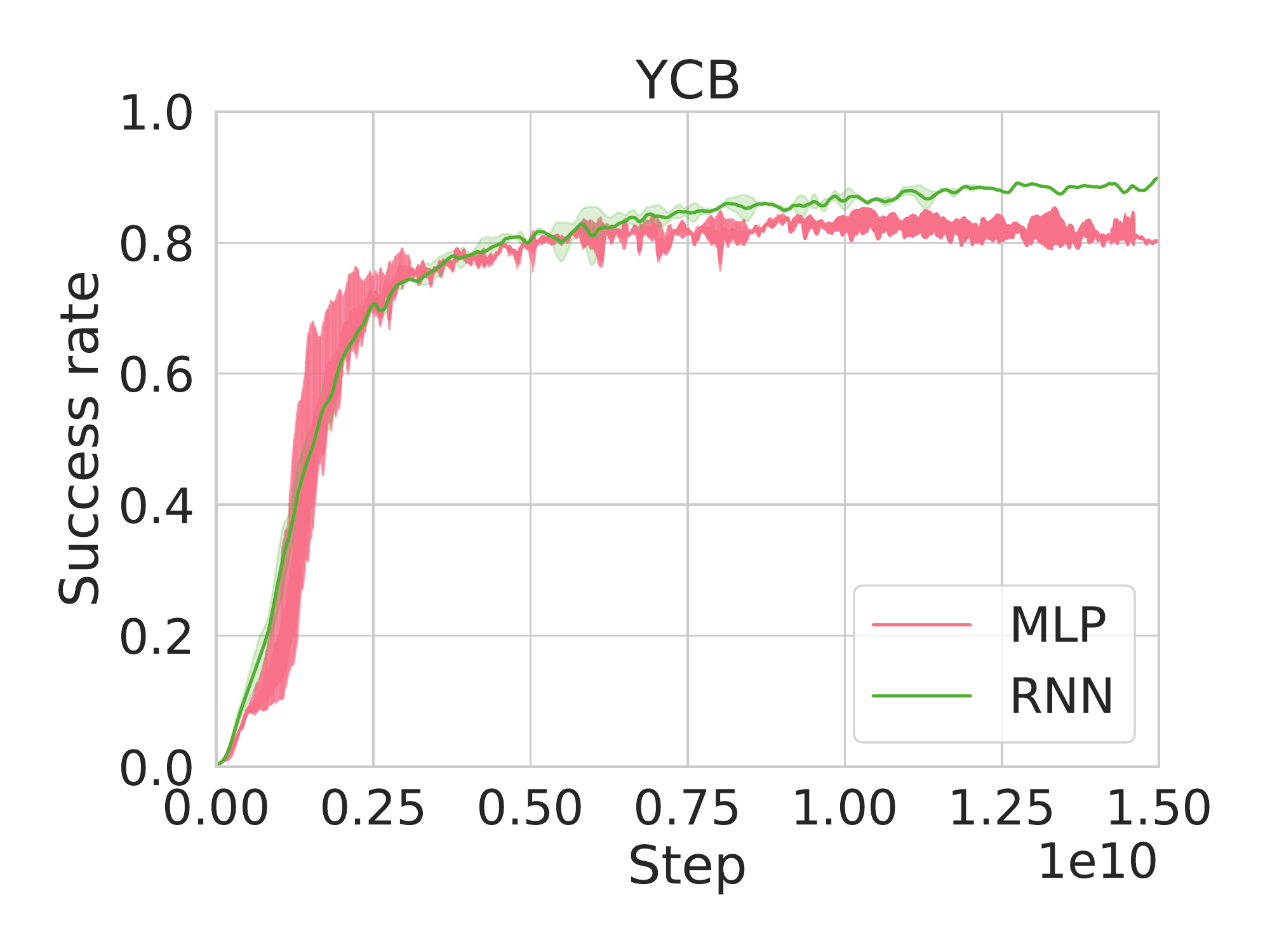}
    \includegraphics[width=0.32\linewidth]{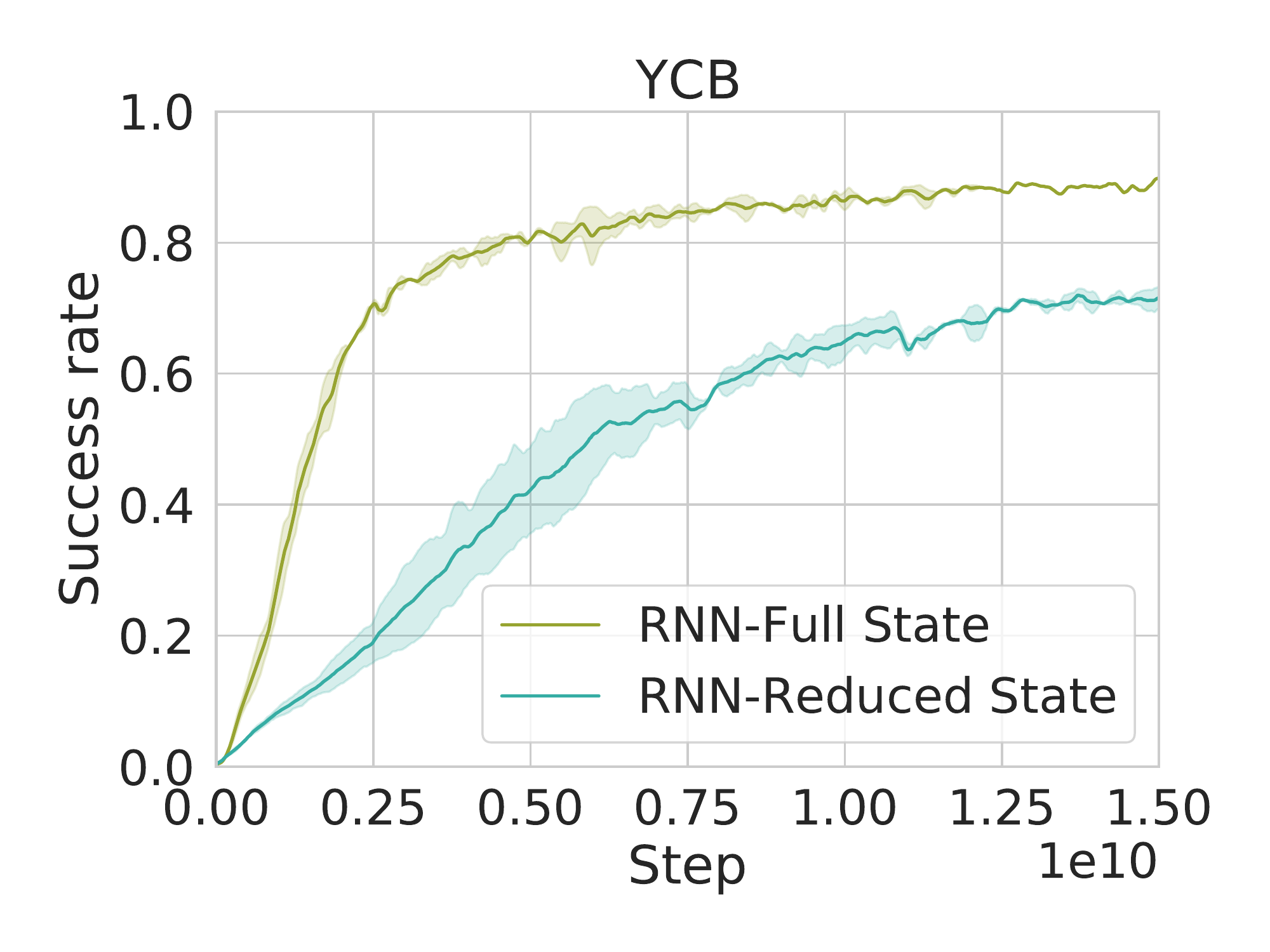}
    \caption{Learning curves of the MLP and RNN policies on the EGAD (\textbf{Left}) and YCB datasets (\textbf{Middle}). The \textbf{Right} plot shows that using the full state information speeds up the policy learning compared to only using the reduced state information.}
    \label{fig:egad_learn_upward}
\end{figure}

\paragraph{Learning curves}
\figref{fig:egad_learn_upward} shows the learning curve of the RNN and MLP policies on the EGAD and YCB datasets. Both policies learn well on the EGAD and YCB datasets. The YCB dataset requires much more environment interactions for the policies to learn. We can also see that using the full-state information can speed up the learning and give a higher final performance. 

\paragraph{Testing performance - Teacher}
The testing results in \tblref{tbl:up_success_rate} show that both the MLP and RNN policies are able to achieve a success rate greater than 90\% on the EGAD dataset (entries A1, B1) and greater than 70\% on the YCB dataset (entries F1, G1) without any explicit knowledge of the object shape.  This result is surprising because intuitively, one would assume that information about the object shape is important for in-hand reorientation.

\paragraph{Testing performance - Student}
We experimented with the following three combinations: (1) distill $\expertmlp$ into $\studentmlp$, (2) distill $\expertmlp$ into $\studentrnn$, (3) distill $\expertrnn$ into $\studentrnn$. The student policy state is $s_t^\stusymbol\in\mathbb{R}^{31}$. In \tblref{tbl:up_success_rate} (entries C1-E1, H1-J1), we can see that $\expertrnn\rightarrow\studentrnn$ gives the highest success rate on $\student$, while $\expertmlp\rightarrow\studentmlp$ leads to much worse performance ($36\%$ drop of success rate in EGAD, and $47\%$ drop in YCB). This shows that $\student$ requires temporal information due to reduced state space. The last two columns in \tblref{tbl:up_success_rate} also show that the policy is more robust to dynamics variations and observation/action noise after being trained with domain randomization.

\renewcommand{\arraystretch}{1.1}
\begin{table}[!tb]
\vspace{-0.5cm}
\centering
\caption{Success rates (\%) of policies tested on different dynamics distribution. $\Bar{\theta}=0.1$\si{\radian}. DR stands for domain randomization and observation/action noise. X$\rightarrow$Y: distill policy X into policy Y.}
\label{tbl:up_success_rate}
\resizebox{0.9\columnwidth}{!}{
\begin{tabular}{cccc|ccc} 
\hline
\multicolumn{4}{c}{}                                                                                                               & 1                                   & 2              & 3                \\ 
\hline
\multirow{2}{*}{Exp. ID} & \multirow{2}{*}{Dataset} & \multirow{2}{*}{State}         & \multicolumn{1}{c}{\multirow{2}{*}{Policy}} & \multicolumn{2}{c}{Train without DR}                 & Train with DR    \\ 
\cline{5-7}
                         &                          &                                & \multicolumn{1}{c}{}                        & \multicolumn{1}{l}{Test without DR} & Test with DR   & Test with DR     \\ 
\hline
A                        & \multirow{5}{*}{EGAD}    & \multirow{2}{*}{Full state}    & MLP                                         & 92.55 $\pm$ 1.3                      & 78.24 $\pm$ 2.4 & \textbf{91.92} $\pm$ \textbf{0.4}   \\
B                        &                          &                                & RNN                                         & \textbf{95.95} $\pm$ \textbf{0.8}                      & \textbf{84.27} $\pm$ \textbf{1.0} & 88.04 $\pm$ 0.6   \\
C                        &                          & \multirow{3}{*}{Reduced state} & MLP$\rightarrow$MLP                        & 55.55 $\pm$ 0.2                      & 25.09 $\pm$ 3.0 & 23.77 $\pm$ 1.8   \\
D                        &                          &                                & MLP$\rightarrow$RNN                       & 85.32 $\pm$ 1.2                     & 68.31 $\pm$ 2.6 & \textbf{81.05} $\pm$ \textbf{1.2}  \\
E                        &                          &                                & RNN$\rightarrow$RNN                        & \textbf{91.96} $\pm$ \textbf{1.5}                     & \textbf{78.30} $\pm$ \textbf{1.2} & 80.29 $\pm$ 0.9   \\ 
\hline
F                        & \multirow{5}{*}{YCB}     & \multirow{2}{*}{Full state}    & MLP                                         & 73.40 $\pm$ 0.2                      & 54.57 $\pm$ 0.6 & 66.00 $\pm$ 2.7   \\
G                        &                          &                                & RNN                                         & \textbf{80.40} $\pm$ \textbf{1.6}                     & \textbf{65.16} $\pm$ \textbf{1.0} & \textbf{72.34} $\pm$ \textbf{0.9}   \\
H                        &                          & \multirow{3}{*}{Reduced state} & MLP$\rightarrow$MLP                        & 34.08 $\pm$ 0.9                      & 12.08 $\pm$ 0.4 & \ \  5.41 $\pm$ 0.3   \\
I                        &                          &                                & MLP$\rightarrow$RNN                         & 69.30 $\pm$ 0.8                      & 47.38 $\pm$ 0.6 & 53.07 $\pm$ 0.9  \\
J                        &                          &                                & RNN$\rightarrow$RNN                         & \textbf{81.04} $\pm$ \textbf{0.5}                      & \textbf{64.93} $\pm$ \textbf{0.2} & \textbf{65.86} $\pm$ \textbf{0.7}  \\
\hline
\end{tabular}
}
\vspace{-0.5cm}
\end{table}

\renewcommand{\arraystretch}{1}

\begin{figure}[!htb]
    \centering
    \begin{subfigure}[t]{0.24\textwidth}
        \centering
        \includegraphics[trim=300 240 80 200, clip, width=\linewidth]{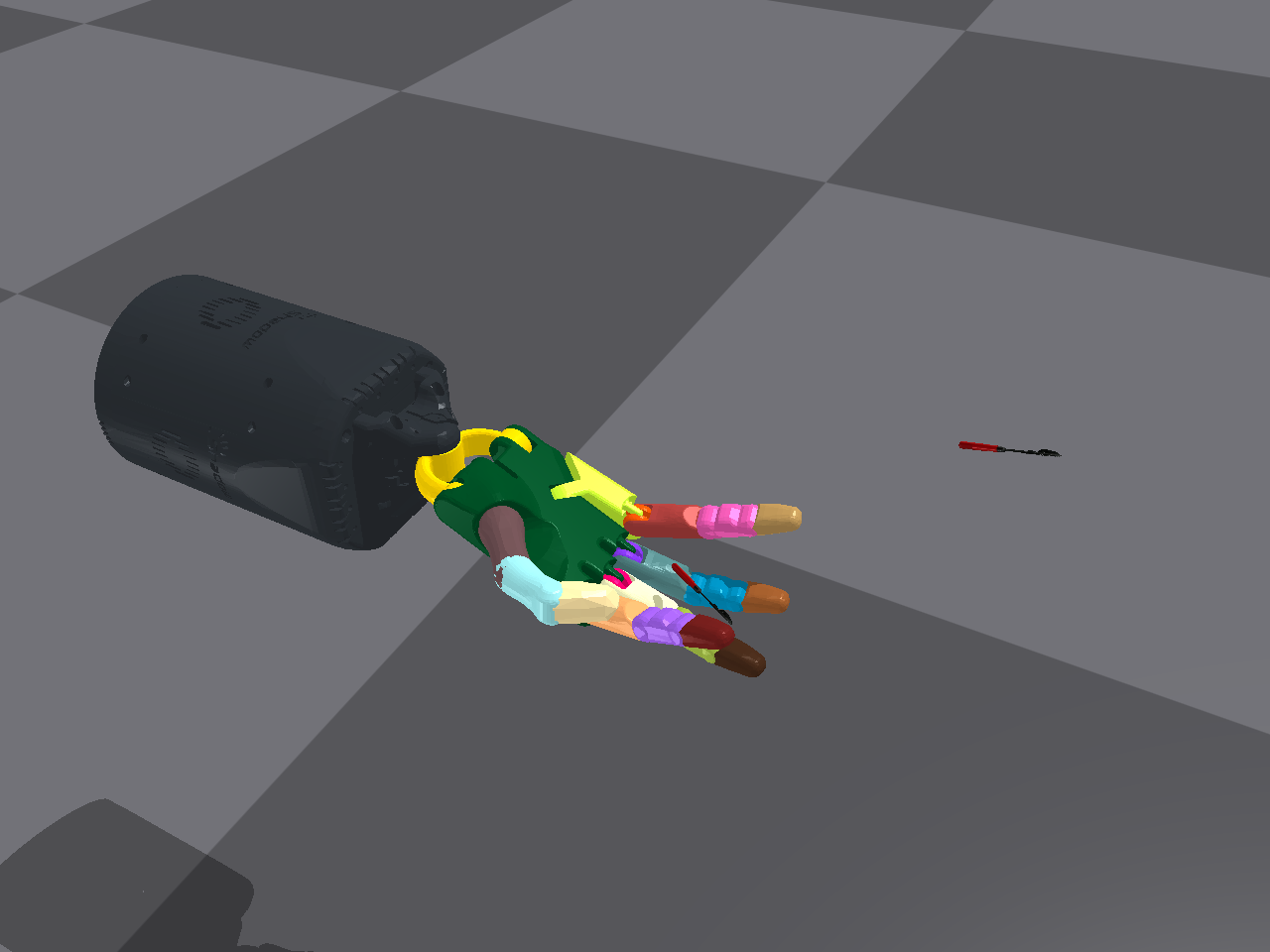} 
        \caption{}
    \end{subfigure}
        \begin{subfigure}[t]{0.24\textwidth}
        \centering
        \includegraphics[trim=300 240 80 200, clip,width=\linewidth]{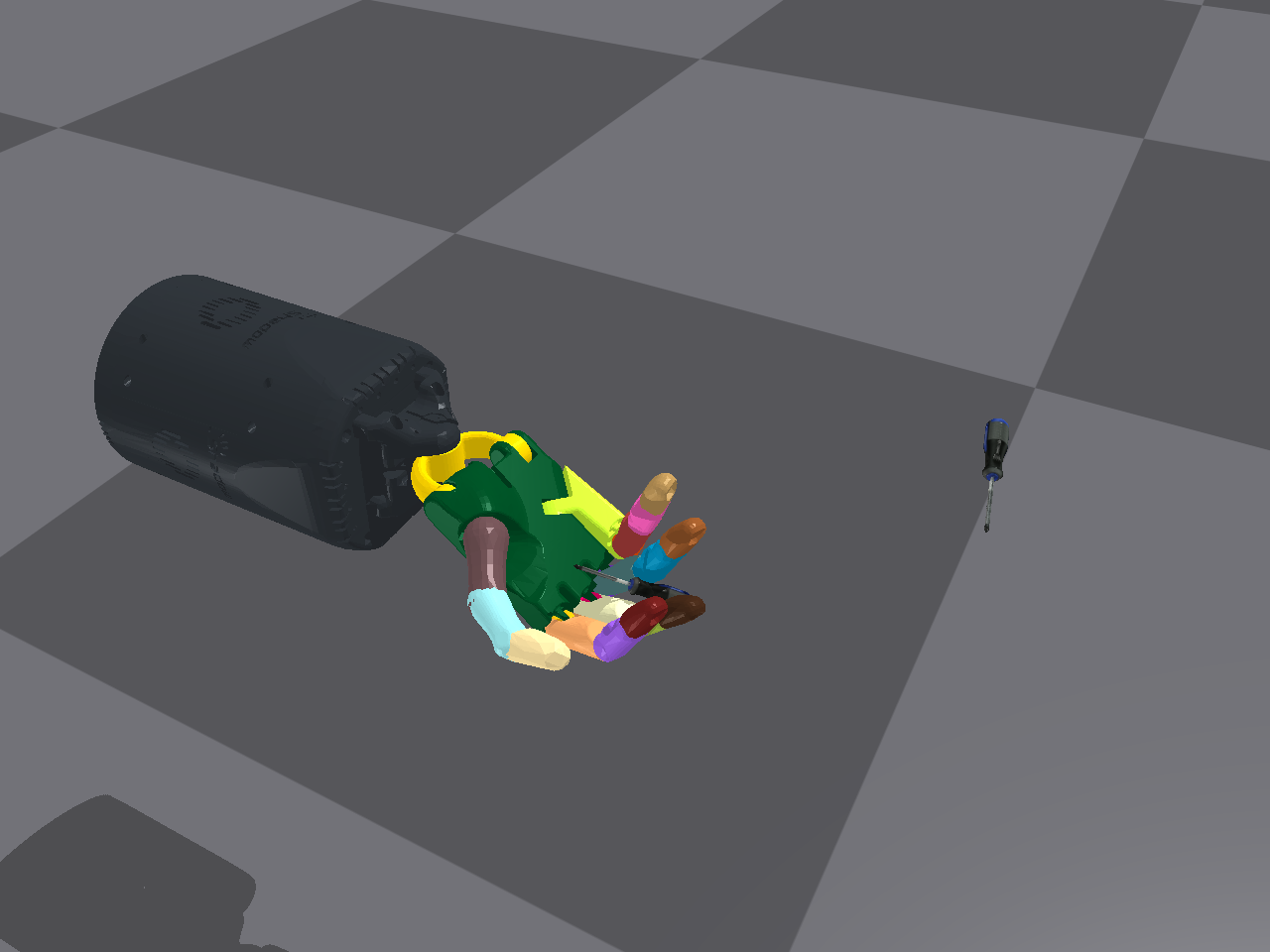} 
        \caption{}
    \end{subfigure}
        \begin{subfigure}[t]{0.24\textwidth}
        \centering
        \includegraphics[trim=300 240 80 200, clip,width=\linewidth]{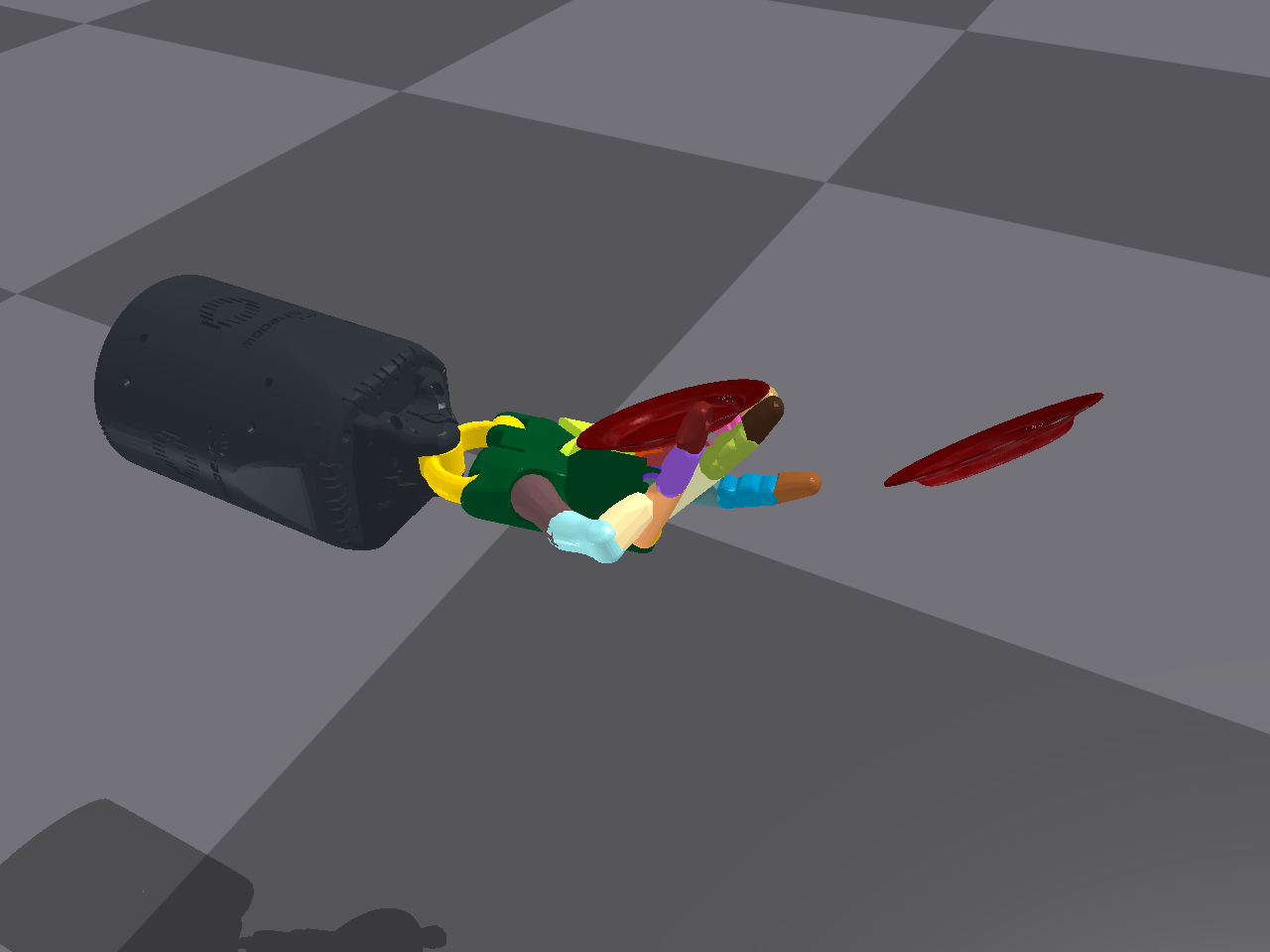} 
        \caption{}
    \end{subfigure}
    \begin{subfigure}[t]{0.24\textwidth}
        \centering
        \includegraphics[trim=300 240 80 200, clip,width=\linewidth]{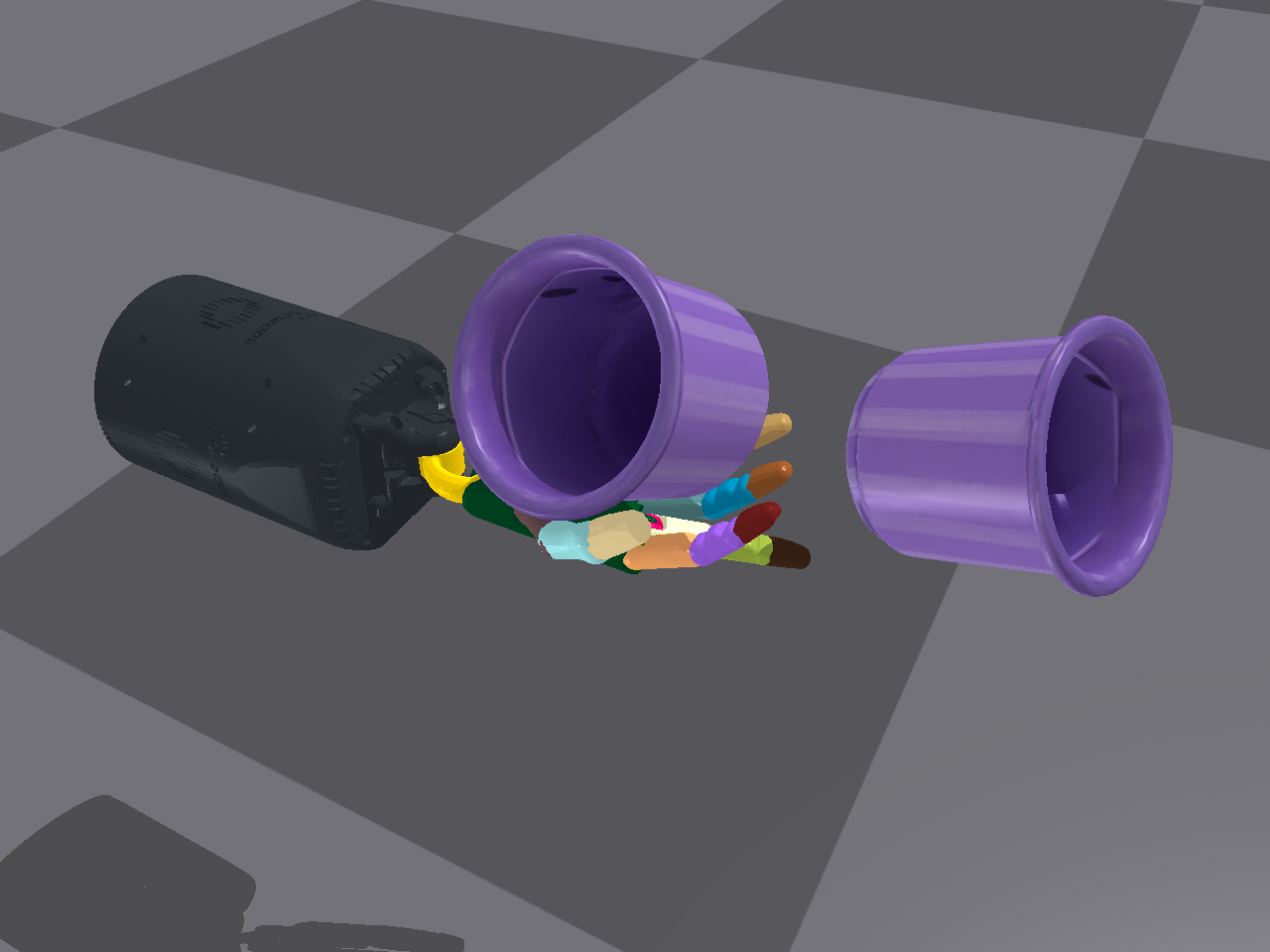} 
        \caption{}
    \end{subfigure}%
    \caption{Examples of failure cases. (a) and (b): objects are too small. (c): the object is reoriented close to the target orientation, but not close enough. (d): the object is too big and initialized around the palm border.} 
    \label{fig:failure_cases_up}
\end{figure}

\paragraph{Failure cases}
\figref{fig:failure_cases_up} shows some example failure cases. If the objects are too small, thin, or big, the objects are likely to fall. If objects are initialized close to the hand border, it is much more difficult for the hand to catch the objects. Another failure mode is that the objects are reoriented close to the goal orientation but not close enough to satisfy $\Delta \theta\leq \Bar{\theta}$.

\subsection{Hand faces downward (in the air)}
\label{app_subsec:hand_down}

\paragraph{Testing performance} For the case of reorienting objects in the air with the hand facing downward \tblref{tbl:down_air_success_rate} lists the success rates of different policies trained with/without domain randomization, and tested with/without domain randomization. 

\renewcommand{\arraystretch}{1.2}
\begin{table}[!htb]
\caption{Success rates (\%) of policies trained with hand facing downward and to reorient objects in the air. Due to the large number of environment steps required in this setup, we fine-tune the model trained without DR with randomized dynamics instead of training models with DR from scratch. }
\label{tbl:down_air_success_rate}
\centering
\resizebox{\columnwidth}{!}{
\begin{tabular}{cccc|ccc} 
\hline
\multicolumn{4}{c}{}                                                                                                                                   & 1               & 2                                                         & 3                 \\ 
\hline
\multirow{2}{*}{Exp. ID} & \multirow{2}{*}{Dataset} & \multirow{2}{*}{State}                             & \multicolumn{1}{c}{\multirow{2}{*}{Policy}} & \multicolumn{2}{c}{Train without DR}                                        & Finetune with DR  \\ 
\cline{5-7}
                         &                          &                                                    & \multicolumn{1}{c}{}                        & Test without DR & Test with DR                                              & Test with DR      \\ 
\hline
K                        & \multirow{4}{*}{EGAD}    & \multirow{2}{*}{Full state}                        & MLP                                         & \textbf{84.29} $\pm$ \textbf{0.9}  &\textbf{38.42} $\pm$ \textbf{1.5}                                            & \textbf{71.44} $\pm$ \textbf{1.3}    \\
L                        &                          &                                                    & RNN                                         & 82.27 $\pm$ 3.3  & \begin{tabular}[c]{@{}c@{}}36.55  $\pm$ 1.4\\\end{tabular} & 67.44 $\pm$ 2.1    \\
M                        &                          & \multicolumn{1}{l}{\multirow{2}{*}{Reduced state}} & MLP$\rightarrow$RNN                         & \textbf{77.05}  $\pm$ \textbf{1.6}  & 29.22  $\pm$ 2.6                                            & 59.23  $\pm$ 2.3   \\
N                        &                          & \multicolumn{1}{l}{}                               & RNN$\rightarrow$RNN                         & 74.10  $\pm$ 2.3  &  \textbf{37.01}  $\pm$ \textbf{1.5}                                            & \textbf{62.64}  $\pm$ \textbf{2.9}    \\ 
\hline
O                        & \multirow{6}{*}{YCB}     & \multirow{3}{*}{Full state}                        & MLP                                         & 58.95  $\pm$ 2.0  & 26.04  $\pm$ 1.9                                            & 44.84  $\pm$ 1.3    \\
P                        &                          &                                                    & RNN                                         & 52.81  $\pm$ 1.7  &\textbf{26.22}  $\pm$ \textbf{1.0}                                            & 40.44  $\pm$ 1.5    \\
Q                        &                          &                                                    & RNN + $g$-curr                              & \textbf{74.74}  $\pm$ \textbf{1.2}  & 25.56  $\pm$ 2.9                                            & \textbf{54.24}  $\pm$ \textbf{1.4}    \\
R                        &                          & \multirow{3}{*}{Reduced state}                     & MLP$\rightarrow$RNN                        & 46.76  $\pm$ 2.5  & \textbf{25.49}  $\pm$ \textbf{1.4}                                            & 34.14  $\pm$ 1.3    \\
S                        &                          &                                                    & RNN$\rightarrow$RNN                         & 45.22  $\pm$ 2.1  & 24.45  $\pm$ 1.2                                            & 31.63  $\pm$ 1.6    \\
T                        &                          &                                                    & RNN + $g$-curr$\rightarrow$ RNN            & \textbf{67.33}  $\pm$ \textbf{1.9}  & 19.77  $\pm$ 2.8                                            & \textbf{48.58}  $\pm$ \textbf{2.3}    \\
\hline
\end{tabular}
}
\end{table}
\renewcommand{\arraystretch}{1}

\begin{figure}[!htb]
\centering
    \begin{subfigure}[t]{0.24\textwidth}
        \centering
        \includegraphics[width=\linewidth]{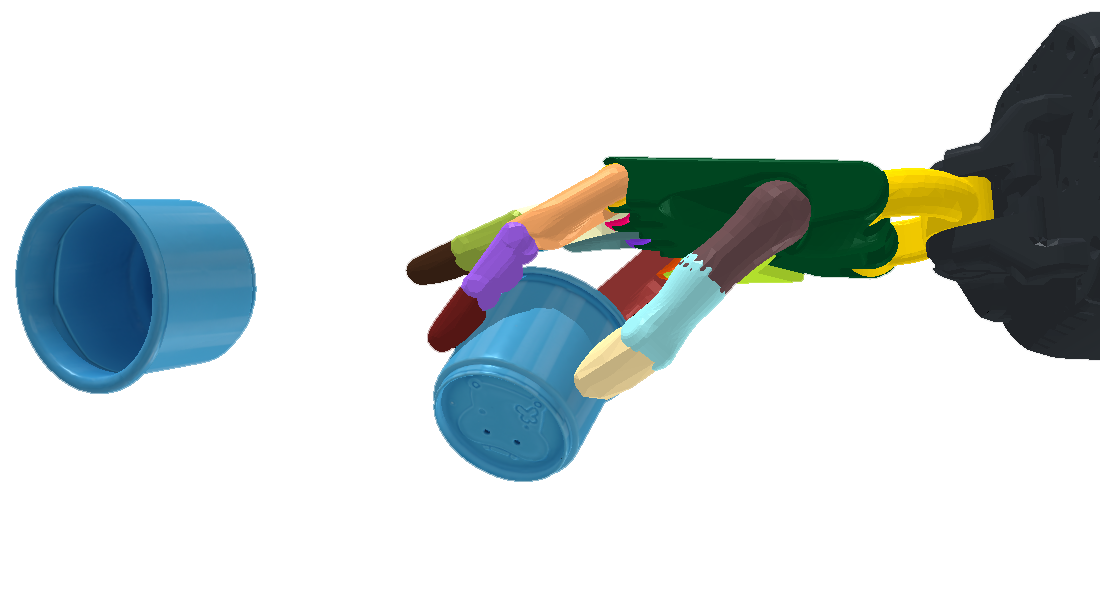}
    \end{subfigure}
    \begin{subfigure}[t]{0.24\textwidth}
        \centering
        \includegraphics[width=\linewidth]{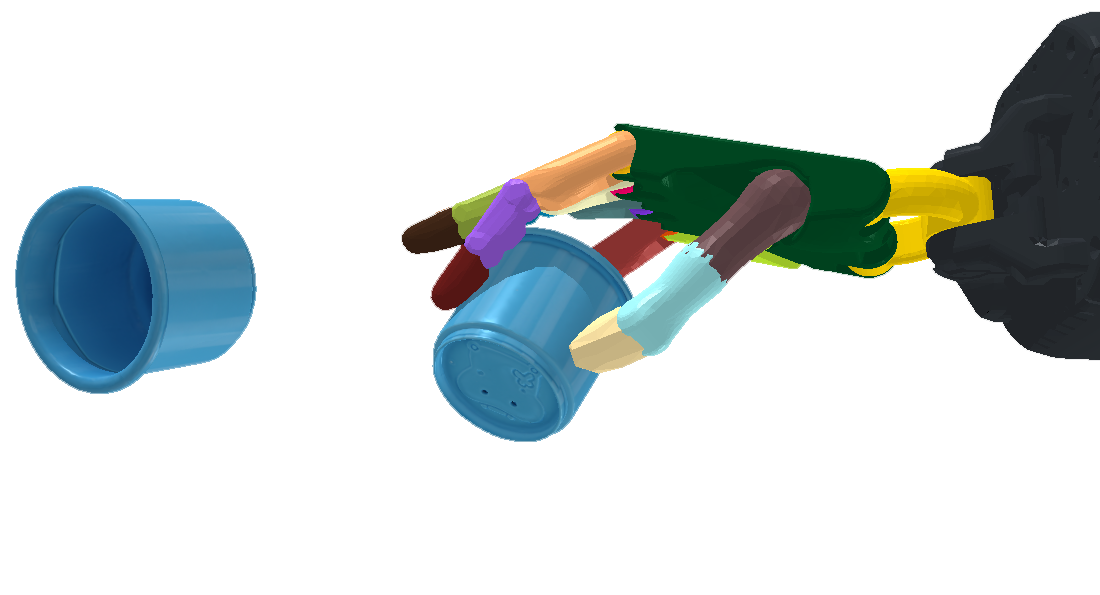} 
    \end{subfigure}
    \begin{subfigure}[t]{0.24\textwidth}
        \centering
        \includegraphics[width=\linewidth]{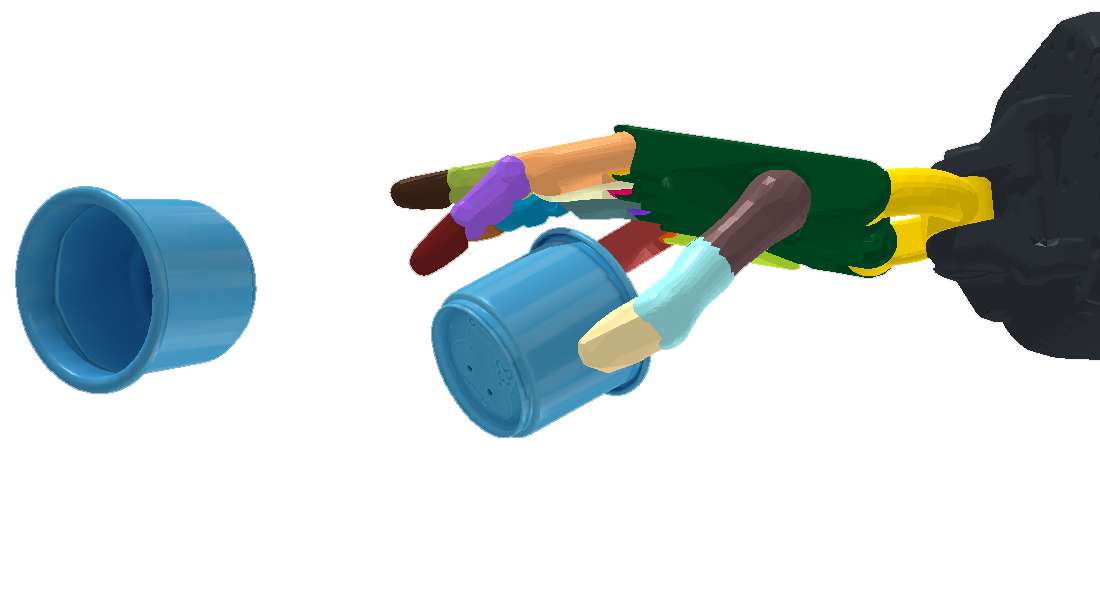} 
    \end{subfigure}
    \begin{subfigure}[t]{0.24\textwidth}
        \centering
        \includegraphics[width=\linewidth]{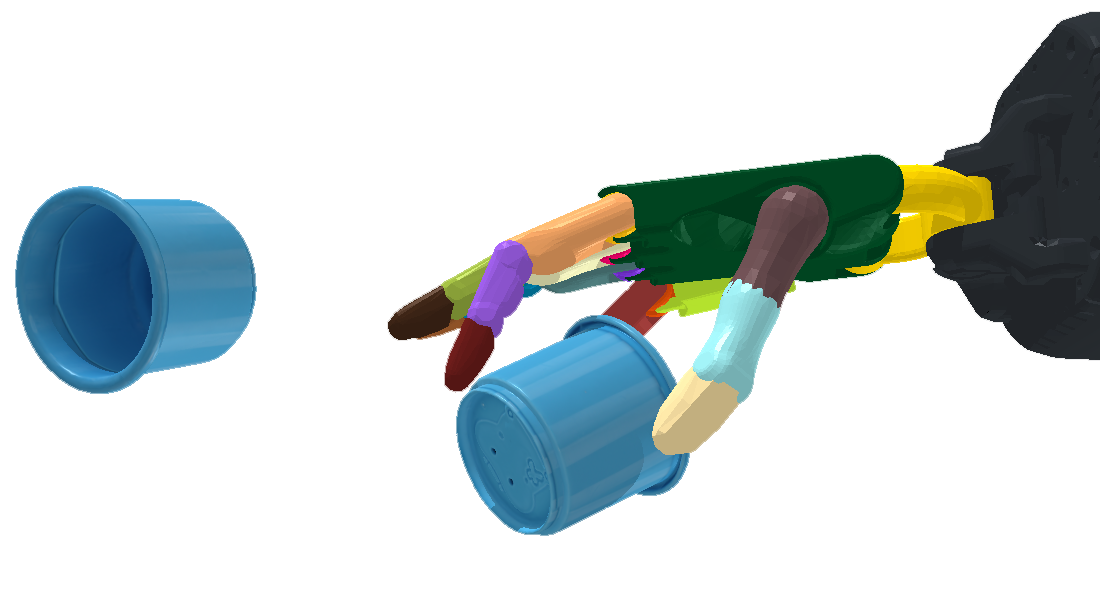} 
    \end{subfigure}%
    \\
    \begin{subfigure}[t]{0.24\textwidth}
        \centering
        \includegraphics[width=\linewidth]{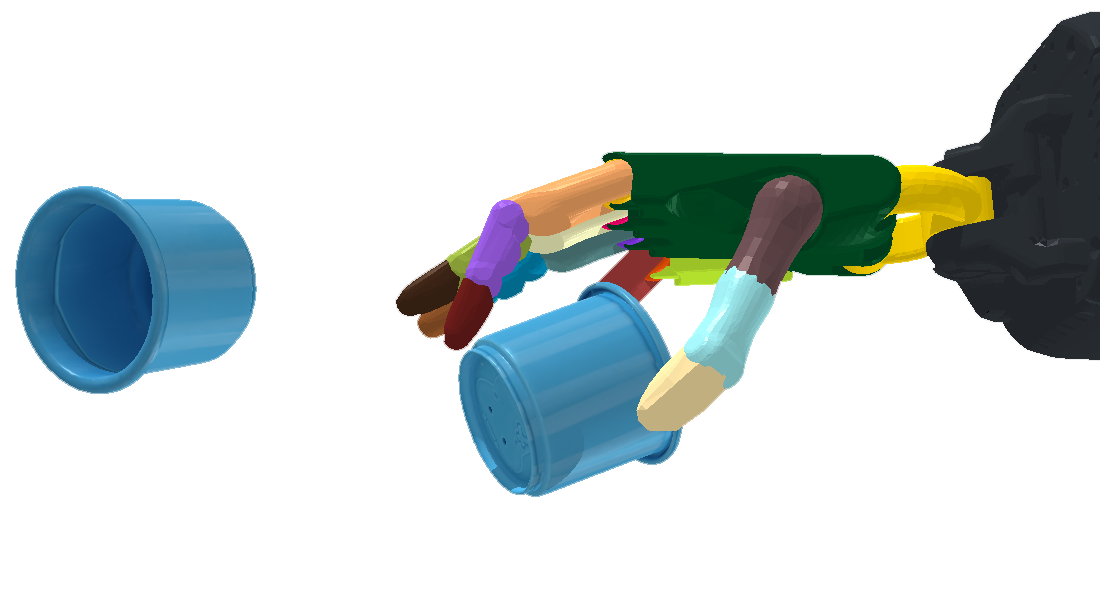} 
    \end{subfigure}
    \begin{subfigure}[t]{0.24\textwidth}
        \centering
        \includegraphics[width=\linewidth]{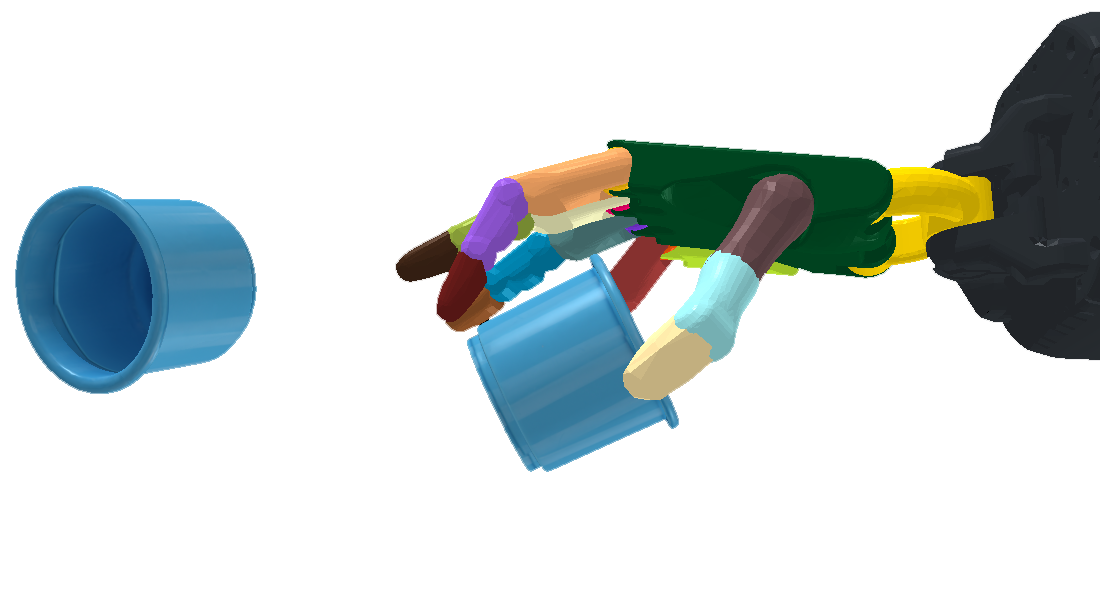} 
    \end{subfigure}
    \begin{subfigure}[t]{0.24\textwidth}
        \centering
        \includegraphics[width=\linewidth]{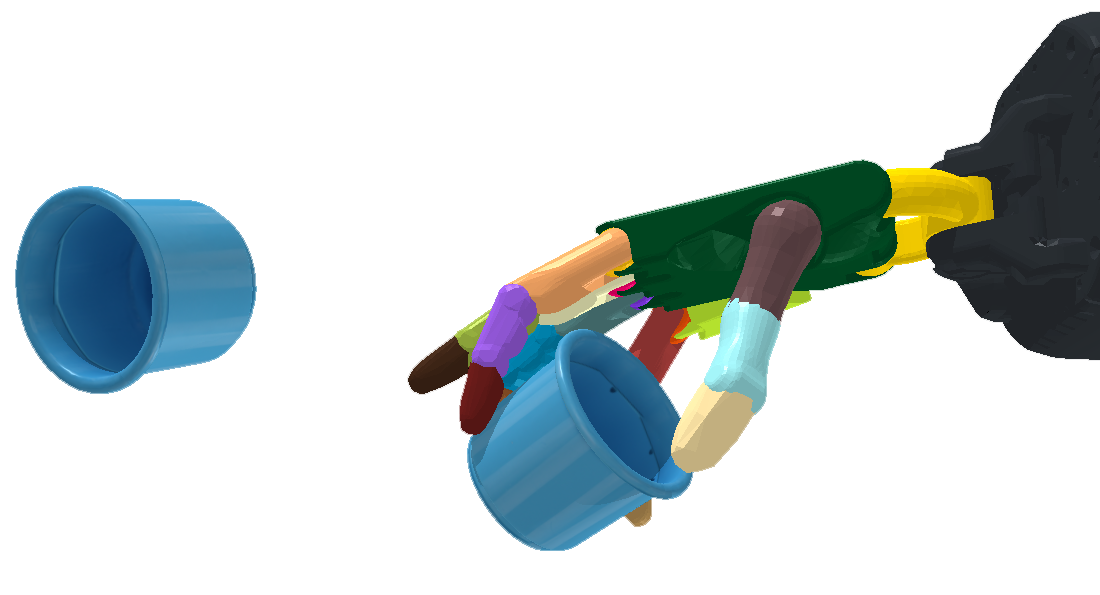} 
    \end{subfigure}
    \begin{subfigure}[t]{0.24\textwidth}
        \centering
        \includegraphics[width=\linewidth]{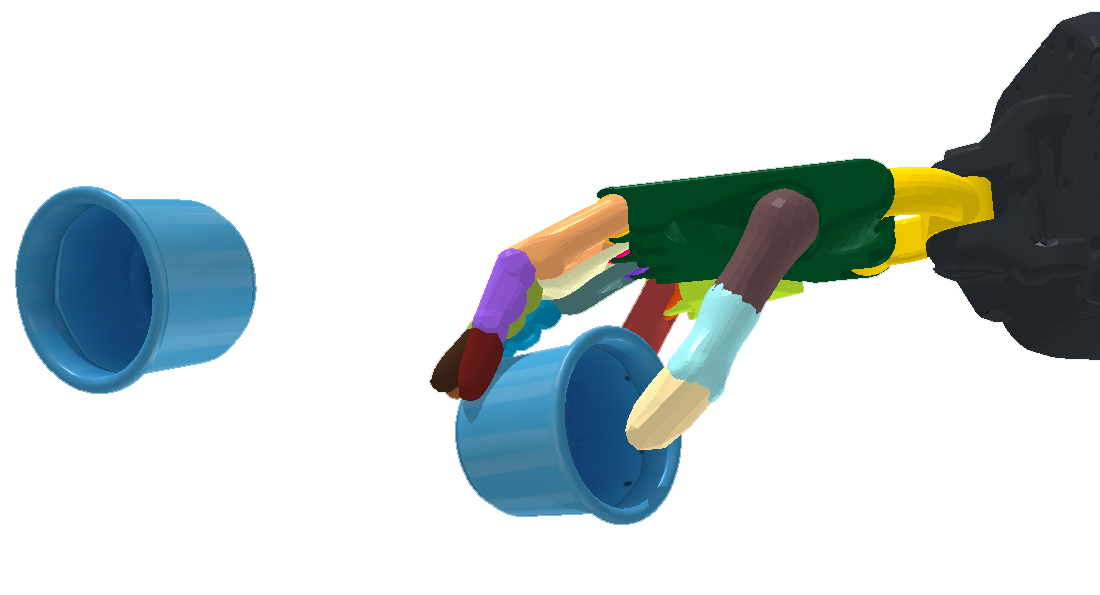} 
    \end{subfigure}%
    \\
     \begin{subfigure}[t]{0.24\textwidth}
        \centering
        \includegraphics[width=\linewidth]{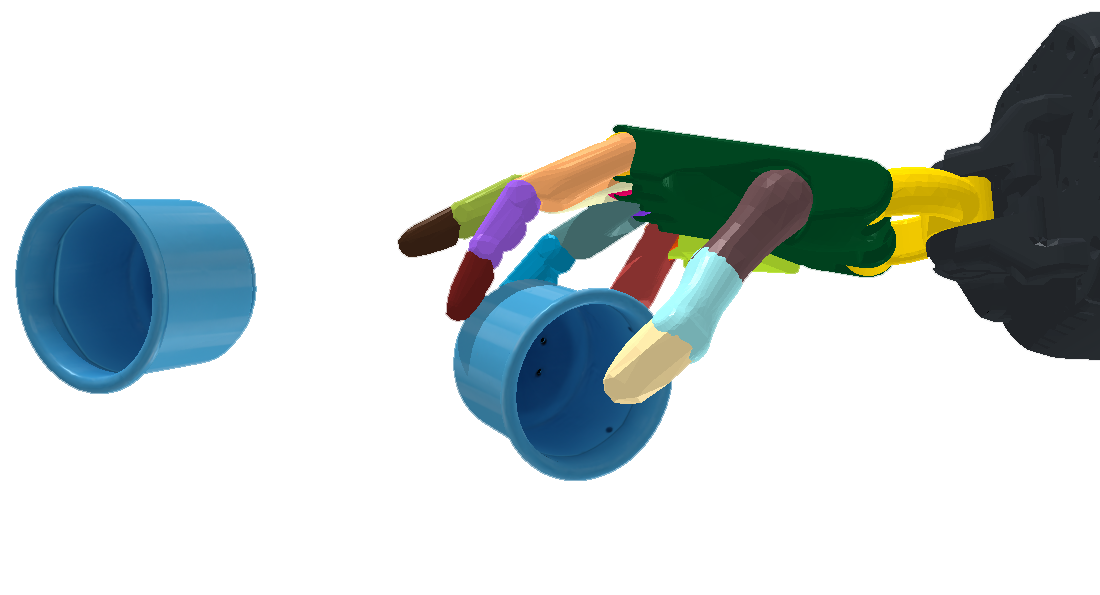} 
    \end{subfigure}
    \begin{subfigure}[t]{0.24\textwidth}
        \centering
        \includegraphics[width=\linewidth]{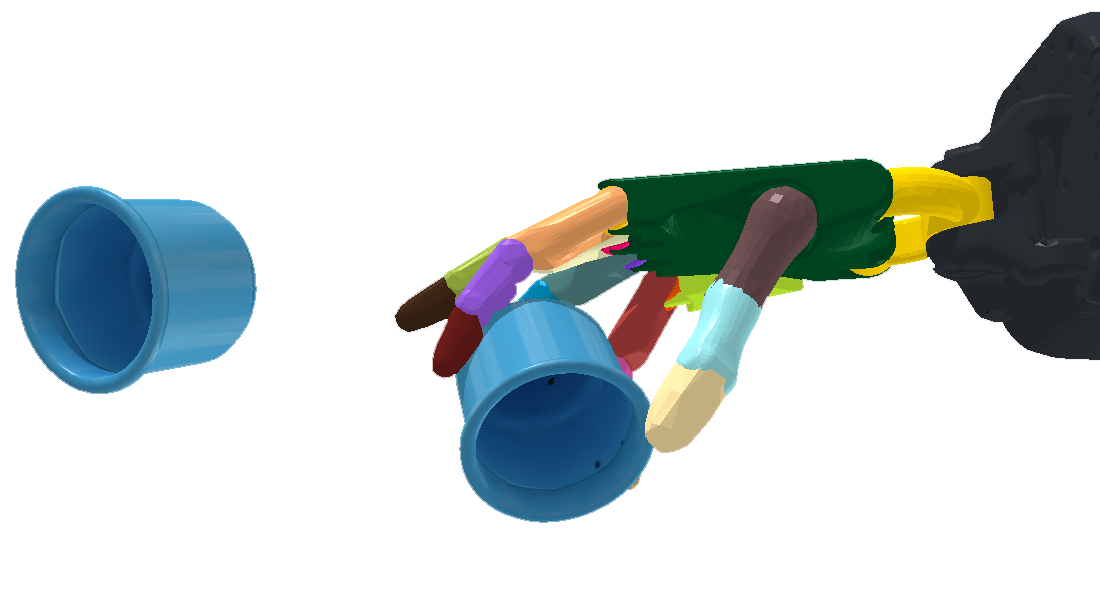} 
    \end{subfigure}
    \begin{subfigure}[t]{0.24\textwidth}
        \centering
        \includegraphics[width=\linewidth]{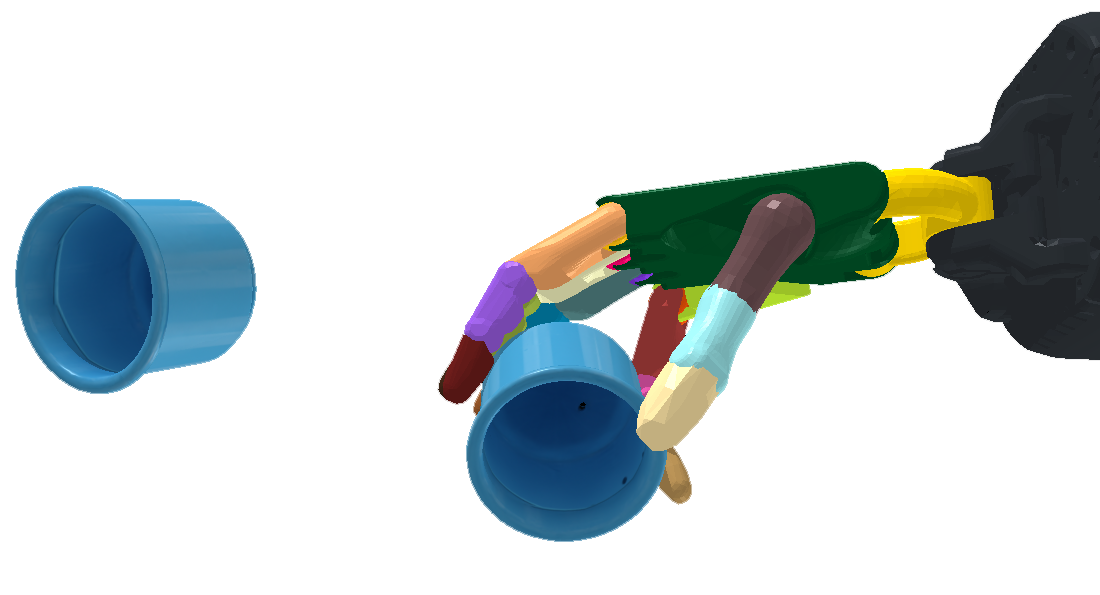} 
    \end{subfigure}
    \begin{subfigure}[t]{0.24\textwidth}
        \centering
        \includegraphics[width=\linewidth]{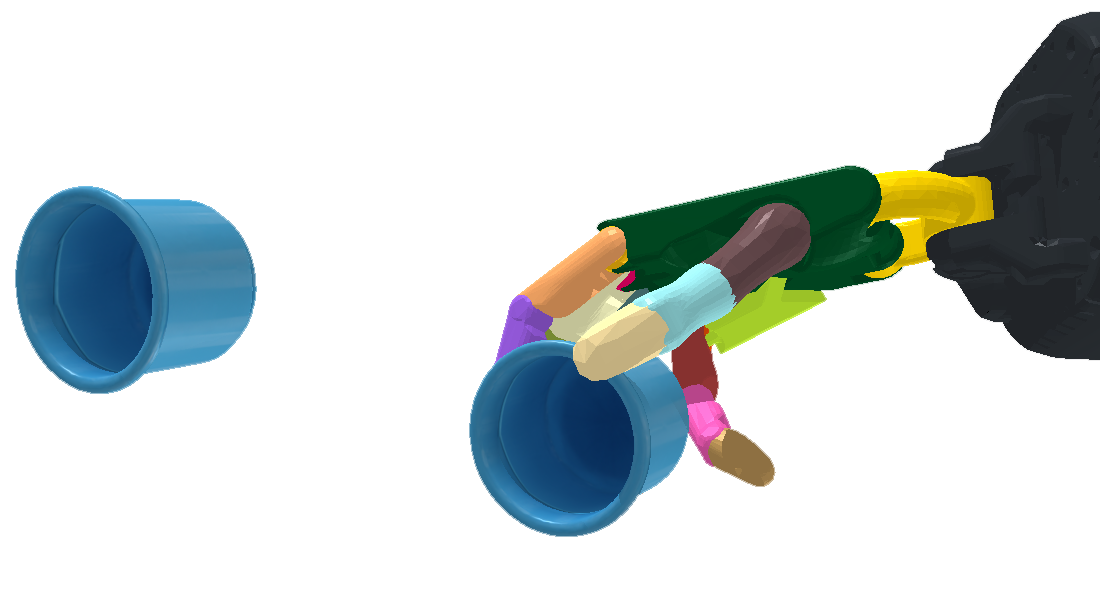} 
    \end{subfigure}%
    \caption{An example of reorienting a cup with the hand facing downward. From left to right, top to bottom, we show the some moments in an episode.}
\label{fig:down_reorient_cup}
\end{figure}

\begin{figure}[!htb]
\centering
    \begin{subfigure}[t]{0.24\textwidth}
        \centering
        \includegraphics[width=\linewidth]{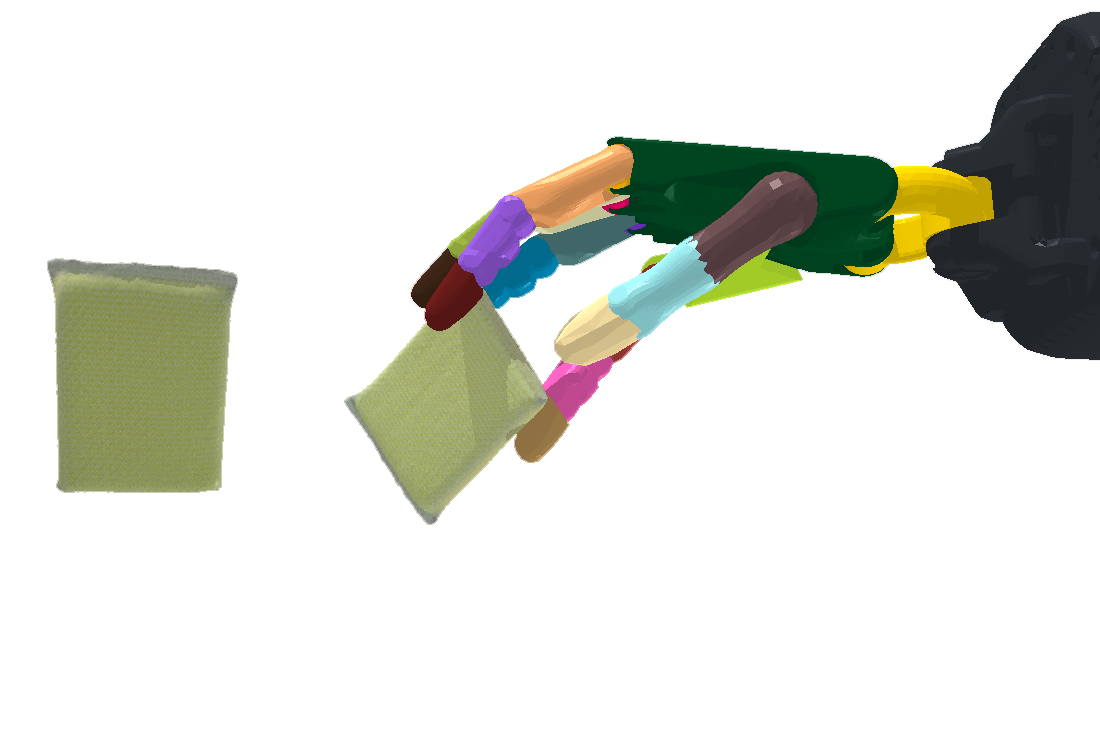}
    \end{subfigure}
    \begin{subfigure}[t]{0.24\textwidth}
        \centering
        \includegraphics[width=\linewidth]{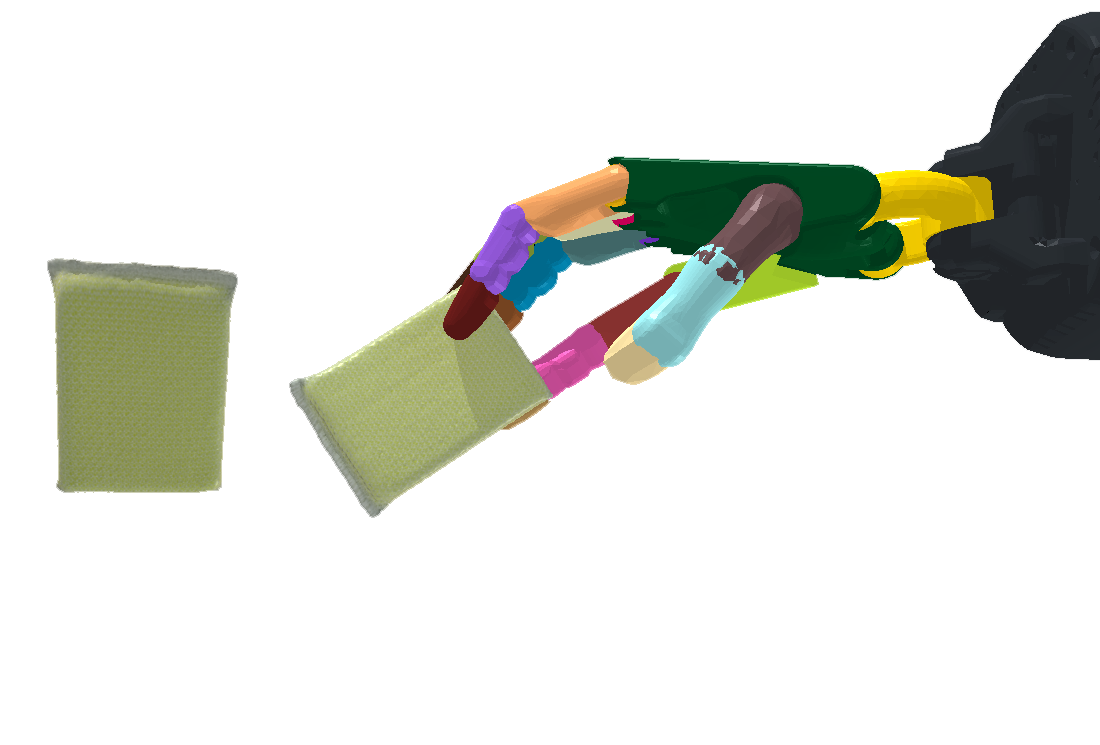} 
    \end{subfigure}
    \begin{subfigure}[t]{0.24\textwidth}
        \centering
        \includegraphics[width=\linewidth]{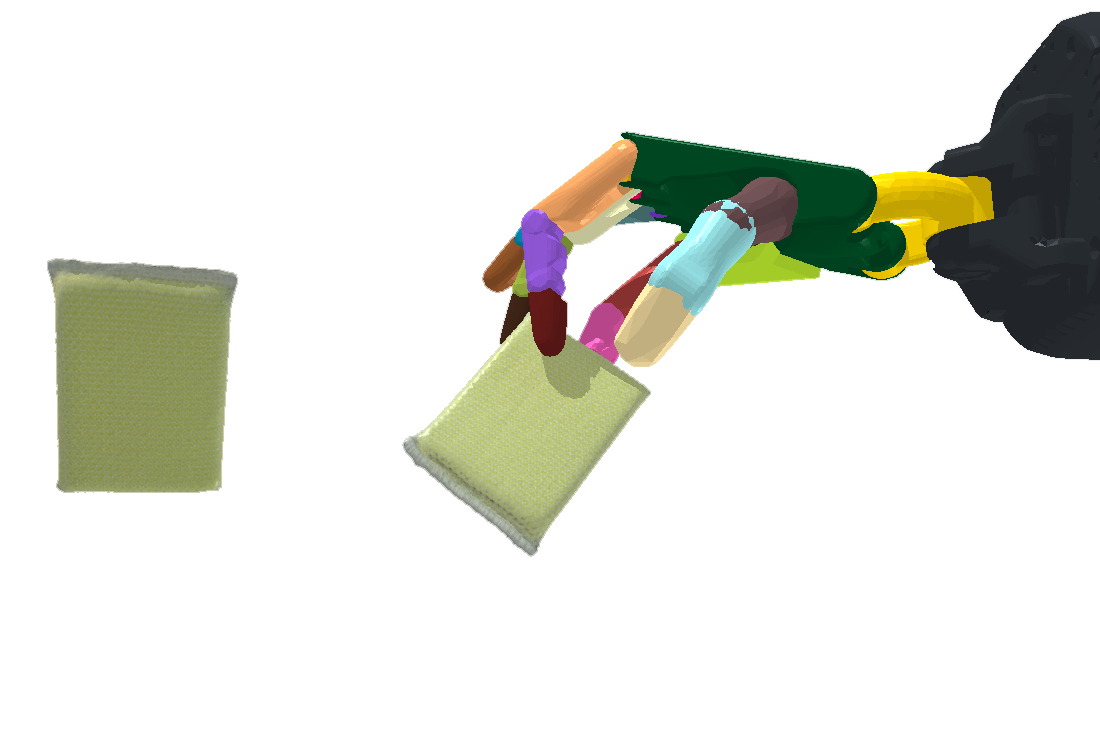} 
    \end{subfigure}
    \begin{subfigure}[t]{0.24\textwidth}
        \centering
        \includegraphics[width=\linewidth]{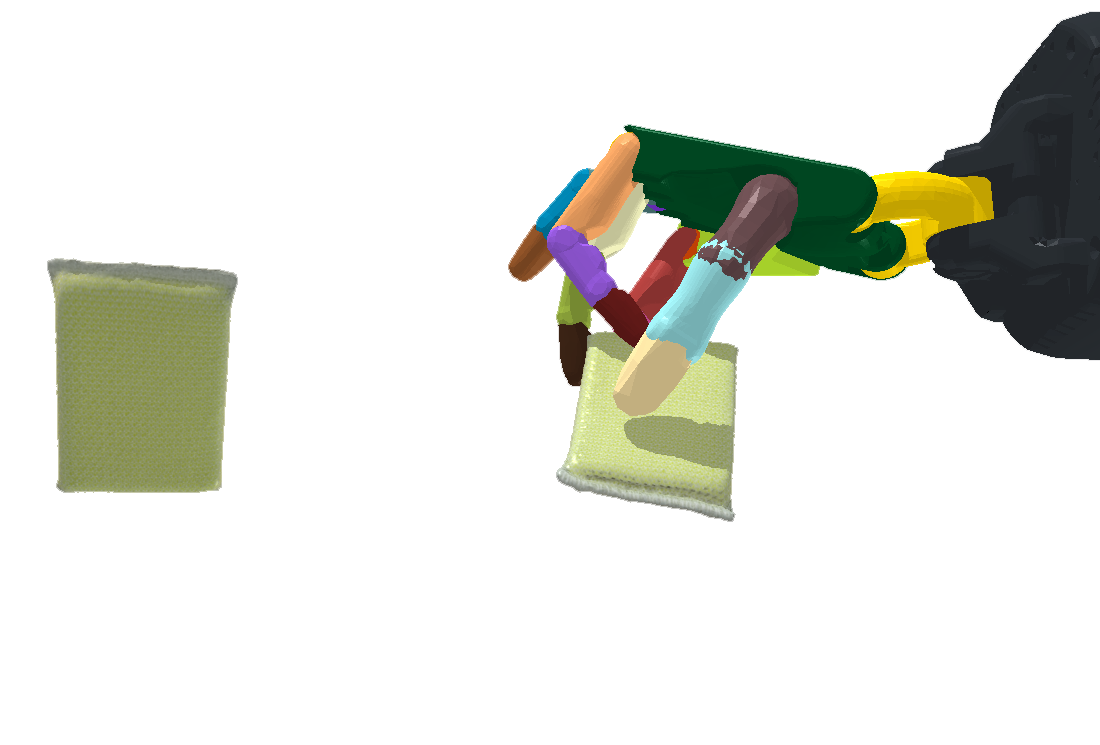} 
    \end{subfigure}%
    \\
    \begin{subfigure}[t]{0.24\textwidth}
        \centering
        \includegraphics[width=\linewidth]{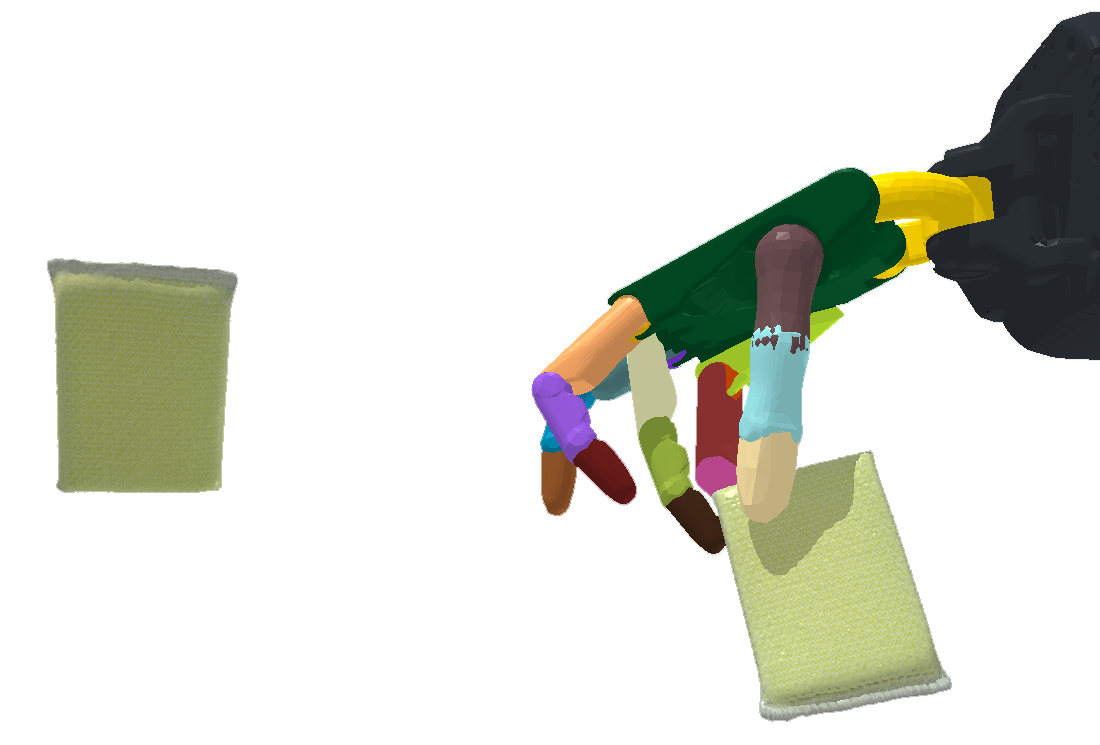} 
    \end{subfigure}
    \begin{subfigure}[t]{0.24\textwidth}
        \centering
        \includegraphics[width=\linewidth]{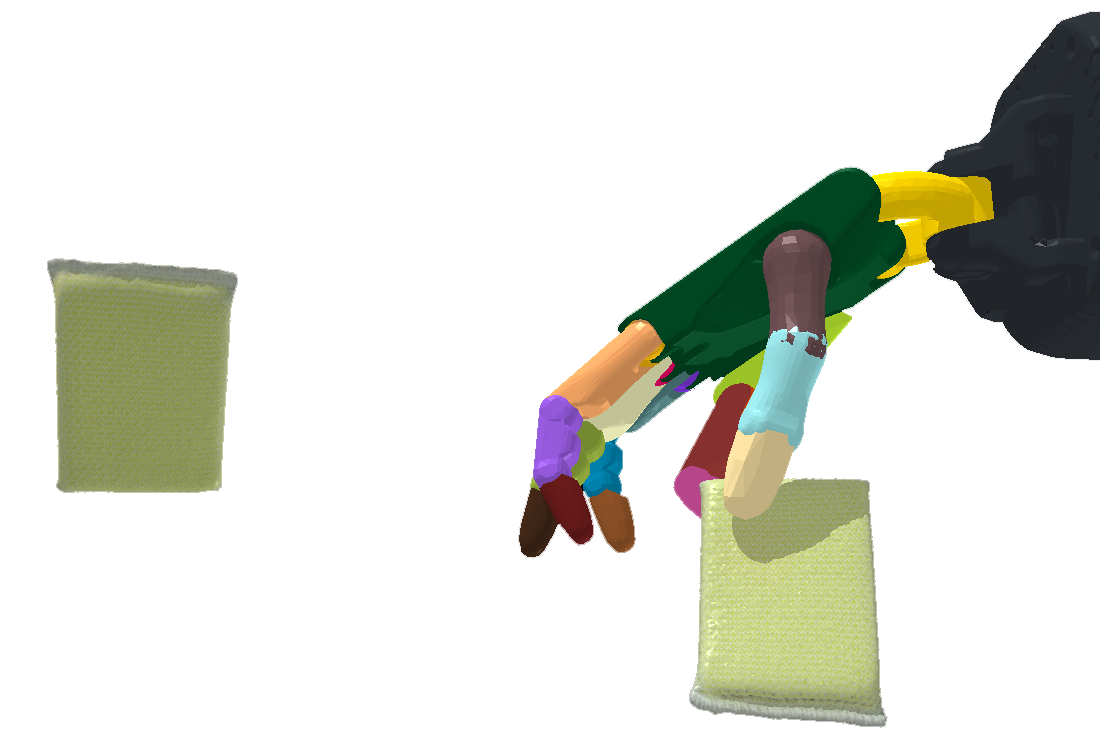} 
    \end{subfigure}
    \begin{subfigure}[t]{0.24\textwidth}
        \centering
        \includegraphics[width=\linewidth]{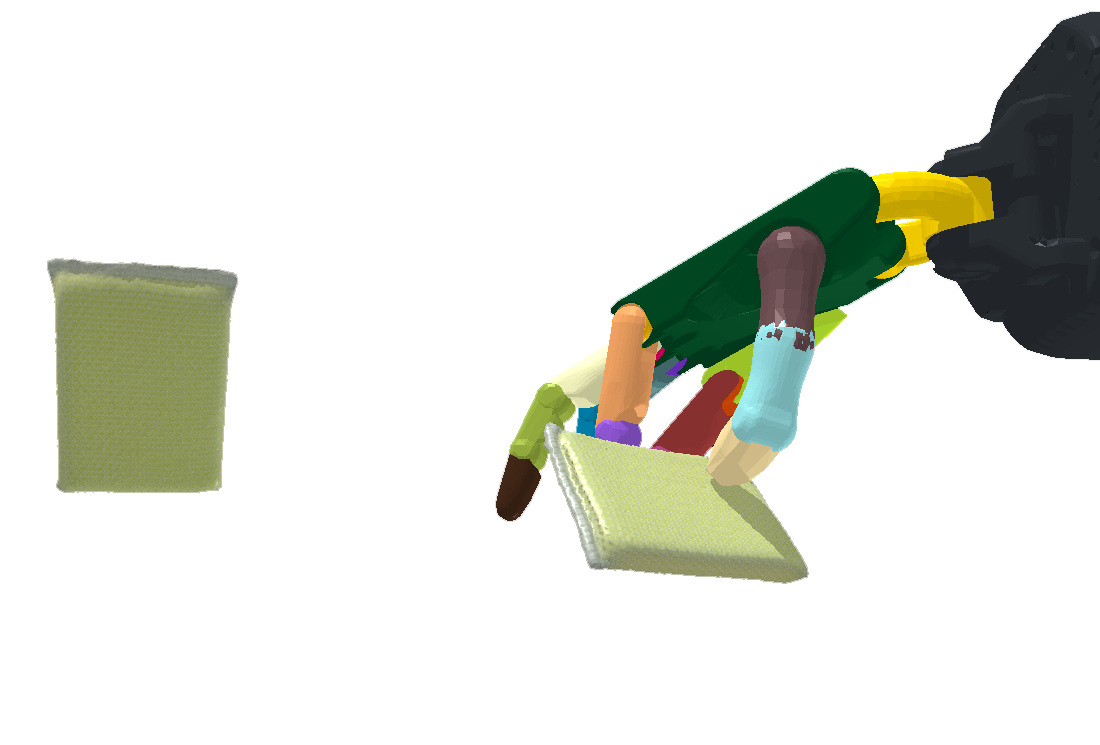} 
    \end{subfigure}
    \begin{subfigure}[t]{0.24\textwidth}
        \centering
        \includegraphics[width=\linewidth]{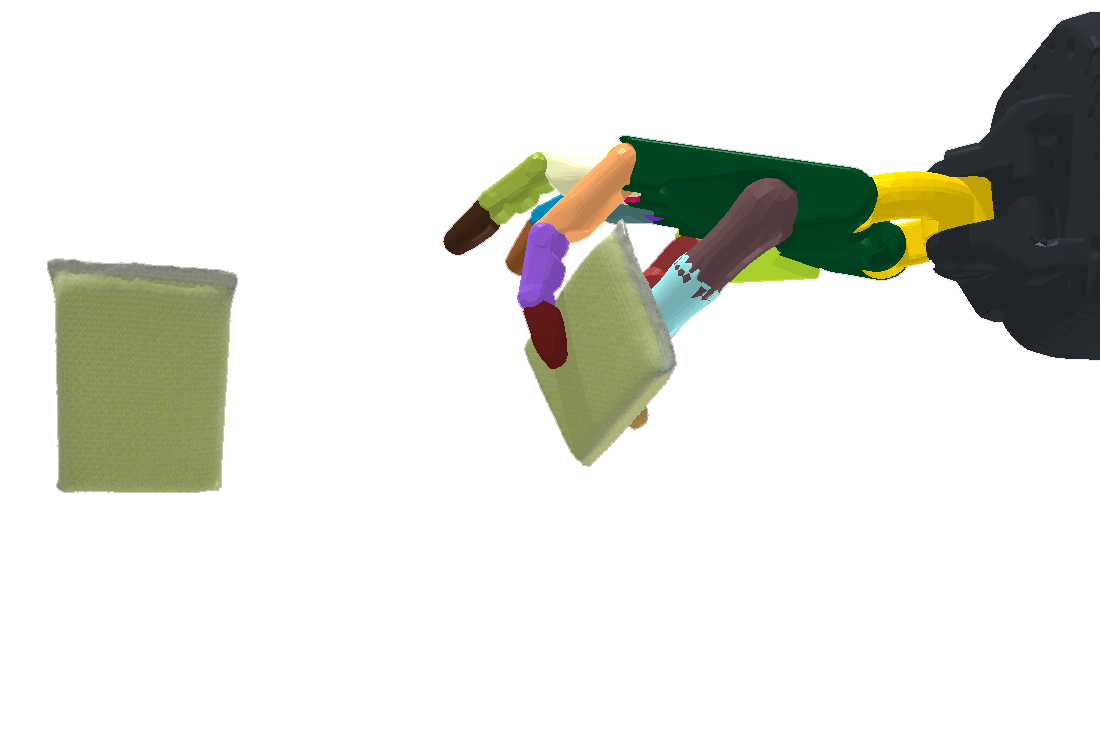} 
    \end{subfigure}%
    \\
     \begin{subfigure}[t]{0.24\textwidth}
        \centering
        \includegraphics[width=\linewidth]{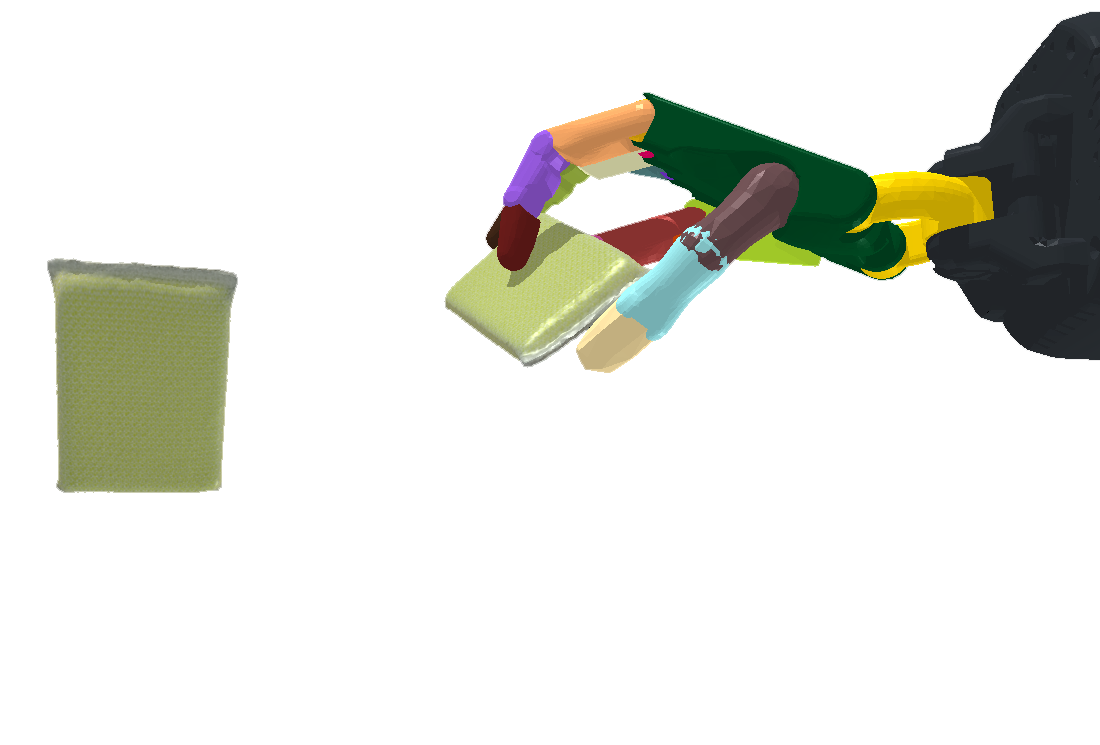} 
    \end{subfigure}
    \begin{subfigure}[t]{0.24\textwidth}
        \centering
        \includegraphics[width=\linewidth]{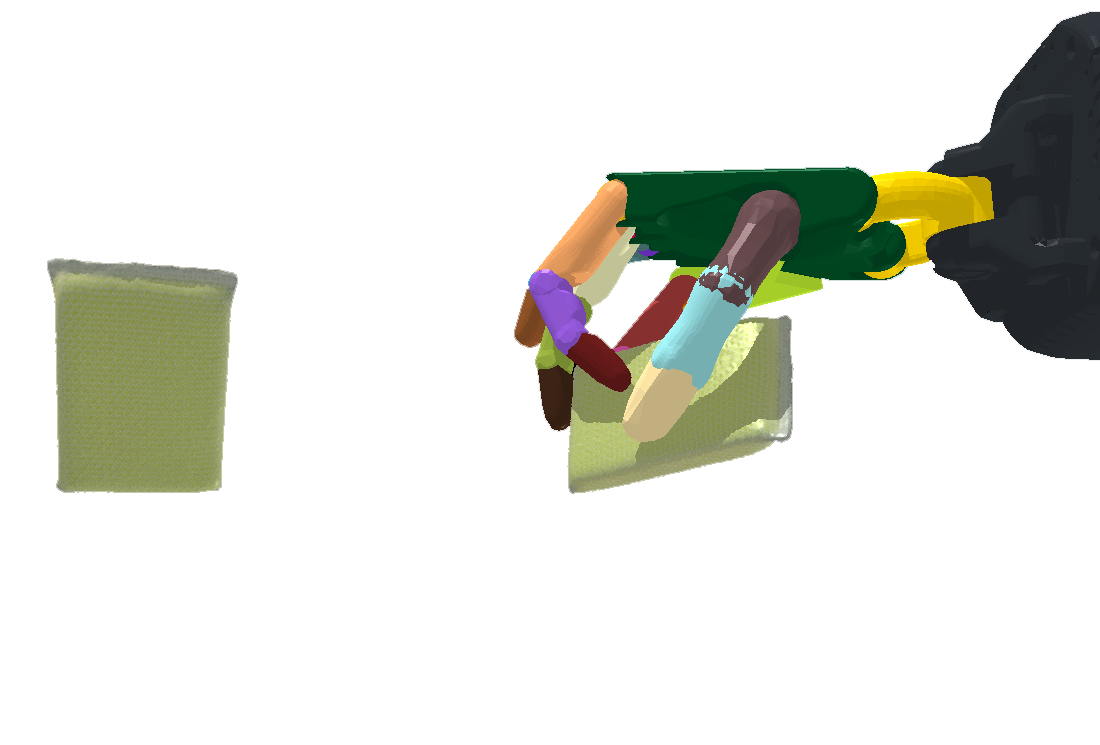} 
    \end{subfigure}
    \begin{subfigure}[t]{0.24\textwidth}
        \centering
        \includegraphics[width=\linewidth]{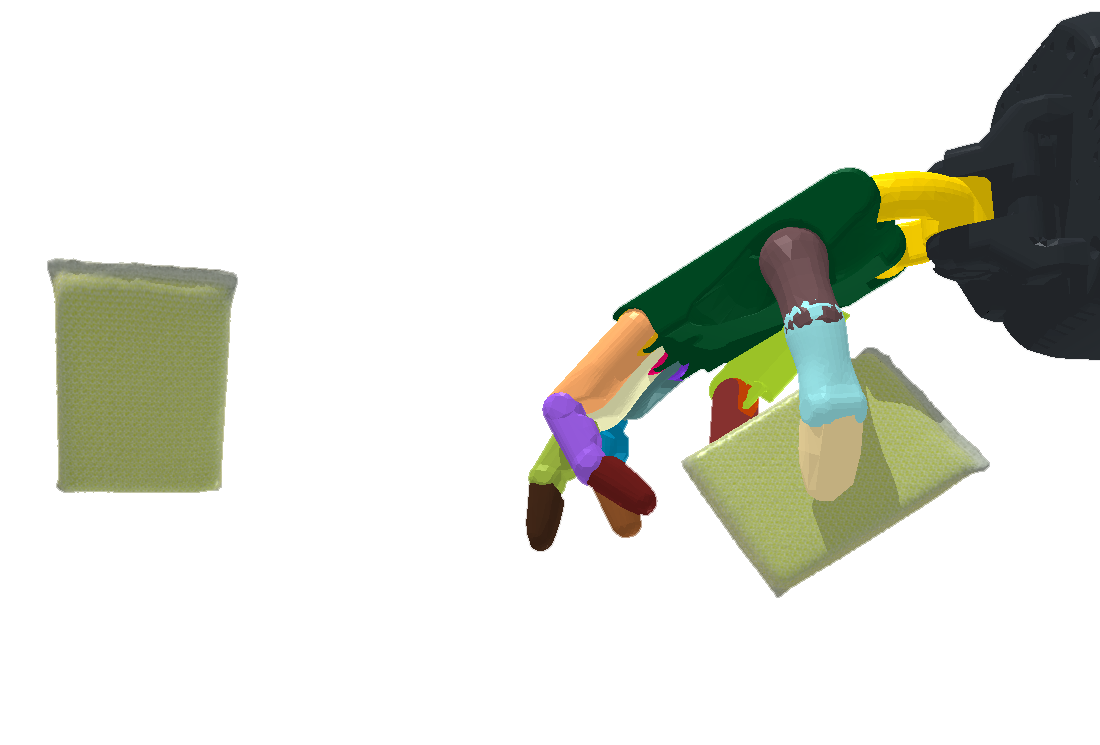} 
    \end{subfigure}
    \begin{subfigure}[t]{0.24\textwidth}
        \centering
        \includegraphics[width=\linewidth]{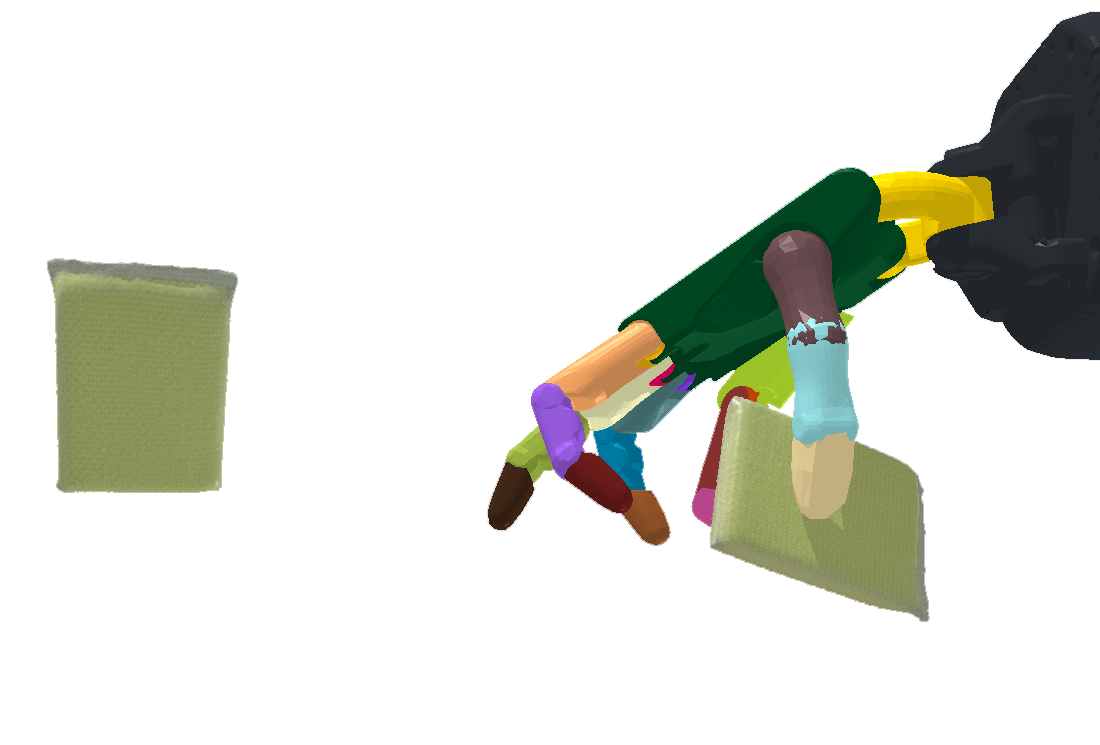} 
    \end{subfigure}%
    \\
         \begin{subfigure}[t]{0.24\textwidth}
        \centering
        \includegraphics[width=\linewidth]{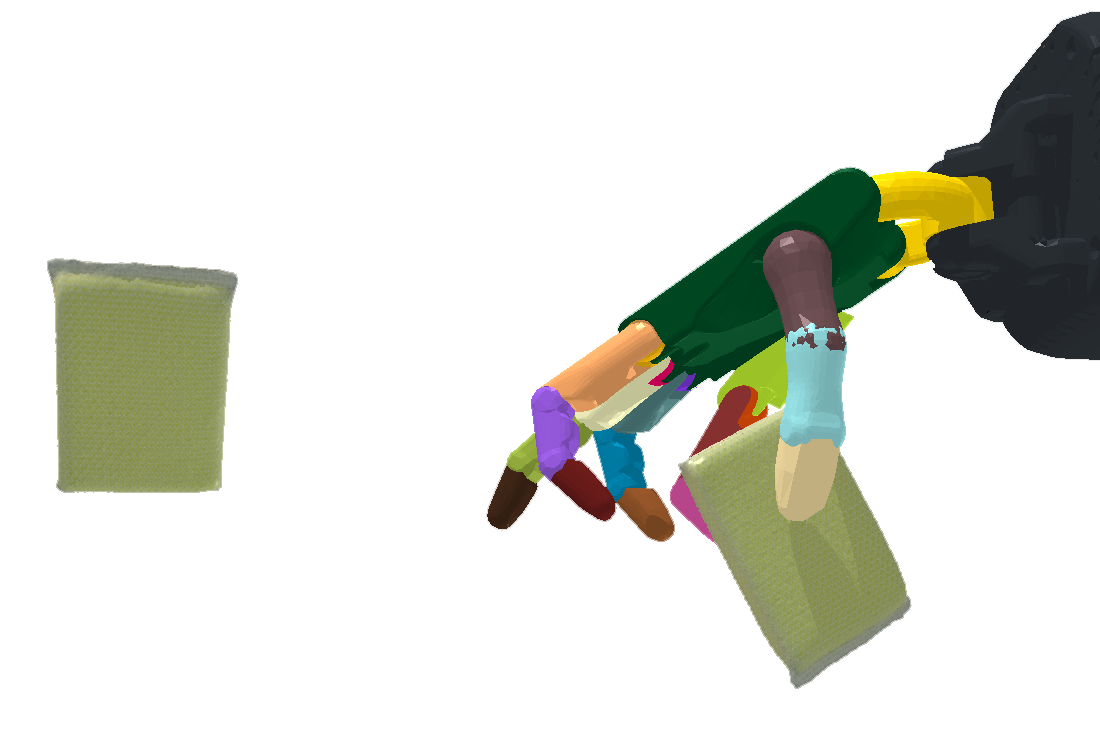} 
    \end{subfigure}
    \begin{subfigure}[t]{0.24\textwidth}
        \centering
        \includegraphics[width=\linewidth]{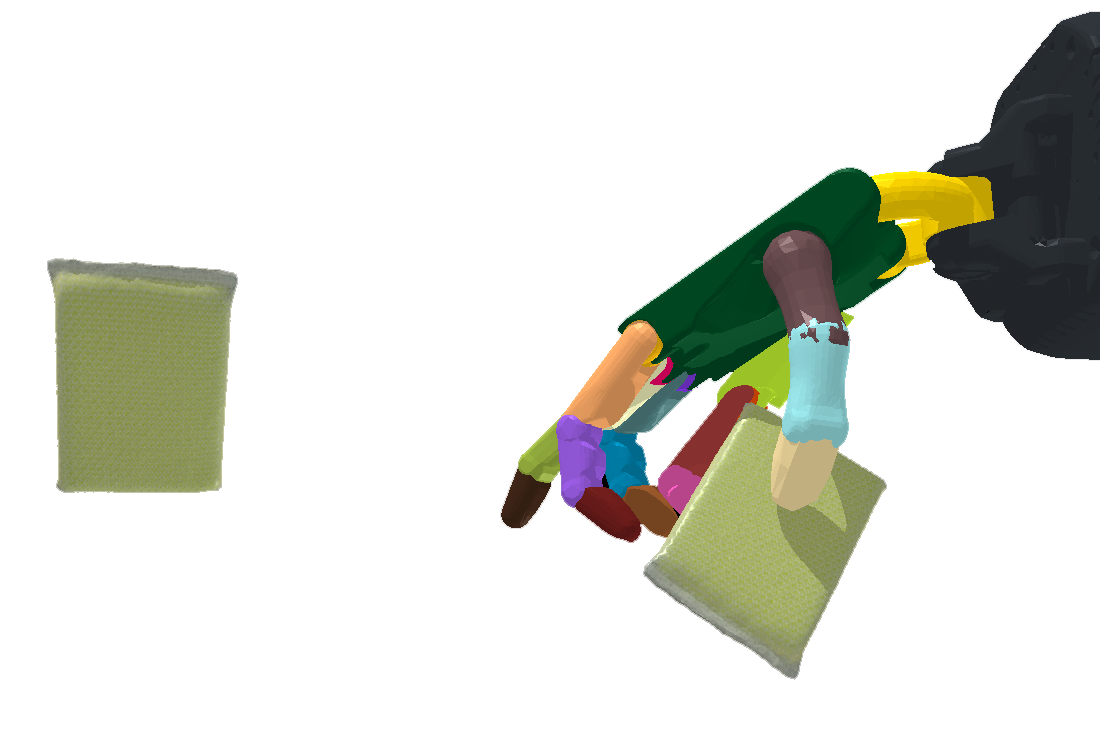} 
    \end{subfigure}
    \begin{subfigure}[t]{0.24\textwidth}
        \centering
        \includegraphics[width=\linewidth]{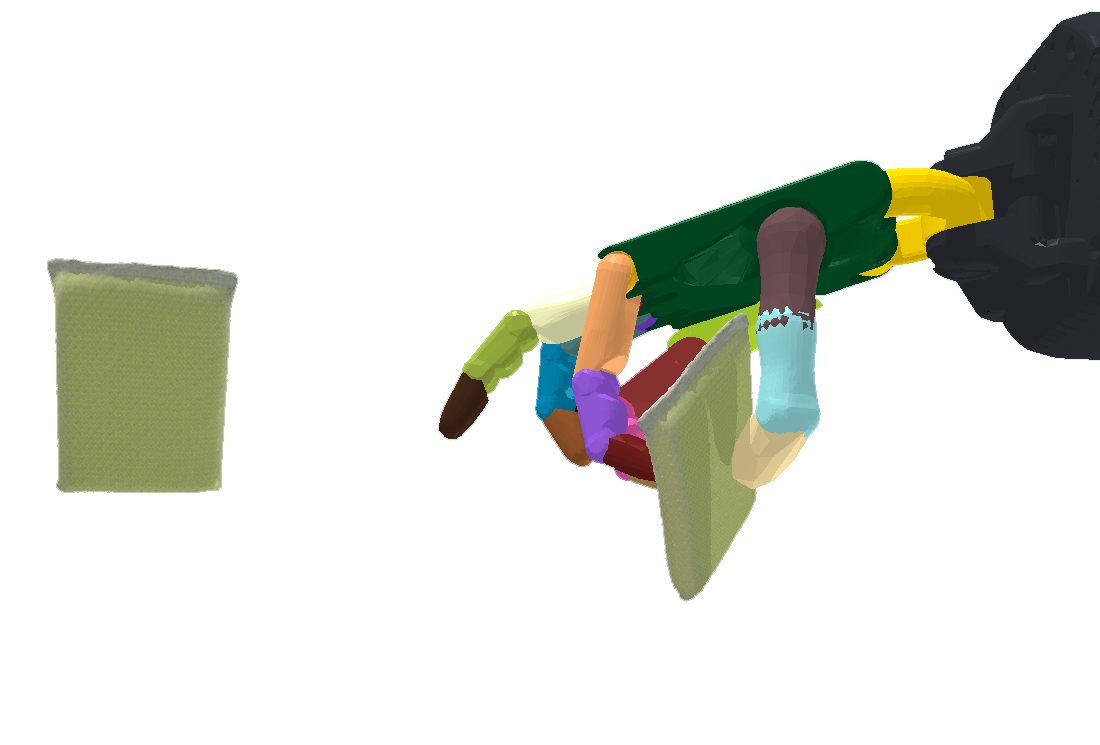} 
    \end{subfigure}
    \begin{subfigure}[t]{0.24\textwidth}
        \centering
        \includegraphics[width=\linewidth]{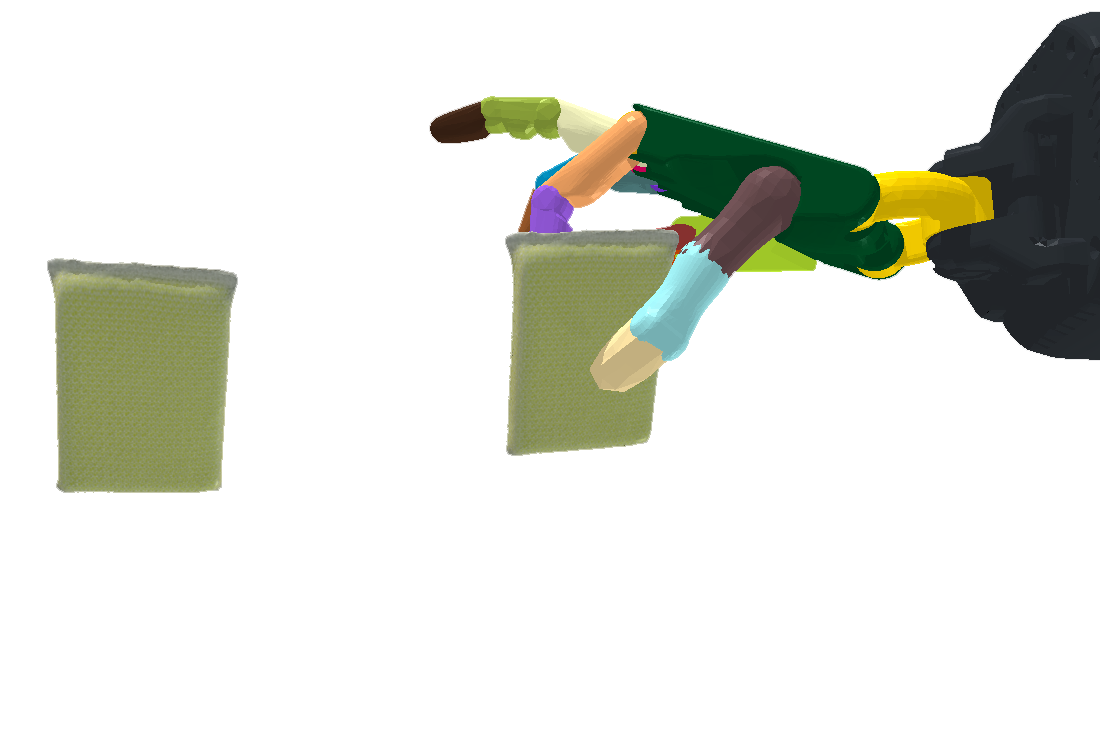} 
    \end{subfigure}%
    \caption{An example of reorienting a sponge with the hand facing downward. From left to right, top to bottom, we show the some moments in an episode.}
\label{fig:down_reorient_sponge}
\end{figure}

\paragraph{Example visualization}
We show an example of reorienting a cup in \figref{fig:down_reorient_cup} and an example of reorienting a sponge in \figref{fig:down_reorient_sponge}. More examples are available at \videourl.

\subsection{Success rate on each type of YCB objects}

\begin{figure}[!htb]
    \centering
    \includegraphics[width=0.95\linewidth]{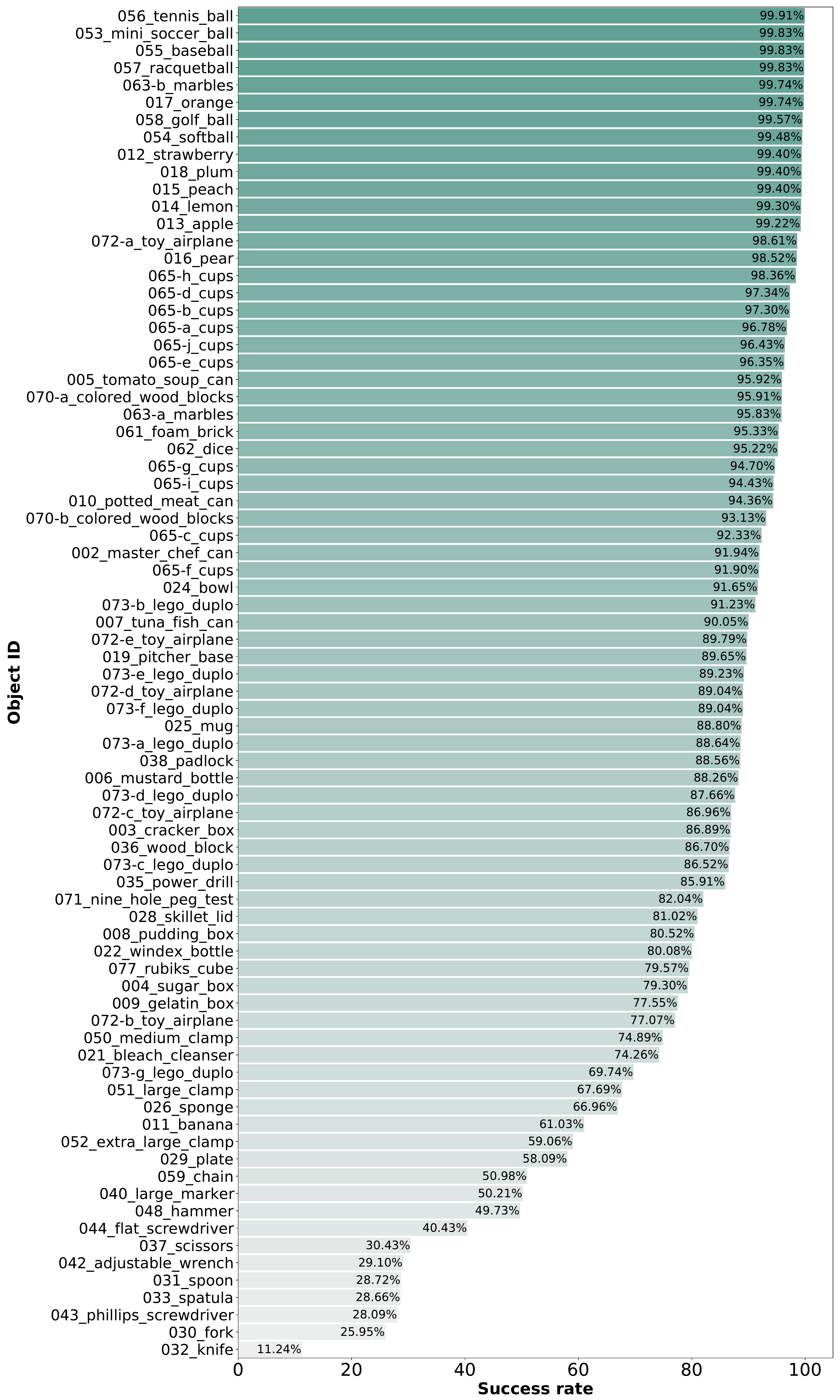}
    \caption{Reorientation success rates for each object in the YCB dataset when the hand faces upward.}
    \label{fig:success_rate_ycb_up}
\end{figure}

\begin{figure}[!htb]
    \centering
    \includegraphics[width=0.95\linewidth]{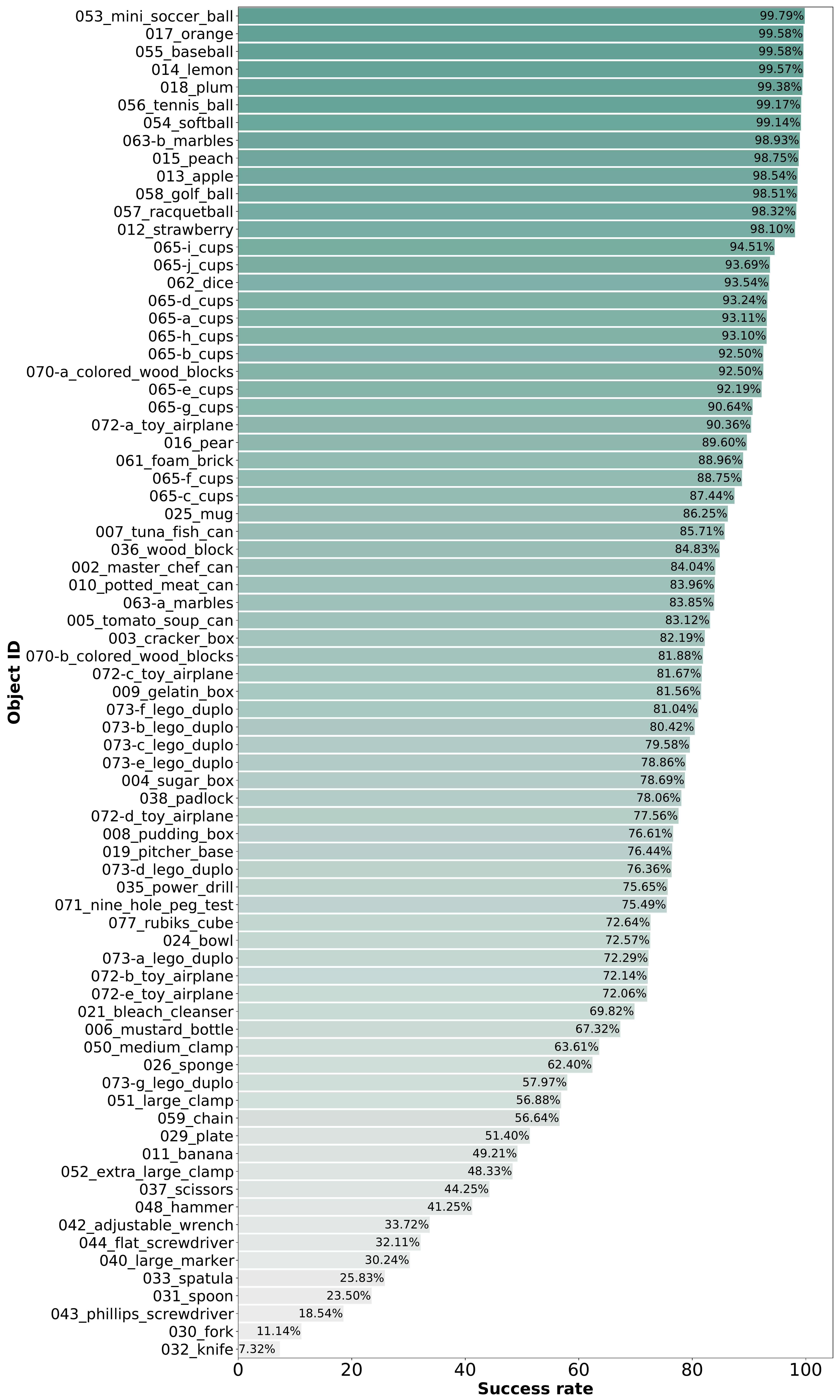}
    \caption{Reorientation success rates for each object in the YCB dataset when the hand faces downward without a table.}
    \label{fig:success_rate_ycb_down}
\end{figure}

We also analyzed the success rates on each object type in the YCB dataset. Using the same evaluation procedure described in \secref{sec:env_setup}, we get the success rates of each object using $\expertrnn$. \figref{fig:success_rate_ycb_up} shows the distribution of the success rates on YCB objects with the hand facing upward while \figref{fig:success_rate_ycb_down} corresponds to the case of reorienting the objects in the air with the hand facing downward. We can see that sphere-like objects such as tennis balls and orange are easiest to reorient. Long or thin objects such as knives and forks are the hardest ones to manipulate.

\subsection{Torque analysis}
\label{app_subsec:torque_analysis}
\begin{figure}[!htb]
    \centering
    \includegraphics[width=\linewidth]{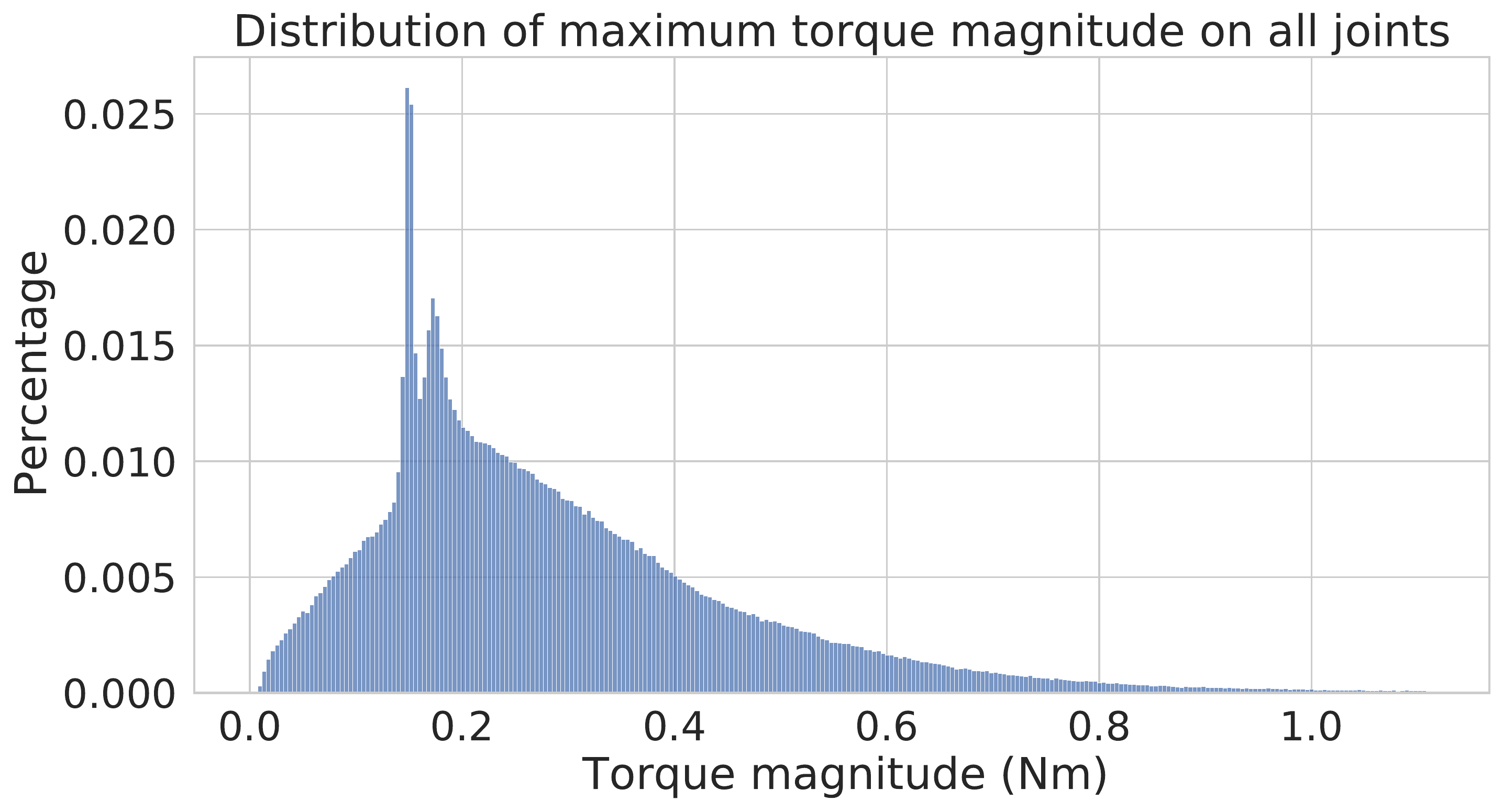}
    \caption{Distribution of the \textit{maximum} absolute joint torque values on all joints for all the time steps. We exclude a few outliers in the plot, i.e., we only plot the data distributions up to $99\%$ quantile.}
    \label{fig:torque}
\end{figure}

We randomly sampled $100$ objects from the YCB dataset, and use our RNN policy trained without domain randomization with the hand facing downward to reorient each of these objects $200$ times. We record the joint torque values for each finger joint at each time step. Let the joint torque value of $i^{th}$ joint at time step $j$ in $k^{th}$ episode be $J_{ij}^k$. We plot the distribution of $\{\max_{i=[\![1,24]\!]}|J_{ij}^k| \mid j\in[\![ 1, T]\!],  k=[\![ 1,20000]\!]\}$, where $[\![a, b]\!]$ represents $\{x\mid x\in[a,b], x\in\mathbb{Z}\}$. \figref{fig:torque} shows that the majority of the maximum torque magnitude is around \SI{0.2}{\newton\meter}. 

\subsection{Vision experiments with noise}
\label{app_subsec:vision_noise}

We also trained our vision policies with noise added to the point cloud input. We added the following transformations to the point cloud input.

We applied four types of transformations on the point cloud:
\begin{itemize}[leftmargin=*]
    \item \textit{RandomTrans}: Translate the point cloud by $[\Delta x, \Delta y, \Delta z]$ where $\Delta x, \Delta y, \Delta z$ are all uniformly sampled from $[-0.04, 0.04]$.
    \item \textit{JitterPoints}: Randomly sample $10\%$ of the points. For each sampled point $i$, jitter its coordinate by $[\Delta x_i, \Delta y_i, \Delta z_i]$ where $\Delta x_i, \Delta y_i, \Delta z_i$ are all sampled from a Normal distribution $\mathcal{N}(0, 0.01)$ (truncated at $-0.015$m and $0.015$m).
    \item \textit{RandomDropout}: Randomly dropout points with a dropout ratio uniformly sampled from $[0, 0.4]$.
    \item \textit{JitterColor}: Jitter the color of points with the following 3 transformations: (1) jitter the brightness and rgb values, (2) convert the color of $30\%$ of the points into gray, (3) jitter the color contrast. Each of this transformation can be applied independently with a probability of $30\%$ if \textit{JitterColor} is applied.
\end{itemize}

Each of these four transformations is applied independently with a probability of $40\%$ for each point cloud at every time step. \tblref{tbl:vision_pol} shows the success rates of the vision policies trained with the aforementioned data augmentations until policy convergence and tested with the same data augmentations. We found that adding the data augmentation in training actually helps improve the data efficiency of the vision policy learning even though the final performance might be a bit lower. As a reference, we show the policy performance trained and tested without data augmentation in \tblref{tbl:vision_pol}. For the mug object, adding data augmentation in training improves the final testing performance significantly. Without data augmentation, the learned policy reorients the mug to a pose where the body of the mug matches how the mug should look in the goal orientation, but the cup handle does not match. Adding the data augmentation helps the policy to get out of this local optimum.

\renewcommand{\arraystretch}{1.2}
\begin{table}[!htb]
    \centering
\centering
        \caption{Vision policy success rates when the policy is trained and tested with/without data augmentation ($\Bar{\theta}=\SI{0.2}{\radian}, \Bar{d}_C=0.01$)}
\label{tbl:vision_pol}
\begin{tabular}{clcc} 
\hline
\multicolumn{2}{c}{Object}                                        & Without data augmentation (noise) & With data augmentation (noise)  \\ 
\hline
\begin{minipage}{.06\textwidth}
\includegraphics[width=\linewidth]{figures/ycb/025_mug.jpeg}
\end{minipage} & 025\_mug &    $36.51 \pm 2.8$ &     $89.67 \pm 1.2$         \\
\begin{minipage}{.06\textwidth}
\includegraphics[width=\linewidth]{figures/ycb/065-d_cups.jpeg}      
\end{minipage}& 065-d\_cups                         &  $79.17\pm 2.3$   &  $68.32\pm 1.9$                        \\
\begin{minipage}{.06\textwidth}
\includegraphics[width=\linewidth]{figures/ycb/072-b_toy_airplane.jpeg}
\end{minipage}&072-b\_toy\_airplane            & $90.25 \pm 3.7$   &  $84.52 \pm 1.4$  \\
\begin{minipage}{.06\textwidth}
\includegraphics[width=\linewidth]{figures/ycb/073-a_lego_duplo.jpeg}
\end{minipage}& 073-a\_lego\_duplo &      $62.16 \pm 3.7$  & $58.16 \pm 3.1$ \\
 \begin{minipage}{.06\textwidth}
\includegraphics[width=\linewidth]{figures/ycb/073-c_lego_duplo.jpeg}
\end{minipage} &073-c\_lego\_duplo &     $58.21 \pm 3.9$  &     $50.21 \pm 3.7$   \\
\begin{minipage}{.06\textwidth}
\includegraphics[width=\linewidth]{figures/ycb/073-e_lego_duplo.jpeg}
\end{minipage}&073-e\_lego\_duplo & $76.57\pm 3.6$ & $66.57\pm 3.1$ \\
\hline
\end{tabular}
\end{table}
\renewcommand{\arraystretch}{1}
\end{appendices}

\end{document}